\pdfoutput=1

\documentclass{article} 

\usepackage{arxiv}

\usepackage{times}
\usepackage{cmap}
\usepackage{url}
\usepackage{hyperref}
\usepackage[T1]{fontenc}
\usepackage{graphicx}
\usepackage{subcaption}
\usepackage{enumitem}
\usepackage{cleveref}
\usepackage{booktabs}
\usepackage{longtable}  
\usepackage{makecell}  
\usepackage{adjustbox}  
\usepackage{colortbl}  

\usepackage[final]{microtype}
\newcommand{\emojirobot}{\includegraphics[height=1em]{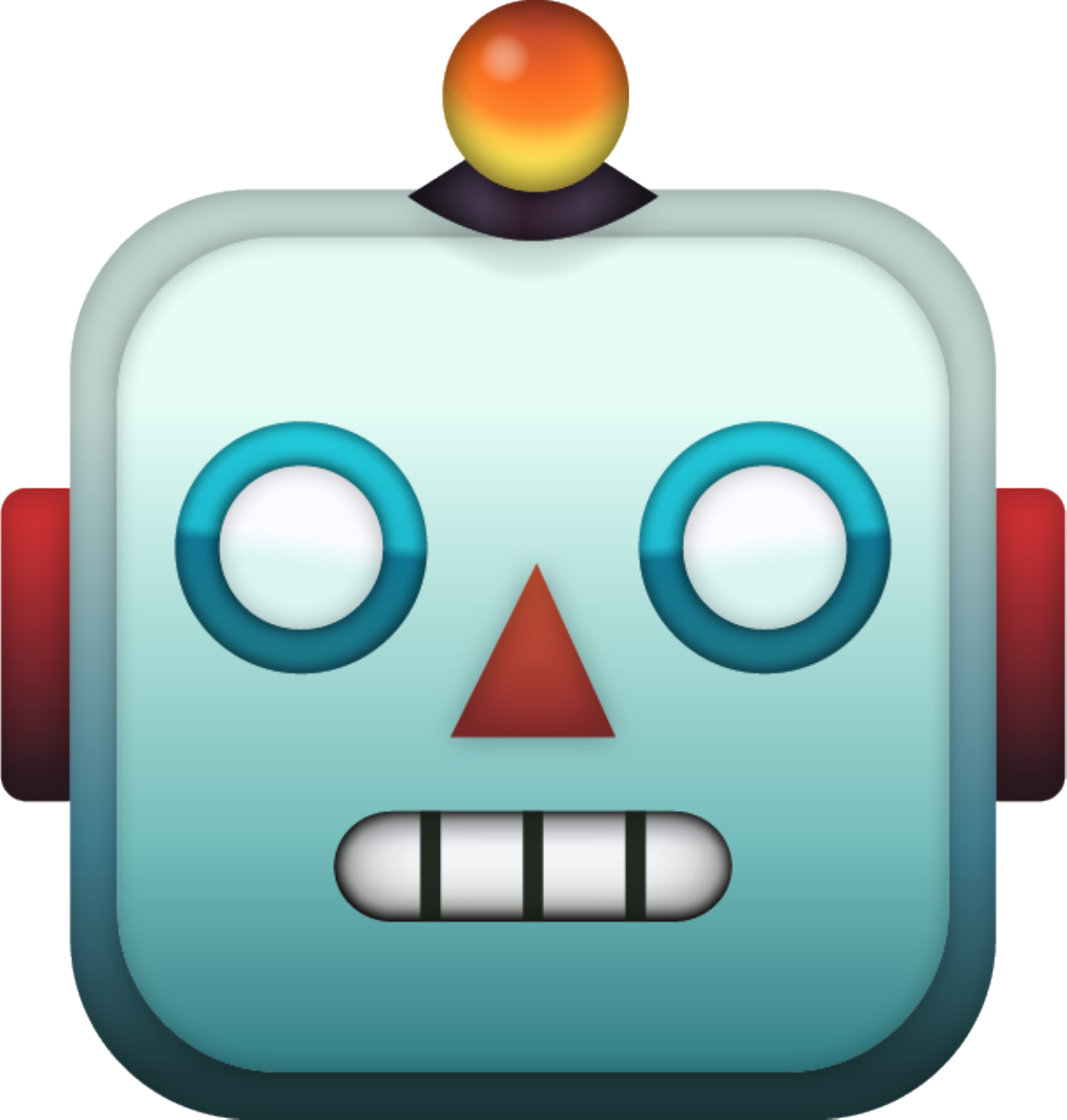}}
\newcommand{\emojiglobe}{\includegraphics[height=1em]{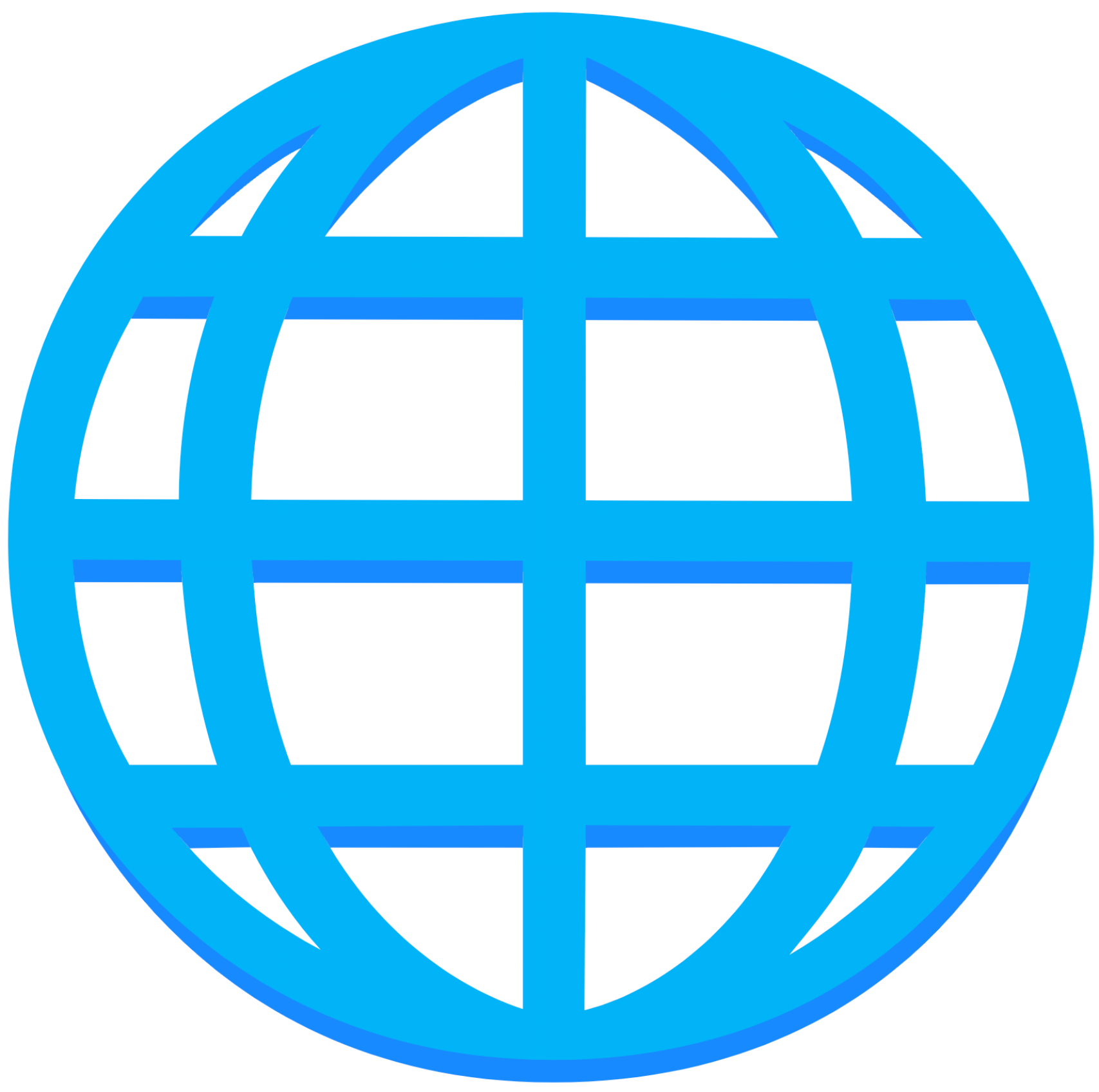}}
\newcommand{\emojiblank}{\phantom{\includegraphics[height=1em]{emojis/robot.pdf}}}
\newcommand{\NCDataCircle}{\includegraphics[height=0.9em]{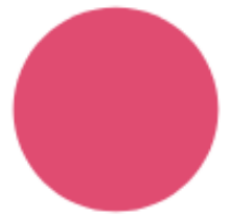}}
\newcommand{\UnspecifiedDataCircle}{\includegraphics[height=0.9em]{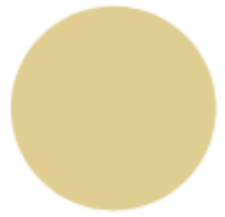}}
\newcommand{\CommercialDataCircle}{\includegraphics[height=0.9em]{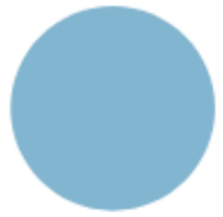}}
\newcommand{\TransparentCircle}{\phantom{\includegraphics[height=0.9em]{emojis/CommercialDataCircle.pdf}}}
\newcommand{\greencheck}{\includegraphics[height=0.9em]{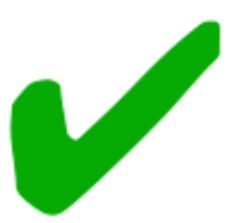}}
\newcommand{\redcross}{\includegraphics[height=0.9em]{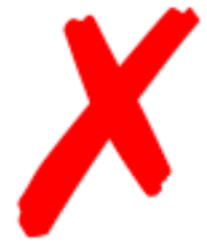}}

\title{Bridging the Data Provenance Gap Across \\Text, Speech, and Video}

\author{%
    \textbf{Shayne Longpre}, \textbf{Nikhil Singh}, \textbf{Manuel Cherep}, \textbf{Kushagra Tiwary}, \textbf{Joanna Materzynska}, \\
    \textbf{William Brannon}, \textbf{Robert Mahari}, \textbf{Naana Obeng-Marnu}, \textbf{Manan Dey}, \textbf{Mohammed Hamdy}, \\
    \textbf{Nayan Saxena},
    \textbf{Ahmad Mustafa Anis}, \textbf{Emad A. Alghamdi}, \textbf{Vu Minh Chien}, \textbf{Da Yin}, \\
    \textbf{Kun Qian}, \textbf{Yizhi Li}, \textbf{Minnie Liang}, \textbf{An Dinh}, \textbf{Shrestha Mohanty}, \textbf{Deividas Mataciunas}, \\ 
    \textbf{Tobin South}, \textbf{Jianguo Zhang}, \textbf{Ariel N. Lee}, \textbf{Campbell S. Lund}, \textbf{Christopher Klamm}, \\
    \textbf{Damien Sileo}, \textbf{Diganta Misra}, \textbf{Enrico Shippole}, \textbf{Kevin Klyman}, \textbf{Lester JV Miranda}, \\
    \textbf{Niklas Muennighoff}, \textbf{Seonghyeon Ye}, \textbf{Seungone Kim}, \textbf{Vipul Gupta}, \textbf{Vivek Sharma}, \\
    \textbf{Xuhui Zhou}, \textbf{Caiming Xiong}, \textbf{Luis Villa}, \textbf{Stella Biderman}, \textbf{Alex Pentland}, \\
    \textbf{Sara Hooker}, \textbf{Jad Kabbara}\\ \\
    \textbf{The Data Provenance Initiative}
}

\iclrfinalcopy  

\begin{document}

\maketitle

\begin{abstract}
Progress in AI is driven largely by the scale and quality of training data.
Despite this, there is a deficit of empirical analysis examining the attributes of well-established datasets beyond text.
In this work we conduct the largest and first-of-its-kind longitudinal audit across modalities---popular text, speech, and video datasets---from their detailed sourcing trends and use restrictions to their geographical and linguistic representation. 
Our manual analysis covers nearly 4000 public datasets between 1990-2024, spanning 608 languages, 798 sources, 659 organizations, and 67 countries.
We find that multimodal machine learning applications have overwhelmingly turned to web-crawled, synthetic, and social media platforms, such as YouTube, for their training sets, eclipsing all other sources since 2019. 
Secondly, tracing the chain of dataset derivations we find that while less than 33\% of datasets are restrictively licensed, over 80\% of the source content in widely-used text, speech, and video datasets, carry non-commercial restrictions.
Finally, counter to the rising number of languages and geographies represented in public AI training datasets, our audit demonstrates measures of \emph{relative} geographical and multilingual representation have failed to significantly improve their coverage since 2013.
We believe the breadth of our audit enables us to empirically examine trends in data sourcing, restrictions, and Western-centricity at an ecosystem-level, and that visibility into these questions are essential to progress in responsible AI.
As a contribution to ongoing improvements in dataset transparency and responsible use, we release our entire multimodal audit, allowing practitioners to trace data provenance across text, speech, and video.
\end{abstract}

\section{Introduction}
\label{sec:introduction}


The capabilities and flaws of multimodal foundation models are often directly attributable to their training data \citep{carlini2023quantifying,rando2022redteamingstablediffusionsafety,carlini2023, parmar2024datadataeverywhereguide, liu2023llava, liu2023improvedllava, nvlm2024}. 
While the importance of \emph{data measurement} has been widely established by prior work  \citep{gadre2024datacomp}, so has a prevailing absence of data documentation \citep{gebru2021datasheets, bender-friedman-2018-data}, transparency \citep{bommasani2023foundation}, and detailed understanding \citep{dodge2021documenting, bandy2021addressing, sambasivan2021everyone}---especially for modalities other than text.
A lack of thorough data analysis has led to significant challenges, including privacy issues \citep{Subramani2023-gj}, retracting datasets with harmful content \citep{birhane2021multimodal,David2023AIDatasetCSAM}, adversarially bypassing safety filters \citep{rando2022redteamingstablediffusionsafety}, facial recognition bias with respect to gender and skin type \citep{Buolamwini2018}, gender bias in hiring \citep{AWSGenderBias}, benchmark contamination from overlapping train and test sets~\citep{platypus2023}, and challenges in copyright \citep{henderson2023foundation}. 
Understanding data provenance can aid mitigation attempts to reduce model bias and toxicity \citep{welbl2021challenges,pozzobon2023} address representation in data \citep{xu2021detoxifying}, contamination \citep{elazars}, and quality \citep{kreutzer2022quality,marion2023moreinvestigatingdatapruning}, as well as practical challenges with identifying copyright-free and permissively licensed sets \citep{min2023SILOLM}.

Despite the urgent need for the provenance and characteristics of widely used datasets, the majority of attention to date has centered on text datasets \citep{elazars, longpre2023data}, or a single feature such as prevalence of hate content  \citep{dodge2021documenting,birhane2021multimodal}. In contrast, in this work, we will critically examine several provenance features of data \emph{across} text, speech, and video. We conduct the largest and most comprehensive multimodal audit of AI data, to date, reviewing nearly 4000 datasets between 1990-2024, covering 443 unique tasks, 608 languages, derived from 798 original sources, and constructed by 659 organizations, spanning 67 countries, over 1T tokens of text, and 1.9M hours of speech and video content (see \Cref{tab:audit-stats}).

There is an unprecedented acceleration in the development of multimodal AI systems, making all the more urgent an understanding of the datasets that underpin these breakthroughs. Our extensive collection of features from unstructured academic papers, websites, and repositories enables us to provide empirical grounding to an ambitious set of research questions surrounding data sourcing trends, intended licenses, and geographical and linguistic representation.
Our key findings include:
\begin{enumerate}[noitemsep]
    \item \textbf{Multimodal data is increasingly sourced from the web, social media platforms, or synthetically generated;} rather than more curated sources such as movies, audiobooks or manually collected. These sources comprise the vast majority of text tokens, as well as speech and video hours in public data. 
    However, while social media platforms provide data scale, heterogeneity and freshness by nature, they are also particularly prone to anti-crawling, copyright, privacy, and factuality concerns.
    
    \item \textbf{Whereas only 25\% of text, speech, and video datasets have non-commercial licenses, over 80\% of content from each modality carries undocumented restrictions in the dataset's sources.} Dataset licenses are inconsistent with their source's restrictions for over 55\% of content. 
    Our audit provides the tools for multimodal developers to identify dataset restrictions, and apply their own standards.
    \item \textbf{Geographical and linguistic representation have not improved for a decade, across the data ecosystem.} 
    While the amount of data from under-represented creators and languages increases each year, to over 600 languages and 60 countries in 2024, their \emph{relative representation} remains consistently western-centric, with no significant improvements from $>0.7$ Gini coefficients.
    While Africa and South America organizations account for $<0.2\%$ of all modality content, North America or European organizations span 93\% of text tokens and 60\%+ hours of speech and video. 
    
\end{enumerate}
Our work provides critical insights into the landscape of available multimodal data. We release the entire audit, collected data, and analysis tools, which we believe will bring immense value for data creators, developers, and researchers interested in promoting the responsible development of AI systems and analysis of the AI data ecosystem.
\vspace{-2mm}
\section{Methodology}
\label{sec:methodology}

\begin{table*}[t!]
\centering
\begin{adjustbox}{width=1.0\textwidth}
\begin{tabular}{l|cc|cc|cc|cc|c|c}

\toprule
& \multicolumn{2}{c}{\textsc{Datasets}} & \multicolumn{2}{c}{\textsc{Sources}}  & \multicolumn{2}{c}{\textsc{Creator Orgs}} & \multicolumn{2}{c}{\textsc{Languages}} & \textsc{Tasks} & \textsc{Licenses}  \\
& \textsc{\#} & \textsc{Size} & \textsc{\#} & \textsc{Domains} & \textsc{\#} & \textsc{Countries} & \textsc{\#} & \textsc{Families} & &  \\
\midrule
\textsc{Text} & 3717 & 2.1T & 713 & 23 & 534 & 60 & 502 & 21 & 395 & 50 \\
\textsc{Speech} & 95 & 775k & 51 & 16 & 124 & 29 & 260 & 36 & 18 & 19 \\
\textsc{Video} & 104 & 1.13M & 44 & 24 & 101 & 23 & - & - & 33 & 11 \\
\midrule
\textsc{Total} & 3916 & - & 798 & 83 & 659 & 67 & 608 & 37 & 443 & 55 \\
\bottomrule
\end{tabular}
\end{adjustbox}
\caption{We quantify the breadth of our audit, including the total number of datasets (\#), their size in tokens or hours, the sources, domains, creator organizations, countries, languages, tasks, and licenses. \textbf{In aggregate, we audited 3916 datasets from 659 organizations in 67 countries, spanning 2.1T tokens, and 1.9M hours. We cataloged nearly 798 unique sources, 443 tasks, and 55 licenses.}}
\label{tab:audit-stats}
\vspace{-3mm}
\end{table*}

\vspace{-2mm}
While many prior works have surveyed the dataset ecosystem \citep{albalak2024survey, liu2024datasets, malik2021automatic, prabhavalkar2023end, 8627985}, few empirically examine data corpora at scale, and those that do focus present a more narrow focus around a specific feature like geographic bias or hate content\citep{birhane2023into,mcmillan2022documenting,shankar2017no} or a single modality \citep{dodge2021documenting, caswell2021quality, elazars, longpre2023data}. The goal of this work is to provide an empirical, ecosystem-level, and multimodal analysis of widely used training datasets \citep{Cen2023}. Our audit focuses on text, speech, and video, as prominent data modalities behind modern multimodal systems, such as Sora, Whisper, Gemini, GPT-4o, and others \citep{sora2024, opensora, radford2023robust, peng2023reproducing, team2023gemini, openAIGPT4o2024}.
Since training data for modalities can often be independent, multimodal models tend to interleave training batches with different combinations of one or two modalities \citep{aghajanyan2023scaling}.
As such, we focus our analysis on datasets that represent one or a pair of these modalities.

\vspace{-2mm}
\paragraph{Annotation Features \& Methodology}
In particular, we analyze data trends for the state of data permissions (licenses and terms), sourcing (the web, human annotation, and synthetic generation), and representation (of tasks, organizations, languages, and countries).
We adopt \citet{longpre2023data}'s methodology, including the license annotation taxonomy and process, to manually audit these features precisely and rigorously.
We go beyond prior work, which considers dataset licenses, by extending the taxonomy to consider the terms of use of the sources of the dataset, either from models used to generate synthetic data (e.g. OpenAI's non-compete clause\footnote{\href{https://openai.com/policies/row-terms-of-use/}{OpenAI Terms of Use}} or Meta's acceptable use policy for Llama 3.1\footnote{\href{https://www.llama.com/llama3_1/use-policy/}{Llama 3.1 Acceptable Use Policy}}), or the source's policy on content restrictions, which can be conveyed in the form of a license, terms of use, or content policy on a website \citep{klyman2024acceptableusepoliciesfoundation}.
For each dataset, the source terms are annotated as Unrestricted, Unspecified, Source Closed or Model Closed, as defined in \Cref{tab:terms-taxonomy}.
For \Cref{fig:license-terms} we combine Source Closed and Model Closed into \emph{Restricted}.

As with prior work \citep{longpre2023data,longpre2024consent}, we engage domain experts for these annotation tasks---AI researchers whose work pertains to the modality and topic.
Because many datasets are iteratively re-packaged before they appear in their final form and often shared on popular dataset marketplaces like HuggingFace, Papers with Code or Github, prior work has found that relevant licensing terms or sourcing information for AI training data is frequently omitted \citep{longpre2023data}.
To ensure we collect this information, we require a full trace of metadata back to their original sources (sometimes a chain of github repositories, websites, or academic papers).
This search can be onerous, especially for terms and licenses, but ensures rigor in the results.
\Cref{tab:audit-stats} enumerates the full statistics of our audit.
All annotations and analysis code will be made publicly available on release.

\vspace{-2mm}
\paragraph{Scope \& Dataset Selection}
For each modality, we define the scope of the audit (detailed separately below), then aggregate resources to distill a list of relevant datasets.
The scope is focused on (a) publicly available datasets, (b) widely used tasks in the context of general-purpose model development, and (c) relevance to generative tasks.
However, we do consider classification-based datasets in text, speech, and video that can and are frequently re-purposed for generative uses (e.g. instruction tuning).
Within the defined audit scope, we use a mix of the HuggingFace Datasets platform, survey papers, survey repositories, workshop proceedings, and expert review to accumulate relevant datasets. More detail about the dataset selection and collection process is given for each modality below.
Each modality requires its own independent process, by virtue of their community dataset ecosystems being unique (discussed in \Cref{sec:discussion}).
Note that text has a wider heterogeneity of published publicly available datasets than speech or video.
Typically those datasets have been aggregated into large, standardized text-to-text collections, and as such we trace both these \emph{Text (Collections)} and their constituent \emph{Text (Datasets)}.
All datasets are described, linked, and attributed in \Cref{app:datasets}.

\vspace{-2mm}
\subsection{Text}

\vspace{-2mm}
\paragraph{Scope} We focus on providing an extensive audit for
\emph{post-training} datasets, used in training language models. We include single and multi-turn formats, encompassing both datasets typically used for instruction finetuning (SFT) and preference alignment \cite{rafailov2023direct}. This scope reflects the prominent role of general-purpose language models, which benefit from multi-task training on heterogeneous collections that span a variety of linguistic, reasoning, and knowledge intensive tasks like question answering, coding, tool use, translation, and classification \citep{weifinetuned, ouyang2022training}.

\vspace{-2mm}
\paragraph{Dataset Selection}
We expand the study conducted by the Data Provenance Collection \citep{longpre2023data}, from $44$ dataset collections (of $1858$ supervised text datasets) to a superset of $108$ collections of $3717$ datasets, prioritizing recent, popular publicly available HuggingFace Datasets introduced between 2022 and April 2024.
Our collection sourced popular datasets from recent survey papers \citep{albalak2024survey, liu2024datasets} and tools \citep{longpre2024responsible}. We additionally reviewed HuggingFace Datasets' most downloaded datasets every month, from April to July 2024, under the Natural Language Processing category, as well as the SFT/DPO datasets associated with popular open model releases.
We also drew from major multilingual data repositories, including the SEACrowd Catalogue \citep{lovenia2024seacrowd}, the Masader Arabic Data Catalogue \citep{alyafeai2022masader}, AI4Bharat \citep{kunchukuttan2020ai4bharat}, and the Aya Collection \citep{singh2024aya}.
Lastly, our list of datasets was reviewed and supplemented by language model experts to fill in notable omissions.
In total, we trace the provenance and features of 3713 text datasets from 108 collections, covering 395 popular tasks, spanning from 1994 to 2024.

\vspace{-2mm}
\subsection{Speech}

\vspace{-2mm}
\paragraph{Scope}
We audit speech datasets for which automatic speech recognition (ASR) was noted as a primary task. We focus on ASR datasets because: (1) ASR is fundamental to many speech technologies, including dictation tools, voice assistants, and chatbots~\citep{aksenova-etal-2021-might, Zhang_2022}; (2) large-scale speech datasets are typically designed for ASR~\citep{li2023yodas}; (3) ASR data follows standardized formats, making comparisons easier (e.g., corpus of audio clips paired with text); and (4) ASR data can often be reused for other tasks like text to speech (TTS)~\citep{itoLJSpeechDataset2017} or language identification~\citep{ardila-etal-2020-common}.

\vspace{-2mm}
\paragraph{Dataset Selection}
To curate a representative sample of popular ASR datasets, we relied on a combination of survey repositories\footnote{\href{https://github.com/RevoSpeechTech/speech-datasets-collection}{The Speech Datasets Collection}}, and HuggingFace Datasets using the ``Automatic Speech Recognition'' and ``Text-to-Speech'' task tags.
We expanded coverage to well-documented datasets on the OpenSLR\footnote{\href{https://openslr.org/}{openslr.org}: Open Speech and Language Resources. OpenSLR is a widely used platform in the speech community, dedicated to hosting resources for speech tasks.} platform, even if they were newer or less widely used. We expect this might reflect datasets that could be adopted more widely in the future.
Finally, we included datasets related to low-resource languages and other languages not well-covered by our initial searches. Speech recognition models are increasingly highly multilingual~\cite{babu2021xls,radford2023robust,pratap2024scaling}, and datasets serving different communities of builders and end-users around the world are a priority for making speech recognition technologies more inclusive.
In total, we trace the provenance and features of 95 speech datasets, covering 18 popular ASR tasks, spanning from 1990 to 2024.

\vspace{-2mm}
\subsection{Video} 
\vspace{-2mm}
\paragraph{Scope}
Early video understanding models primarily focused on video classification, detection and action recognition, where short clips were categorized into predefined classes \citep{video_obj_survey,zhu2020comprehensivestudydeepvideo}. More advanced tasks such as temporal action segmentation, video question answering, and video captioning were later introduced to build upon these foundational tasks \citep{moctezuma2022videocaptioningcomparativereview, zhu2023deeplearningvideotextretrieval}. Recently, following the success in the field of image generation, video generation from text has become a new task that has shown promising results \citep{sora2024,opensora, blattmann2023stablevideodiffusionscaling, esser2023structurecontentguidedvideosynthesis}. Given the scarcity of datasets for text-to-video and the often undocumented sources of data used in recent video generation models \citep{MauranOpenAI}, we take a broader approach to our collection of video datasets. We focus on annotating popular video tasks and limit our scope to datasets corresponding to video tasks that are either published, highly cited, or have 100+ downloads on HuggingFace. This approach is justified by three key factors: (1) the usefulness of video data to the research community stems from its collection and presentation in peer-reviewed work, (2) datasets can often be repurposed between different tasks, allowing for applicability to new tasks such as video generation from text, and (3) focusing on highly cited datasets ensures that datasets' quality and relevance has been validated by the research community.

\vspace{-2mm}
\paragraph{Dataset Selection} We include datasets tagged with ``Video Classification'', ``Text-to-Video'', and ``Video-Text-to-Text'' from HuggingFace Datasets.
We augmented this with datasets tagged by \href{https://paperswithcode.com/task/video-understanding}{``Video Understanding''} or \href{https://paperswithcode.com/task/video-generation}{``Video Generation''} in PapersWithCode, as well as datasets listed in a popular \href{https://github.com/xiaobai1217/Awesome-Video-Datasets}{Github survey repository}.
We also consulted the proceedings of recent video workshops: the \href{https://holistic-video-understanding.github.io/}{Large Scale Video Understanding} and \href{https://egovis.github.io/cvpr24/}{Egocentric Vision} workshops.
We separately consulted a committee of non-author video experts to supplement the list with relevant datasets published at CVPR, ICCV, ECCV, and IJCV. 
In total, we trace the provenance and features of 104 video datasets, covering 33 popular video tasks, spanning from 2009 to 2024.
\vspace{-2mm}
\section{Results}
\label{sec:results}

We discuss three key results related to (1) the rising use of web, social media and synthetic sources, (2) inconsistent and opaque restrictions on data use, and (3) a lack of improvement in geographical or linguistic representation.
Each of these findings holds across modalities, at the ecosystem level.

\vspace{-2mm}
\subsection{Rising Use of Web, Social Media \& Synthetic Data}
\label{sec:data-sources}

\begin{figure*}[!htb]
    \centering
    \includegraphics[width=\textwidth]{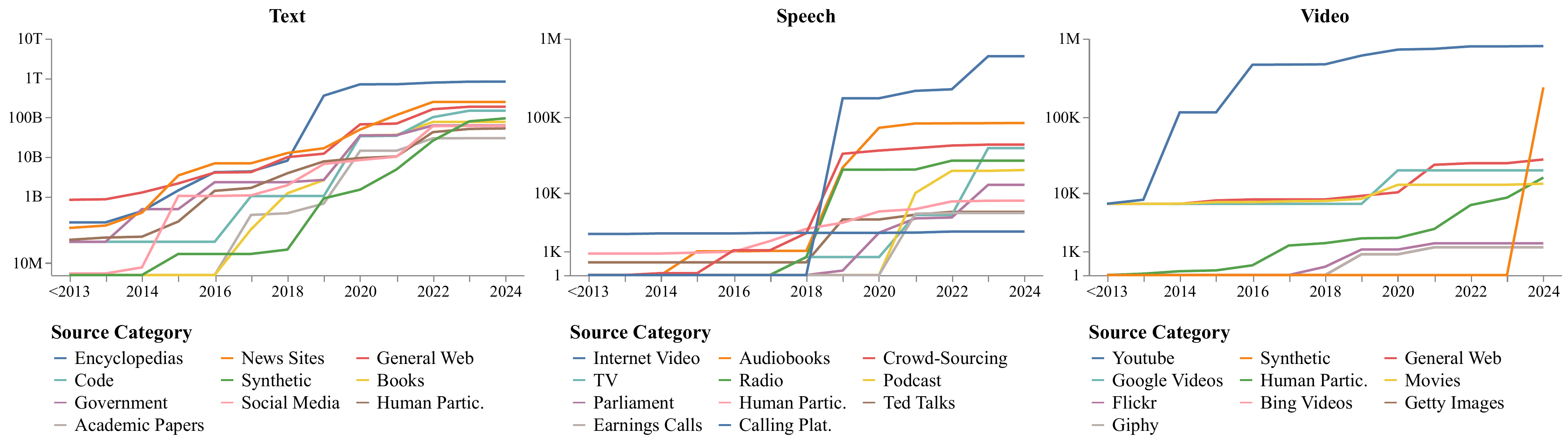}
    \caption{The cumulative size of data (log-scale tokens for text, hours for speech/video) from each source category, across modalities. The source categories in the legend are ordered by descending quantity. \textbf{Speech and video sources are increasingly dominated by internet videos and YouTube. Whereas text is predominantly web or encyclopedia-based (wiki) sources, synthetic text is rising in popularity.} }
    \label{fig:temporal-source-categories}
    \vspace{-3mm}
\end{figure*}

\vspace{-2mm}
\paragraph{The need for scale, and heterogeneity have driven rising use of data from web-crawled, social media, and synthetic data sources.}
Developers have sought out ever larger and conveniently accessible sources of training data \citep{hoffmann2022training, henighan2020scalinglawsautoregressivegenerative}.
While small, human-curated datasets are often sufficient and sometimes preferred due to higher quality, these sources often do not scale to present demands \citep{kaplan2020scaling, henighan2020scalinglawsautoregressivegenerative}.
In \Cref{fig:temporal-source-categories}, we empirically measure the rising use of web crawling and social media (or ``forum'') websites that provide some of the most scalable and fresh content.
While web-sourced data was always prominent, the balance of sources becomes much more skewed after 2018---note the use of the y-axis log scale.
We find for Speech and Video that by far the most prominent source of data has become internet videos, and specifically YouTube. 
Nearly 1M hours each of Speech and Video data from this source far outstrips the next most common sources, which comprise less than 100K hours.
For Speech, the primary data sources used to be Calling Platforms (pre-2017), content manually collected with Human Participation, and Audiobooks, but since 2018 internet videos have supplanted these other sources.
For Video, since 2013, YouTube, synthetic, and general web data sources all constitute a significantly larger portion of data used in prominent video datasets, outstripping the use of Movies, Flickr, Getty, or human curated sources.
Among text post-training datasets, we see a similar trend with general or news web-based sources, including encyclopedic sources (mainly Wikipedia), providing the majority of tokens over time.
Encyclopedic sources alone now contribute over 1T tokens in total.

\paragraph{Synthetic data sources are rising the most rapidly.}
Within the video modality, the introduction of VidProm \citep{wang2024vidprom} in 2024, consisting of nearly 7M synthetically generated videos, offered a large shift in the video source distribution.
Within the textual modality, from \cref{fig:temporal-source-categories}, synthetic data represented <0.1\% of the quantity of Web Encyclopedia data in 2020, but is now 10\% its proportion in 2024, making up the 5th largest source of tokens.
The top models used in generating datasets are mainly from OpenAI. The top 5 consist of ChatGPT, version unspecified (15.0\% of synthetic datasets), GPT-4 (14.4\%), BART (10.1\%), GPT-3 (8.3\%) and GPT-3.5-Turbo (4.9\%). 
The average synthetic dataset also has notably longer turns (in tokens) than the average natural dataset: 1,756 tokens vs 1,065.
The task distribution of textual synthetic datasets is shifted towards longer form, open-generation and creative tasks. For example, 88.1\% of natural datasets contain classification tasks, compared to only 66.3\% of synthetic datasets. Natural data is also more likely to cover translation than synthetic data (72.4\% of datasets vs only 22.9\% of synthetic datasets). 

\begin{figure*}[htbp]
    \centering
    \begin{subfigure}[b]{0.99\textwidth}
        \centering
        \includegraphics[width=\linewidth]{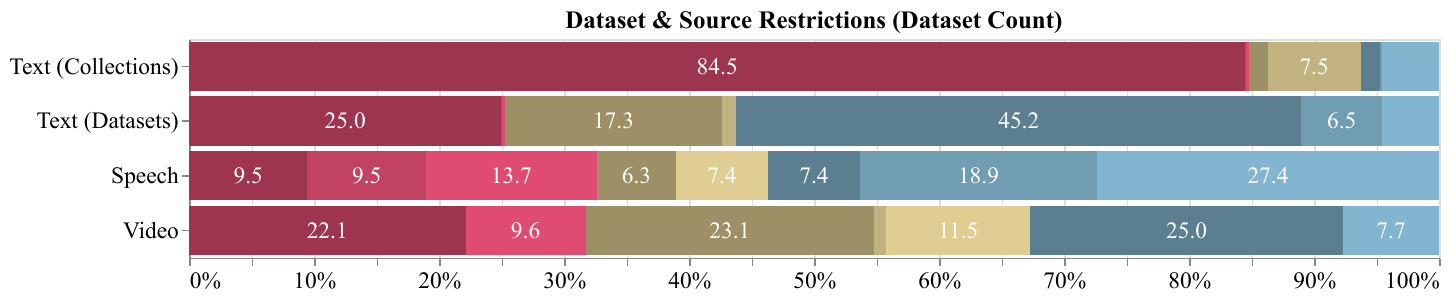} 
    \end{subfigure}
    \vspace{0.3cm}
    \begin{subfigure}[b]{0.99\textwidth}
        \centering
        \includegraphics[width=\linewidth]{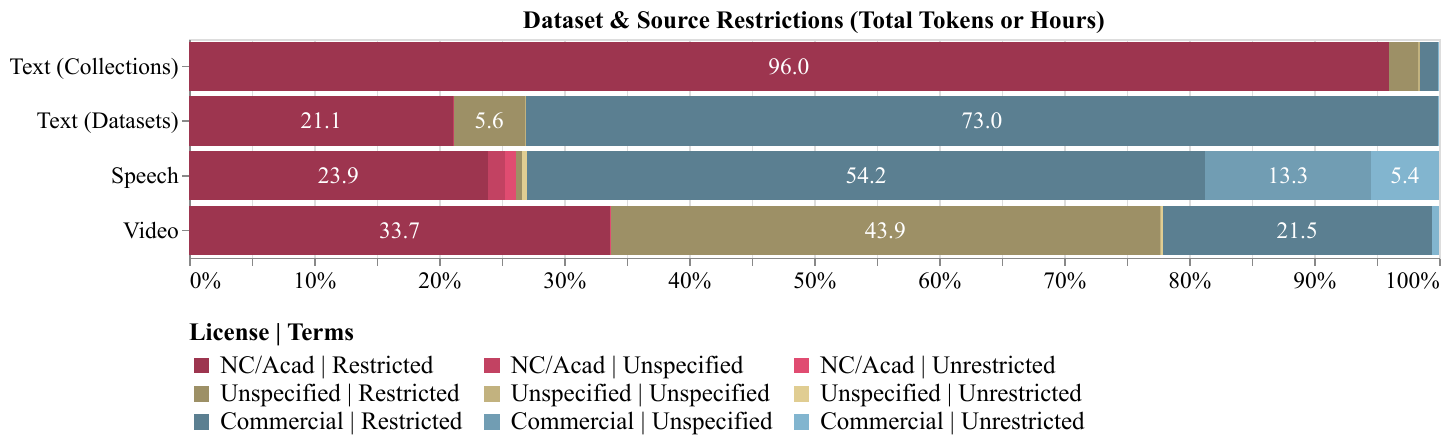} 
    \end{subfigure}
    \caption{The distribution of restrictions from dataset \emph{licenses} and their sources' \emph{terms}. We break this down by the count of datasets (top), as well as total tokens or hours (bottom). 
    Each license is categorized as Non-commercial/Academic (NC/Acad), Unspecified, or Commercially licensed.
    Each dataset may also have terms from the source: Restricted to non-commercial use, Unspecified restrictions, or Unrestricted.
    \textbf{Two main findings across modalities emerge: (1) Commercially licensed datasets represent a larger set of tokens and hours, relative to number of datasets; however, (2) the vast majority of those commercially licensed tokens/hours bare restrictions from their sources.}
    \Cref{tab:license_terms_breakdown,tab:license_terms_breakdown_count} in the appendix provide detailed numbers.}
    \label{fig:license-terms}
    \vspace{-3mm}
\end{figure*}

\vspace{-2mm}
\subsection{Inconsistent Use Restrictions}
\label{sec:use-restrictions}

In the United States, creators of a work automatically have a copyright interest that gives them exclusive rights to make copies and derivatives of the work (17 U.S.C. § 106). \emph{Licenses} are legal documents through which the owners of a work express how others may use their work. By contrast, \emph{Terms of Service} express a contract between a platform and its users to spell out how a platform and its content may be used~\citep{robinson2020beyond}.  
For simplicity, we use \emph{``Licenses''} to refer to dataset restrictions, and \emph{``Terms''} to refer to restrictions on the sources of datasets.
There remain open questions about whether certain data licenses are enforceable, but these licenses signal the intention of data creators and therefore warrant consideration as the data creators may be best positioned to understand the sensitivities of the data (privacy, copyright, representation, etc.), and the most impacted by its downstream use~\citep{morton2023licensed, lee2023talkin, mahari2023discit, mahari2023comment}.
The extent to which a practitioner adheres to dataset licenses or source terms remains an open question, and may depend on jurisdiction or the desired model's use cases \citep{lee2023talkin}. 
\emph{This work does not propose one standard for all developers.}
For these reasons we restrict our treatment and discussion here to tracing the lineage and distribution of licenses and terms for a given modality. 
    
\vspace{-2mm}
\paragraph{Data source terms are much more restrictive than the dataset's documented license restrictions.}
In \Cref{fig:license-terms}, we find only 25\%, 33\%, and 32\% of text/speech/video datasets are licensed non-commercially.
This value is even lower if we consider the proportion of tokens or hours, with 21\%, 26\%, and 33\% of text/speech/video quantities carrying license restrictions.
However, a staggering 99.8\%, 78\%, and 99\% of those quantities carry some form of non-commercial restriction on one of their sources.
For text, these restrictions are frequently from being generated by OpenAI or other models with a non-compete clause, while for speech and videos this is often since the datasets are derived from web or social media sources.

\vspace{-2mm}
\paragraph{Inconsistencies between dataset licenses and their source's restrictions pose challenges to practitioners.}
A large amount of datasets have permissive or unspecified licenses, but some set of their sources carry non-commercial restrictions.
This inconsistency is measurable---representing 79\% of tokens in text datasets, 55\% of speech hours, and 65\% of video hours.
Additionally, 19\%, 14\%, and 36\% of text, speech, and video datasets have no license or intended use documentation (from our audit of the datasets' documentation on Hugging Face Datasets, GitHub, and Papers with Code).
A lack of centralized documentation around these restrictions means it can be misleading to developers who are attempting to source data according to their own legal standards for copyright and privacy.
Furthermore, lack of documentation can hamper developers following best practices around data preparation and transparency \citep{gebru2021datasheets, bommasani2023foundation}.

\vspace{-2mm}
\paragraph{Large quantities of commercially licensed text datasets are locked in collections without clear information to separate them from restrictive datasets.}
In \Cref{fig:license-terms} (top and bottom), we see the number of datasets and number of tokens \emph{without} restrictions is significantly higher for Text (Datasets) than Text (Collections).
Specifically, 60\% more Datasets (or 75\% more tokens) are commercially licensed, than for Collections.
This demonstrates that many collections contain significant amounts of commercially licensed data.
While our audit traces licenses for all datasets within a collection, most collections do not aggregate or expose this documentation.
As a result, practitioners may be left without easy access to filter for the  subsets appropriate for their sourcing standards.

\vspace{-2mm}
\subsection{Geographical \& Linguistic Representation is Not Improving}
\label{sec:representation}

\begin{figure*}[!ht]
  \centering
  \begin{minipage}{0.99\textwidth}
    \centering
    \includegraphics[width=\textwidth]{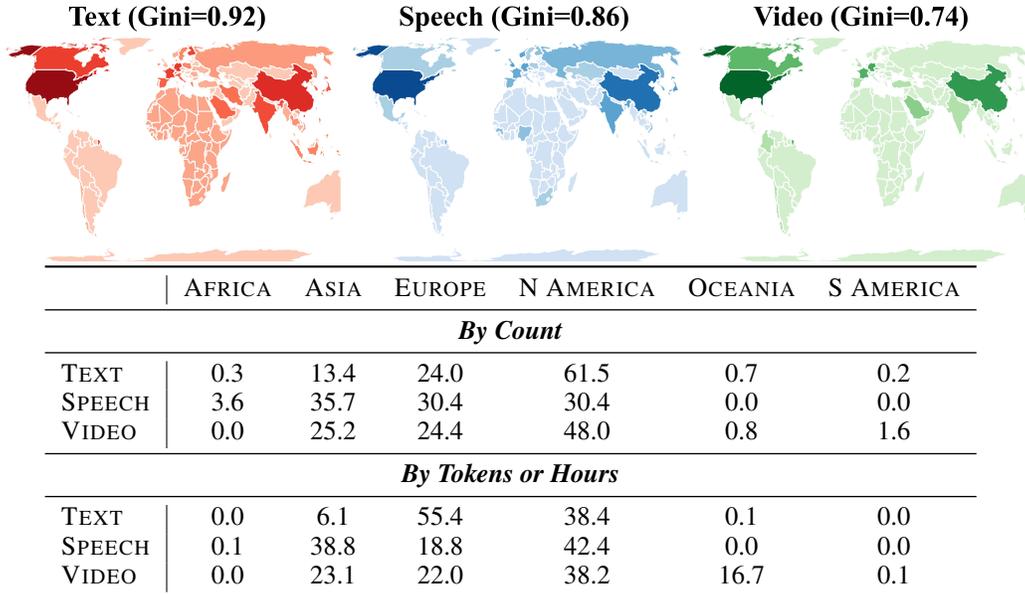}
  \end{minipage}\hfill
  \begin{minipage}{0.99\textwidth}
    \centering
    \begin{tabular}{l|cccccc}
        \toprule
         & \textsc{Africa} & \textsc{Asia} & \textsc{Europe} & \textsc{N America} & \textsc{Oceania} & \textsc{S America} \\
        \midrule
        \multicolumn{7}{c}{\textbf{\emph{By Count}}} \\
        \midrule
        \textsc{Text} & 0.3 & 13.4 & 24.0 & 61.5 & 0.7 & 0.2 \\
        \textsc{Speech} & 3.6 & 35.7 & 30.4 & 30.4 & 0.0 & 0.0 \\
        \textsc{Video} & 0.0 & 25.2 & 24.4 & 48.0 & 0.8 & 1.6 \\
        \midrule
        \multicolumn{7}{c}{\textbf{\emph{By Tokens or Hours}}} \\
        \midrule
        \textsc{Text} & 0.0 & 6.1 & 55.4 & 38.4 & 0.1 & 0.0 \\
        \textsc{Speech} & 0.1 & 38.8 & 18.8 & 42.4 & 0.0 & 0.0 \\
        \textsc{Video} & 0.0 & 23.1 & 22.0 & 38.2 & 16.7 & 0.1 \\
        \bottomrule
    \end{tabular}
  \end{minipage}
  \vspace{1mm}
  \caption{The geographical distribution of countries (world maps) and continents (table) represented by dataset creators. \textbf{Despite some differences in European, Russian, and Middle Eastern representation, creators are heavily concentrated in the US, China, and  Western Europe, with little to no representation in South America or Africa, across modalities.} The current Gini coefficient for (Text, Speech, Video) = (0.92, 0.86, 0.74), where higher values indicate more concentration.}
  \label{fig:creator-worldmaps}
  \vspace{-3mm}
\end{figure*}

\vspace{-2mm}
\paragraph{The importance and progress of representation in AI training data.}
Diversity and representation in training datasets, and among their creators, are widely acknowledged as essential to building AI models that are less biased, more useful, and more equitable \citep{joshi2020state,singh2024aya,ustun2024aya,adelani2021masakhaner,adelani2024irokobenchnewbenchmarkafrican,aakanksha2024multilingualalignmentprismaligning, mcmillanmajor2022documentinggeographicallycontextuallydiverse, porgali2023casualconversationsv2dataset, monfort2019moments, sigurdsson2016hollywoodhomescrowdsourcingdata}.
Prior work has measured the diversity of languages in data along with cultural, ideological, and geographical imbalances \citep{faisal2022dataset,shankar2017no,mcmillan2022documenting,de2019does,mahadev2021understanding}.
These studies have exposed significant flaws, often in the form of bias and discrimination, stemming directly from poor representation in data \citep{buolamwiniGenderShades2018, birhane2021multimodal}.
As this problem has now been widely acknowledged for decades, recent efforts have foregrounded sourcing data multilingually and multi-culturally, from native speakers and creators (e.g. ROOTS \citep{NEURIPS2022_ce9e92e3}, the Aya Dataset \citep{singh2024aya}, the SEACrowd Catalogue \citep{lovenia2024seacrowd}, the Masader Catalogue \citep{alyafeai2022masader}, Common Voice \citep{ardila2019common}, Causal Conversations V2 \citep{porgali2023casualconversationsv2dataset} or Moments in Time \citep{monfort2019moments}).

\begin{figure*}[htbp]
    \centering
    \begin{subfigure}[b]{0.49\textwidth}
        \centering
        \includegraphics[width=\linewidth]{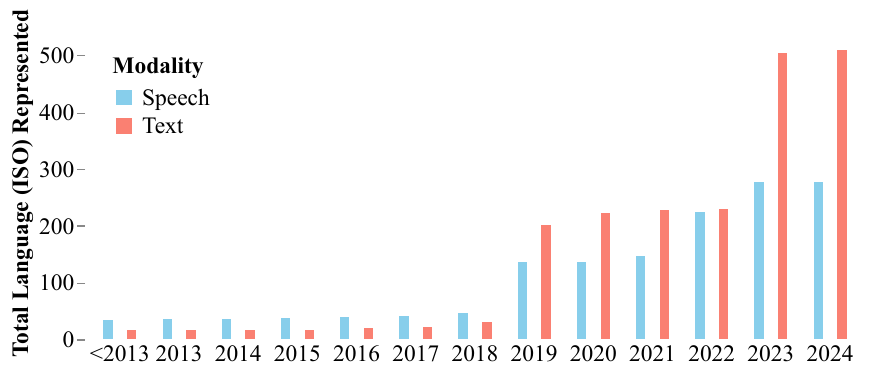} 
        \caption{Total Language Representation}
    \end{subfigure}
    \hfill
    \begin{subfigure}[b]{0.49\textwidth}
        \centering
        \includegraphics[width=\linewidth]{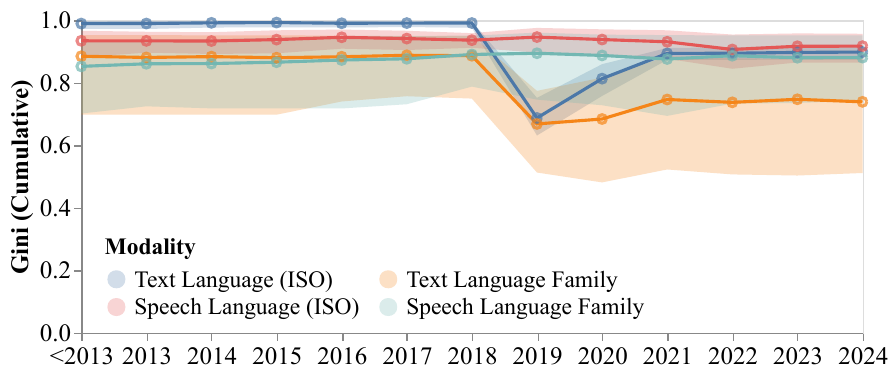} 
        \caption{Gini-coefficient for Language Representation}
    \end{subfigure}

    \vspace{0.3cm}

    \begin{subfigure}[b]{0.49\textwidth}
        \centering
        \includegraphics[width=\linewidth]{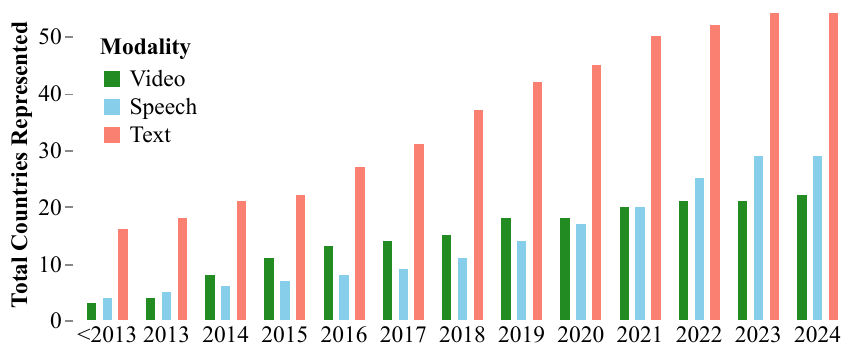} 
        \caption{Total Geographical Representation}
    \end{subfigure}
    \hfill
    \begin{subfigure}[b]{0.49\textwidth}
        \centering
        \includegraphics[width=\linewidth]{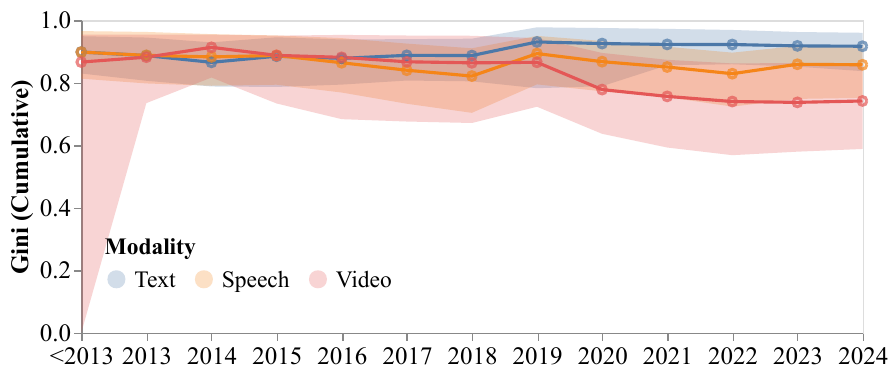} 
        \caption{Gini-coefficient for Geographical Representation}
    \end{subfigure}
    
    \caption{The cumulative totals (left) of languages and countries represented in the data over time, and the 95\% confidence intervals of the gini-coefficients over time (right) to measure the representativeness of these variables. Gini-coefficients are a measure of statistical dispersion, frequently used to quantify inequality. 
    A Gini coefficient of 1 indicates highest concentration, and lower values more broad representation.
    \textbf{While the number of represented languages and geographies continue to rise (left), the equality of their distribution has in most cases, not significantly changed.} 
    }
    \label{fig:representation}
\end{figure*}

\vspace{-2mm}
\paragraph{Measuring geographical and linguistic representation.}
Naturally, we aim to use our audit to measure the progress of these efforts on geographical and linguistic representation in the AI ecosystem.
We measure the progress of two forms of representation: (1) language diversity of text and speech data, and (2) geographical diversity of the creators, in all three modalities.
For languages, we use the ISO 639-1 and 639-3 language codes and categories of language families from Glottolog 5.0.\footnote{We use top level \href{https://glottolog.org/}{Glottolog} families.}
In \Cref{fig:representation}(a, c) we display the cumulative sum of unique languages and countries present across all audited datasets, at each time period since 2013.
While these measurements illustrate the absolute rise in diversity, we also hope to measure the relative dispersion, or equality of languages and countries in the distribution.
In \Cref{fig:representation}(b, d), we use the Gini Index \citep{e7461144-3bdb-3344-8d83-b11a52be7f2f, atkinson1970measurement}, a traditional measure of statistical dispersion, frequently used to quantify inequality.
This allows us to understand if the distributions of languages and creators are more representative of the international community over the last decade, or equally concentrated despite apparent efforts at the margins.

\vspace{-2mm}
\paragraph{Inequality in geographical representation remains very high, with few organizations creating datasets from the Global South.}
For every dataset, our audit recorded the organizational affiliations of each creator of the dataset.\footnote{A dataset creator, following \citep{longpre2023data}, is defined as an organization associated with the release of the dataset as created for machine learning---not any of the upstream sources. More details in \Cref{app:datasets}.}
These organizations were then manually mapped to the country in which they are headquartered. 
Occasionally, organizations like BigScience, BigCode, or Masakhane have international or continental representation, and were counted as such.
In \Cref{fig:creator-worldmaps}, we measure the current state of diversity among these creator organizations---where a Gini coefficient of 1 indicates highest concentration, and lower values more broad representation.
Without taking up the normative question of what a truly ``fair'' score would be, these values provide useful comparisons across modalities and over time.
We find that Text dataset developers are particularly homogeneous, with a Gini-coefficient of 0.92; followed by Speech, at 0.86 and Video at 0.74, which remain high, but are meaningfully less concentrated. \Cref{fig:creator-worldmaps} also illustrates that even this limited diversity is still concentrated in North America, Europe, East Asia, and less so in the Global South.

In \Cref{fig:creator-worldmaps}, we also compare the distribution of datasets, and of tokens or hours by continent. 
Dataset creators affiliated with African or South American organizations account for fewer than 0.2\% of all tokens or hours, in each modality.
In contrast, Asian affiliated organizations represent large proportions of the data, particularly for speech (39\% of hours, attributed predominantly to YODAS \citep{li2023yodas}).
Much of this driven by Chinese, Indian, Russian, and Saudi Arabian creators.
Most prominently, the combination of North American and European datasets comprises 93\% of text tokens, 61\% of speech hours, and 60\% of video hours.

\vspace{-2mm}
\paragraph{Geographical representation has not significantly improved for over a decade.}
In \Cref{fig:representation}(c), we measure the total unique number of countries represented across all dataset creator organizations. While individual creators will have varying ethnic and national affiliation, we treat this as an estimate for the influence of each locale in dataset development. We find that while the number of represented countries has risen steadily each year, for each modality, this represents only an illusion of progress. Empirically, the Gini coefficient for each modality has not significantly changed since the start of the period we examine in 2013. Geographic diversity has increased only among Video datasets, and these increases are not significant at the $p = 0.05$ level. Text and Speech geographical representations appear to remain stable over the last decade of AI development.

\vspace{-2mm}
\paragraph{Multilingual representation has not improved by most measures.}
Similar to geographical representation, we measure the cumulative number of ISO 639-1 languages and language families over time, as well as the per-modality Gini-coefficient.
\Cref{fig:representation}(a) shows significant increases in the number of languages available for speech and text, especially in 2019, and 2023, with the introduction of large sets like Flores \citep{goyal2022flores}, xP3x \citep{muennighoff2023crosslingual}, Common Voice \citep{ardila2019common}, and the Aya Collection \citep{singh2024aya}. However, once again, when measuring the cumulative dispersion of these datasets in \Cref{fig:representation}(b), only Text language families demonstrate any improvement from pre-2013 to the present. Improvements in the Gini coefficient appear to be largely driven by individual large-scale projects like xP3x and Common Voice, both introduced in 2019. Subsequently, newer datasets remain predominantly monolingual, causing measures of concentration in text languages, speech languages, and language families to remain consistently high.

\begin{figure*}[!htb]
    \centering
    \includegraphics[width=\textwidth]{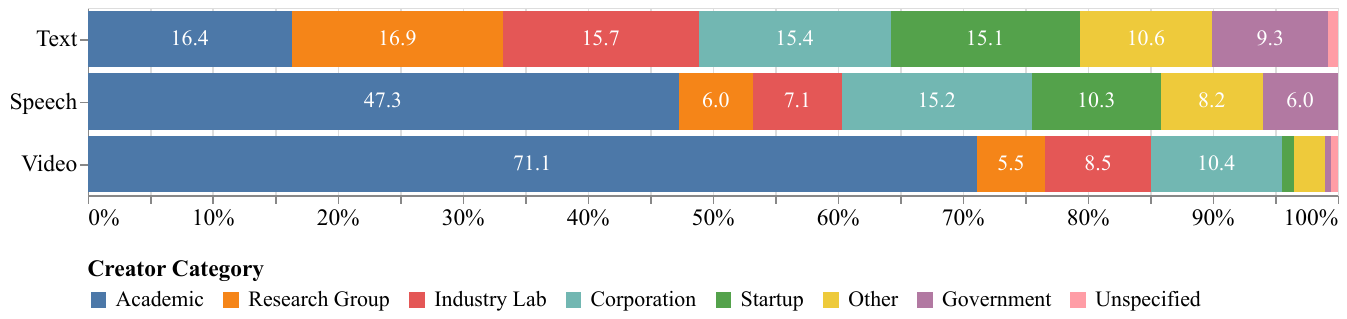}
    \caption{The distribution of creator organizations by modality. \textbf{Most public speech and video datasets are developed by academic organizations, whereas text datasets are developed by a wide mix of academia, non-profit or industry labs, as well as startups.} 
    }
    \label{fig:creator-orgs}
    \vspace{-3mm}
\end{figure*}

\vspace{-2mm}
\paragraph{Academia, research non-profits, and industry labs continue to drive public dataset development.}
As well as understanding the geographic associations of the organizations creating popular datasets, we manually categorize them into: Academic Organization (e.g., universities), Research Groups (e.g., non-profits such as BigScience, EleutherAI or AI2), Industry Labs (e.g., Cohere For AI, Google DeepMind), Corporations (e.g. Google, Meta), Startups (e.g., OpenAI, Anthropic), Governments, Unspecified (datasets where owner affiliation is not shared), or Other.
When a dataset is released in collaboration between organizations, we record each organization.
In \Cref{fig:creator-orgs}, we find that universities and other academic organizations account for 16\%, 47\%, and 71\% of all recorded dataset releases, across Text, Speech, and Video respectively.
Research groups, industry labs and even corporations are also significant contributors, especially for Text datasets, where ecosystem contributors are far more distributed.
The significant role of academic organizations in Video and Speech may suggest that the risk profile of releasing Text datasets differs somewhat from Video and Speech datasets, which may have more distinct privacy concerns.
\section{Discussion}
\label{sec:discussion}

\vspace{-2mm}
\paragraph{The rise of web-based, social media, and synthetic datasets may pose greater risks to privacy, copyright, and bias.}
\Cref{sec:data-sources} discusses the rise of web-based sources and particularly social media as primary sources for speech and video.
\Cref{fig:temporal-source-categories} shows these sources now exceed more traditional, curated sources such as movies, audiobooks, radio, TV, or content hand-crafted by human participants---by at least one order of magnitude.
These websites made of mostly user-generated content are a natural choice, given that they scale in the quantity, freshness, and heterogeneity that is best suited to train general-purpose models \citep{longpre2023pretrainers,aghajanyan2023scaling}.
However, prior work suggests that crowd-sourced, user-generated web content also introduces more challenges than curated content, particularly for privacy, copyright, bias, harm, and factuality.

Web-based and particularly user-generated content is disproportionately likely to include personally identifiable information (PII) \cite{Luccioni2021-iq,Subramani2023-gj,elazars}, and copyrighted content \citep{meese2019mundane,lee2023talkin}.
These can be reproduced in the outputs of AI models \citep{Carlini2022-nk,Chen2023-lw}, creating privacy and copyright concerns \citep{Zhang2023-ln}. 
Open datasets being used to train GPAI often attempt to filter---but frequently miss---PII and copyrighted data \citep{soldaini2024dolma,Subramani2023-gj} (although not all do \citep{Penedo2023-cr}). 
Social media, in particular, is also known to have bias, toxicity and factuality issues \citep{olteanu2019social}, which can manifest in trained models, even after alignment \citep{kotha2023understanding}.
Lastly, while synthetic data can help reduce the prevalence of PII, copyright, or bias in data, it comes with its own challenges \citep{Kurakin2023-ip,Liu2024-ga}.

\vspace{-2mm}
\paragraph{Social Media websites have become one of the most prominent data sources, but their Terms often restrict crawling or commercial use.}

We find that 71\% of Video data and 69\% of Speech data is from YouTube which has become a prominent source of data, given its scale, freshness, and multimodality (containing videos, speech, images, and text) \citep{abu2016youtube, NEURIPS2018_35309226, NEURIPS2020_2cd4e8a2, NEURIPS2023_5c61452d, coats2023dialect, li2023yodas}.
However, YouTube is a social media platform owned by Google and its Terms of Service\footnote{\href{https://www.youtube.com/static?template=terms}{YouTube Terms of Service}.} prohibit third parties from crawling YouTube.
While content creators maintain their ownership rights in the material they upload to YouTube, the YouTube Terms of Service also grant Google a license to reproduce, modify, display, and use the content for purposes connected to YouTube's ``business'', which may include building machine learning models; even if the copyright holder has selected a permissive license, YouTube's Terms   disallow external parties from crawling that data.
Model developers such as Nvidia and OpenAI have been sued in the U.S. by content creators who allege that they unlawfully trained on YouTube videos \citep{Cole2024, Skolnik2024}. 
Large social media platforms and forums have also adopted restrictive terms in recent years, including Reddit and StackOverflow.\footnote{\href{https://redditinc.com/policies/user-agreement-september-25-2023}{Reddit User Agreement} and \href{https://stackoverflow.com/legal/terms-of-service/public}{StackOverflow Terms of Service}.}
As these data sources become critical to scaling AI systems, access has been made exclusive, which may hamper academic, non-profit, or open source model development---to the extent that social media platforms can enforce their terms against third party developers.\footnote{We treat the enforceability of licenses and terms as an open legal question, beyond the scope of our work.}

\vspace{-2mm}
\paragraph{Ambiguous and poorly documented use restrictions may significantly inhibit model developers adhering to cautious legal and ethical data sourcing standards.}
In \Cref{sec:use-restrictions}. we find that a significant amount of data carry non-commercial restrictions in their sources, rather than on the final dataset, which can contain no license or a permissive one.
For text and video, these restrictions can equate to 99\% of all tokens and hours.
These inconsistencies are the result of datasets being iteratively re-packaged and re-licensed, without carrying on documentation \citep{longpre2023data}.
While not every developer will employ the same filtering standards, our work shows that the challenges to separate and identify appropriate datasets remain difficult across these modalities.
Without continued audits and documentation, practitioners may be forced to forego large collections of partially viable data, hampering data scaling laws \citep{kaplan2020scaling}, or take on avoidable risk.
We hope this released audit will provide greater tools for practitioners to apply their own standards, to make informed decisions on training data use.

\vspace{-2mm}
\paragraph{The limitations of measures of geographical and linguistic representation.}
It is important to note that measures of geographical and linguistic representation are imperfect. We are limited by partial information about the developers' identities (including for privacy reasons), limited transparency into how frequently these datasets are used, and the extent to which proprietary datasets may fill in representation gaps behind closed doors.
Nonetheless, we believe the breadth and rigour of the audit make this the best available empirical measure of representation in \emph{publicly} documented datasets.
Further, we propose the goal of measuring representation in AI data as essential to understanding progress, or its absence, towards AI systems that fairly serve the broader community of users.
\Cref{fig:creator-worldmaps} and \Cref{fig:representation} demonstrate that despite the absolute rise of geographical and linguistic representation, the relative western-centric concentration persists, across thousands of surveyed datasets.
We release all audit materials for transparency and replicability, and for further use by the research community.

\vspace{-2mm}
\paragraph{Conducting representative analyses of an ecosystem comes with assumptions.}
First, an ecosystem for AI is by nature, not centralized or organized.
Widely used datasets for Text are often hosted on Hugging Face, but this is frequently not the case for Speech or Video.
Similarly, while Text data undergoes frequent dataset re-packaging for general-purpose post-training, this is not true to the same extent for other modalities.
As such, the scope and dataset selection process need to be designed for each modality, rather than a single, simple protocol, which inevitably will not accurately represent one modality at its ecosystem-level.
Similarly, we chose a subset of modalities of interest to foundation model development \citep{sora2024, radford2023robust}, but note there are many other left for future work (e.g., images, 3D representations, tabular, time series, graphs, and geospatial data).
\subsubsection*{Acknowledgments}
\label{sec:acknowledgements}
This research was conducted by the Data Provenance Initiative, a collective of independent and academic researchers volunteering their time to data transparency projects. The Data Provenance Initiative is supported by the Mozilla Data Futures Lab Infrastructure Fund.

\printbibliography  

\clearpage

\appendix
\section{Extended Related Work}
\label{app:related-work}
Progress in machine learning across modalities from speech \citep{radford2023robust} to vision \citep{dosovitskiy2021image} to text \citep{browngpt3, weifinetuned} has benefited from advancements in large pre-training and fine-tuning corpora. The development of multimodal corpora has also been key to several recent advances, as with CLIP in the image/text domain \cite{clip_radford2021learning}, CLAP for audio/text settings \cite{elizalde2022clap}, and a number of other models involving both text and images, audio or video \citep{radford2023robust, ramirez2024anatomy, singerMakeVideoTextVideoGeneration2022, rameshHierarchicalTextConditionalImage2022}.

The datasets powering these advances are not, however, always well-documented, despite the existence of standards and frameworks for recording and annotating dataset metadata that range from `data statements' \citep{bender-friedman-2018-data} to `datasheets for datasets' \citep{gebru2021datasheets} and others \citep{mitchell2019model}. The key problem is not a deficiency of any particular framework, but rather inconsistent adoption and fragmentation \citep{Longpre2024Data}. Much prior work has argued for the need to document and audit these datasets \citep{rogers-2021-changing, paullada2021data}, motivated by concerns from reproducibility \citep{kapoor2022leakage} to interpretability \citep{longpre2023pretrainers} to bias and fairness problems that may stem from problematic content in training data \citep{birhane2021multimodal}. 

There have been several attempts to carry out such audits, with prior work examining pretraining data \citep{longpre2024consent}, general web corpora \citep{gao2020pile, dodge2021documenting}, instruction fine-tuning datasets \citep{longpre2023data}, and the documentation fields of the HuggingFace Datasets platform in particular \citep{yang2024navigatingdatasetdocumentationsai}. For speech and vision, there has been less work, with many discussions of datasets in the aggregate occurring in survey papers \citep{SchiappaSSLV2023, chaquetVideoDatasetSurvey2013}, research aimed directly at improving model performance \cite{GadreDataComp2023} or close examinations of questions like bias in small groups of datasets \citep{buolamwiniGenderShades2018,romanou2024includeevaluatingmultilinguallanguage}. 

Prior work has also examined the identities, affiliations and national origin of paper authors \citep{movva2024topics} in AI, but an analogous look at the producers of datasets is lacking. We aim to carry out such analyses: replicating those for pretraining and text finetuning datasets in video and audio domains, and surveying provenance and legal status. Finally, there has also been significant recent attention to legal questions in the collection and use of AI training data \citep{Sag2020TheNL, henderson2023foundation}. The complex process involved in preparing these datasets \citep{lee2023talkin}, and the ambiguous licensing of inputs, can make understanding the legal status of the final output quite difficult.
\section{Dataset Licenses \& Terms}
\label{app:licenses-and-terms}

\paragraph{Detailed taxonomy}
We code the legal restrictions placed on use of datasets along two axes. First, we identify whether a dataset's license permits commercial use (``Commercial'' in Table \ref{tab:license_terms_breakdown}), only non-commercial / academic use (``NC / Acad''), or does not clearly specify what is permitted (``Unspecified''). 
The latter category includes datasets for which we were unable to locate a license.
Datasets which are in the public domain and not subject to a license are counted as commercially usable.
Second, we annotate the contractual or terms-of-use restrictions placed on dataset use by the source of each dataset. There are four levels, defined in Table \ref{tab:license_terms_breakdown}. Note that the Model Closed status can only apply to datasets that are AI-generated, at least in part.
Some datasets can carry both Model Closed and Source Closed status, but we count the Model Closed first for simplicity.

\begin{table*}[t!]
\centering
\begin{adjustbox}{width=1.0\textwidth}
\begin{tabular}{p{0.19\textwidth}|p{0.81\textwidth}}
\toprule
\textsc{Label} & \textsc{Definition} \\
\midrule
\textsc{Model Closed} & A model used to generate part or all of the dataset prohibits using its outputs commercially, to develop a competing AI model, or in general. \\
\textsc{Source Closed} & The source has a license or terms that prohibits use of the data, either commercially, from being crawled, to develop AI, or in general. \\
\textsc{Unspecified} & No information can be found relevant to restrictions, or lack thereof, for this source. \\
\textsc{Unrestricted} & The source has a commercially permissive license, such as CC BY, or explicitly states the data is open for broad use. \\
\bottomrule
\end{tabular}
\end{adjustbox}
\caption{\textbf{The taxonomy used to determine use restrictions on each dataset source.} Each source in a dataset is examined and fit into one of these categories. The dataset Terms are then labelled according to the strictest terms across the sources, with Model Closed and Source Closed considered stricter than Unspecified which is in turn stricter than Unrestricted.}
\label{tab:terms-taxonomy}
\end{table*}

\paragraph{Detailed breakdown}
Tables \ref{tab:license_terms_breakdown} and \ref{tab:license_terms_breakdown_count} present crosstabs of these two dimensions, according to respectively the total amount of content and the number of datasets. The most notable finding, as discussed in the main text, is the frequency of clashing restriction status between licenses and terms. By amount of content, fully 73.0\% of text content, 55.0\% of speech content, and 21.6\% of video content is subject to a license permitting commercial use but also to terms restrictions forbidding it, or the reverse. The absolute level of restrictions is also high, with < 0.1\% of text content, 5.4\% of speech content, and 0.6\% of video content usable for commercial purposes under both licenses and terms.

\begin{table*}[!htb]
\centering
\begin{adjustbox}{width=0.9\textwidth}
\begin{tabular}{l|rrr|r}
\toprule
\textsc{License / Terms} & \textsc{Restricted} & \textsc{Unspecified} & \textsc{Unrestricted} & \textsc{Total} \\
\midrule
\multicolumn{5}{c}{\textbf{\emph{Text Collections}}} \\
\midrule
\textsc{NC/Acad} & 96.0 & 0.0 & 0.0 & 96.0 \\
\textsc{Unspecified} & 2.3 & 0.1 & 0.0 & 2.4 \\
\textsc{Commercial} &1.5 & 0.0 & 0.0 & 1.6 \\
\midrule
\textsc{Total} & 99.8 & 0.1 & 0.1 & \\
\midrule
\multicolumn{5}{c}{\textbf{\emph{Text Datasets}}} \\
\midrule
\textsc{NC/Acad} & 21.1 & 0.0 & 0.0 & 21.2 \\
\textsc{Unspecified} & 5.7 & 0.1 & 0.0 & 5.7 \\
\textsc{Commercial} & 73.0 & 0.0 & 0.0 & 73.1 \\
\midrule
\textsc{Total} & 99.8 & 0.1 & 0.1 & \\
\midrule
\multicolumn{5}{c}{\textbf{\emph{Speech Datasets}}} \\
\midrule
\textsc{NC/Acad} & 23.9 & 1.4 & 0.8 & 26.2 \\
\textsc{Unspecified} & 0.5 & 0.0 & 0.4 & 0.9 \\
\textsc{Commercial} & 54.2 & 13.3 & 5.4 & 73.0 \\
\midrule
\textsc{Total} & 78.6 & 14.7 & 6.7 & \\
\midrule
\multicolumn{5}{c}{\textbf{\emph{Video Datasets}}} \\
\midrule
\textsc{NC/Acad} & 33.7 & 0.0 & 0.1 & 33.8 \\
\textsc{Unspecified} & 43.9 & 0.1 & 0.1 & 44.1 \\
\textsc{Commercial} & 21.5 & 0.0 & 0.6 & 22.1 \\
\midrule
\textsc{Total} & 99.1 & 0.1 & 0.8 & \\
\bottomrule
\end{tabular}
\end{adjustbox}
\caption{\textbf{A breakdown of the percentage of license and terms restrictions across datasets}, by total tokens or hours of content. The much higher frequency of restrictions at the collection level is because we consider a collection's license or terms status to be the most restrictive of those for its datasets. Note that percentages may not add to exactly 100\% because of rounding. 
}
\label{tab:license_terms_breakdown}
\vspace{-2mm}
\end{table*}

\begin{table*}[!htb]
\centering
\begin{adjustbox}{width=0.9\textwidth}
\begin{tabular}{l|rrr|r}
\toprule
\textsc{License / Terms} & \textsc{Restricted} & \textsc{Unspecified} & \textsc{Unrestricted} & \textsc{Total} \\
\midrule
\multicolumn{5}{c}{\textbf{\emph{Text Collections}}} \\
\midrule
\textsc{NC/Acad} & 84.5 & 0.0 & 0.3 & 84.8 \\
\textsc{Unspecified} & 1.5 & 7.5 & 0.0 & 8.9 \\
\textsc{Commercial} & 1.5 & 0.2 & 4.5 & 6.3 \\
\midrule
\textsc{Total} & 87.5 & 7.7 & 4.8 &  \\
\midrule
\multicolumn{5}{c}{\textbf{\emph{Text Datasets}}} \\
\midrule
\textsc{NC/Acad} & 25.0 & 0.0 & 0.3 & 25.3 \\
\textsc{Unspecified} & 17.3 & 1.2 & 0.0 & 18.5 \\
\textsc{Commercial} & 45.2 & 6.5 & 4.5 & 56.2 \\
\midrule
\textsc{Total} & 87.5 & 7.7 & 4.8 &  \\
\midrule
\multicolumn{5}{c}{\textbf{\emph{Speech Datasets}}} \\
\midrule
\textsc{NC/Acad} & 9.5 & 9.5 & 13.7 & 32.6 \\
\textsc{Unspecified} & 6.3 & 0.0 & 7.4 & 13.7 \\
\textsc{Commercial} & 7.4 & 18.9 & 27.4 & 53.7 \\
\midrule
\textsc{Total} & 23.2 & 28.4 & 48.4 &  \\
\midrule
\multicolumn{5}{c}{\textbf{\emph{Video Datasets}}} \\
\midrule
\textsc{NC/Acad} & 22.1 & 0.0 & 9.6 & 31.7 \\
\textsc{Unspecified} & 23.1 & 1.0 & 11.5 & 35.6 \\
\textsc{Commercial} & 25.0 & 0.0 & 7.7 & 32.7 \\
\midrule
\textsc{Total} & 70.2 & 1.0 & 28.8 & \\
\bottomrule
\end{tabular}
\end{adjustbox}
\caption{\textbf{A breakdown of the percentage of license and terms restrictions} by dataset count. The much higher frequency of restrictions at the collection level is because we consider a collection's license or terms status to be the most restrictive of those for its datasets. Note that percentages may not add to exactly 100\% because of rounding.
}
\label{tab:license_terms_breakdown_count}
\vspace{-2mm}
\end{table*}

\section{Additional Results}
\label{app:results}
\begin{figure*}[!htb]
    \centering
    \includegraphics[width=\textwidth]{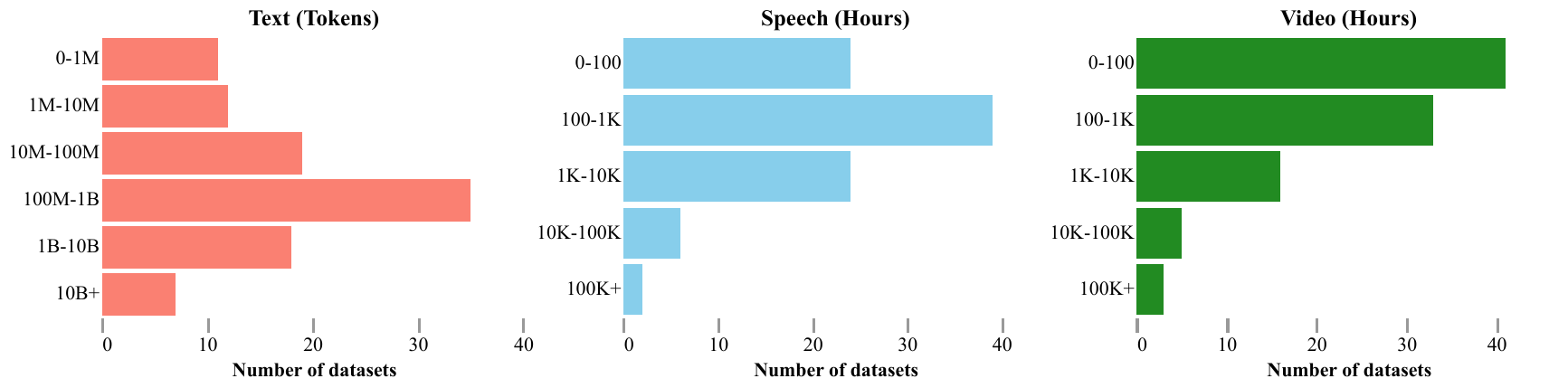}
    \caption{The distribution of dataset sizes for each modality. Most text data collections are between 100M-1B tokens. \textbf{Speech datasets average 100-1k hours, and video datasets are usually the smallest, commonly less than 100 hours.} }
    \label{fig:dimensions}
\end{figure*}

\begin{figure*}[!htb]
    \centering
    \includegraphics[width=\textwidth]{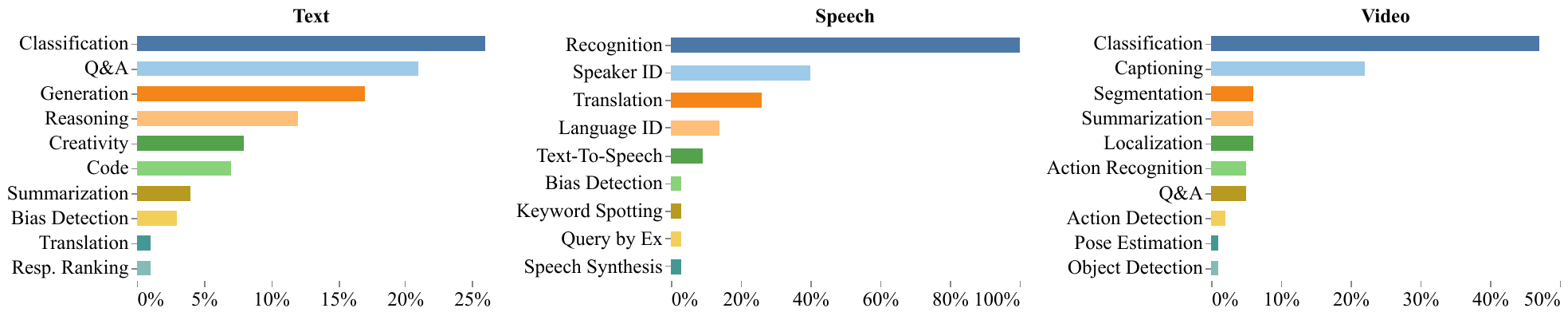}
    \caption{The task distribution of datasets, across modalities. Post-training text and video datasets are predominantly based on classification. For text, generation and reasoning are rising categories. All speech datasets are recognition-based, particularly for speaker, language, or in the process of translation.}
    \label{fig:tasks}
\end{figure*}

Figures \ref{fig:dimensions} and \ref{fig:tasks} report the size distributions of the datasets. We measure size differently for different types of datasets: Text datasets are in tokens, and audio/video in hours of content. The lack of standard tokenization or preprocessing schemes for those modalities makes it simplest to report raw dataset size.

Notably, we find quite different size distributions by modality. The distribution of dataset sizes has the thickest right tail for text, followed by speech and then by video. Most video datasets are short in hour terms, with speech datasets tending to be somewhat longer and text datasets having a greater prevalence of both very small and very large datasets relative to the mean size.

Dataset tasks, meanwhile, reflect traditional approaches and research programs for each modality. Classification is the most common task for both text and video, with the video community's long-standing interest in captioning also visible in its role as the second most common task for video datasets. Q\&A occupies a similar role for text, though text datasets have a more balanced distribution over other, increasingly prominent tasks like generation and reasoning. Given our selection criteria, all datasets for speech are for ASR tasks, but other tasks like speaker identification and translation are also represented.

\section{Datasets}
\label{app:datasets}
This section provides a detailed overview of the datasets we have collected and analyzed. \autoref{tab:collections-text} summarizes the text datasets, \autoref{tab:collections-speech} the audio datasets, and \autoref{tab:collections-video} the video datasets. Each of these tables lists broad collections of data, sorted in chronological order, and provides information about their properties, sizes, sources and permissions. Each collection can include multiple datasets, and they generally reflect the ways dataset creators have grouped their datasets (such as in the same paper). Because of the large number of datasets, we provide detailed information about their licenses and original published papers, where applicable, in the supplementary Attribution Card in \Cref{sec:attribution-card}.

\paragraph{Annotation Details: Text}
For post-training text datasets it is common to package many together as collections, such as Flan \citep{weifinetuned} or P3 \citep{sanh2021multitask}. 
This practice is not common to the same extent for speech or video datasets.
For much of the text analysis, where possible, we chose to analyze statistics at the collection-level, since practitioners are more likely to adopt a collection for general-purpose post-training, than an individual dataset within the collection.
Also, in dataset-level statistics, metadata for a single collection with many datasets can get repeated and overwhelm the statistics unfairly (e.g. the dataset aggregator/creator being repeated hundreds of times).
Consequently, our collection-level analysis of the text modality is reflected in  \Cref{fig:temporal-source-categories}, \Cref{fig:creator-worldmaps}, \Cref{fig:creator-orgs}, \Cref{fig:representation}, \Cref{fig:tasks}, and \Cref{fig:dimensions}.
However, for \Cref{fig:license-terms} we draw the distinction between collection and dataset metrics, as practitioners may wish to unpack collections to extract only commercially licensed data.
In that case a Collection inherits the most restrictive license and terms of its constituent datasets.

For annotating creator organizations, we follow prior work's instructions \citep{longpre2023data}.
For each dataset they record the affiliations listed on the academic paper or GitHub or HuggingFace object in which the dataset was released.
This does not include the organizations who created or owned the sources from which the data was derived.
For instance, the SQuAD dataset \citep{rajpurkar2016squad} would be associated with Stanford (the authors' affiliation), but not Wikipedia, which the data was partially derived from.
For a dataset that has authors affiliated with multiple organizations, the dataset will be counted towards each organization.

\paragraph{Annotation Details: Speech}
In many cases, multiple versions of a dataset exist due to datasets being expanded or updated. In these scenarios, we used the release date from the initial version (since release dates for subsequent versions were not always clear), but used metadata from the most recently released version for which information was available to offer an overview of the current landscape of data. However, if the dataset versions could not be meaningfully aggregated (e.g. different licenses), or did not appear to be cumulatively designed (non-overlapping or otherwise semantically disjoint data), we maintained separate records. We kept only datasets for which ASR was noted as a primary task. For example, if a dataset was primarily intended for text-to-speech or speaker recognition, we did not keep it even if it could conceivably be repurposed for ASR. When computing hours, we excluded any hours without supervisory transcripts/scripts (unlabeled data), but kept hours with ``weak supervision'' (e.g. model-generated transcripts from speech audio). 
We recognize the difficulty in comprehensively covering all relevant datasets. 

\paragraph{Annotation Details: Video} In video, a single dataset can be re-purposed and annotated to address different tasks \cite{monfort2019moments, monfortSpokenMomentsLearning2021}. We consider these as two different datasets even if they have the same video source since now they can be used for different computer vision tasks. 

\clearpage

\setlength{\tabcolsep}{1.9pt}
\definecolor{colorRiddleSensePropertyCounts_Datasets}{RGB}{240,223,178}
\definecolor{colorMathInstr.PropertyCounts_Datasets}{RGB}{240,223,178}
\definecolor{colorNoRobotsPropertyCounts_Datasets}{RGB}{240,223,178}
\definecolor{colorNectarPropertyCounts_Datasets}{RGB}{240,223,178}
\definecolor{colorMetaMathQAPropertyCounts_Datasets}{RGB}{245,238,218}
\definecolor{colorMegaWikaPropertyCounts_Datasets}{RGB}{240,243,243}
\definecolor{colorMedInstr.PropertyCounts_Datasets}{RGB}{240,223,178}
\definecolor{colorMathDialPropertyCounts_Datasets}{RGB}{240,223,178}
\definecolor{colorPII-Masking-200kPropertyCounts_Datasets}{RGB}{240,223,178}
\definecolor{colorPure-DovePropertyCounts_Datasets}{RGB}{240,223,178}
\definecolor{colorLMSYS-Chat-1MPropertyCounts_Datasets}{RGB}{240,223,178}
\definecolor{colorPygmalionAI-PIPPAPropertyCounts_Datasets}{RGB}{240,223,178}
\definecolor{colorHelpSteerPropertyCounts_Datasets}{RGB}{240,223,178}
\definecolor{colorSeaBenchPropertyCounts_Datasets}{RGB}{245,238,220}
\definecolor{colorOpenAsst.v2PropertyCounts_Datasets}{RGB}{245,241,232}
\definecolor{colorFeedbackColl.PropertyCounts_Datasets}{RGB}{240,223,178}
\definecolor{colorGlaiveCodeAsst.PropertyCounts_Datasets}{RGB}{240,223,178}
\definecolor{colorEverythingLMPropertyCounts_Datasets}{RGB}{240,223,178}
\definecolor{colorBactrian-XPropertyCounts_Datasets}{RGB}{245,236,212}
\definecolor{colorCOBRAFramesPropertyCounts_Datasets}{RGB}{240,223,178}
\definecolor{colorUltraFeedbackArgillaPropertyCounts_Datasets}{RGB}{245,238,220}
\definecolor{colorExpertQAPropertyCounts_Datasets}{RGB}{240,223,178}
\definecolor{colorChatDoctorPropertyCounts_Datasets}{RGB}{245,233,200}
\definecolor{colorCapybaraPropertyCounts_Datasets}{RGB}{245,239,224}
\definecolor{colorUltraChat-200kPropertyCounts_Datasets}{RGB}{240,223,178}
\definecolor{colorCollectiveCognitionPropertyCounts_Datasets}{RGB}{240,223,178}
\definecolor{colorThaiGenAIPropertyCounts_Datasets}{RGB}{245,238,220}
\definecolor{colorDeita10KPropertyCounts_Datasets}{RGB}{246,232,195}
\definecolor{colorSelFeePropertyCounts_Datasets}{RGB}{240,223,178}
\definecolor{colorChatbotArenaPropertyCounts_Datasets}{RGB}{240,223,178}
\definecolor{colorOpenGPTHealthcarePropertyCounts_Datasets}{RGB}{245,233,200}
\definecolor{colorOrca-MathPropertyCounts_Datasets}{RGB}{240,223,178}
\definecolor{colorOpenMathInstr.-1PropertyCounts_Datasets}{RGB}{246,232,195}
\definecolor{colorWildChatPropertyCounts_Datasets}{RGB}{246,232,195}
\definecolor{colorMagpie-ProPropertyCounts_Datasets}{RGB}{240,223,178}
\definecolor{color10kPromptRankedPropertyCounts_Datasets}{RGB}{240,223,178}
\definecolor{colorSynth.-GSM8K-Refl.PropertyCounts_Datasets}{RGB}{240,223,178}
\definecolor{colorLongAlign-10kPropertyCounts_Datasets}{RGB}{240,223,178}
\definecolor{colorLlama2-MedTuned-Instr.PropertyCounts_Datasets}{RGB}{240,223,178}
\definecolor{colorKIWIPropertyCounts_Datasets}{RGB}{240,223,178}
\definecolor{colorIndic-Instr.PropertyCounts_Datasets}{RGB}{245,238,218}
\definecolor{colorGretelText-to-SQLPropertyCounts_Datasets}{RGB}{240,223,178}
\definecolor{colorConiferPropertyCounts_Datasets}{RGB}{240,223,178}
\definecolor{colorCidarPropertyCounts_Datasets}{RGB}{240,223,178}
\definecolor{colorAyaPropertyCounts_Datasets}{RGB}{235,242,241}
\definecolor{colorAgentInstructPropertyCounts_Datasets}{RGB}{245,236,212}
\definecolor{colorInstArPropertyCounts_Datasets}{RGB}{245,243,238}
\definecolor{colorDynosaurPropertyCounts_Datasets}{RGB}{179,226,219}
\definecolor{colorMedicalMeadowPropertyCounts_Datasets}{RGB}{245,238,218}
\definecolor{colorOpen-PlatypusPropertyCounts_Datasets}{RGB}{245,239,222}
\definecolor{colorPMC-LLaMAInstr.PropertyCounts_Datasets}{RGB}{245,237,216}
\definecolor{colorCOIGPropertyCounts_Datasets}{RGB}{245,241,232}
\definecolor{colorDialogStudioPropertyCounts_Datasets}{RGB}{231,241,240}
\definecolor{colorReasoningPropertyCounts_Datasets}{RGB}{240,223,178}
\definecolor{colorRiddleSensePropertyCounts_Tasks}{RGB}{245,240,228}
\definecolor{colorMathInstr.PropertyCounts_Tasks}{RGB}{245,240,228}
\definecolor{colorNoRobotsPropertyCounts_Tasks}{RGB}{224,240,237}
\definecolor{colorNectarPropertyCounts_Tasks}{RGB}{240,223,178}
\definecolor{colorMetaMathQAPropertyCounts_Tasks}{RGB}{245,236,210}
\definecolor{colorMegaWikaPropertyCounts_Tasks}{RGB}{240,223,178}
\definecolor{colorMedInstr.PropertyCounts_Tasks}{RGB}{240,223,178}
\definecolor{colorMathDialPropertyCounts_Tasks}{RGB}{245,236,210}
\definecolor{colorPII-Masking-200kPropertyCounts_Tasks}{RGB}{245,236,210}
\definecolor{colorPure-DovePropertyCounts_Tasks}{RGB}{245,243,240}
\definecolor{colorLMSYS-Chat-1MPropertyCounts_Tasks}{RGB}{218,238,235}
\definecolor{colorPygmalionAI-PIPPAPropertyCounts_Tasks}{RGB}{245,240,228}
\definecolor{colorHelpSteerPropertyCounts_Tasks}{RGB}{242,244,244}
\definecolor{colorSeaBenchPropertyCounts_Tasks}{RGB}{245,243,240}
\definecolor{colorOpenAsst.v2PropertyCounts_Tasks}{RGB}{245,243,240}
\definecolor{colorFeedbackColl.PropertyCounts_Tasks}{RGB}{245,236,210}
\definecolor{colorGlaiveCodeAsst.PropertyCounts_Tasks}{RGB}{245,236,210}
\definecolor{colorEverythingLMPropertyCounts_Tasks}{RGB}{224,240,237}
\definecolor{colorBactrian-XPropertyCounts_Tasks}{RGB}{245,243,240}
\definecolor{colorCOBRAFramesPropertyCounts_Tasks}{RGB}{240,223,178}
\definecolor{colorUltraFeedbackArgillaPropertyCounts_Tasks}{RGB}{196,232,227}
\definecolor{colorExpertQAPropertyCounts_Tasks}{RGB}{245,240,228}
\definecolor{colorChatDoctorPropertyCounts_Tasks}{RGB}{240,223,178}
\definecolor{colorCapybaraPropertyCounts_Tasks}{RGB}{193,231,226}
\definecolor{colorUltraChat-200kPropertyCounts_Tasks}{RGB}{229,241,239}
\definecolor{colorCollectiveCognitionPropertyCounts_Tasks}{RGB}{235,242,241}
\definecolor{colorThaiGenAIPropertyCounts_Tasks}{RGB}{211,237,233}
\definecolor{colorDeita10KPropertyCounts_Tasks}{RGB}{211,237,233}
\definecolor{colorSelFeePropertyCounts_Tasks}{RGB}{242,244,244}
\definecolor{colorChatbotArenaPropertyCounts_Tasks}{RGB}{245,243,240}
\definecolor{colorOpenGPTHealthcarePropertyCounts_Tasks}{RGB}{245,243,240}
\definecolor{colorOrca-MathPropertyCounts_Tasks}{RGB}{240,223,178}
\definecolor{colorOpenMathInstr.-1PropertyCounts_Tasks}{RGB}{245,240,228}
\definecolor{colorWildChatPropertyCounts_Tasks}{RGB}{229,241,239}
\definecolor{colorMagpie-ProPropertyCounts_Tasks}{RGB}{218,238,235}
\definecolor{color10kPromptRankedPropertyCounts_Tasks}{RGB}{206,235,231}
\definecolor{colorSynth.-GSM8K-Refl.PropertyCounts_Tasks}{RGB}{245,240,228}
\definecolor{colorLongAlign-10kPropertyCounts_Tasks}{RGB}{245,240,228}
\definecolor{colorLlama2-MedTuned-Instr.PropertyCounts_Tasks}{RGB}{245,243,240}
\definecolor{colorKIWIPropertyCounts_Tasks}{RGB}{240,223,178}
\definecolor{colorIndic-Instr.PropertyCounts_Tasks}{RGB}{229,241,239}
\definecolor{colorGretelText-to-SQLPropertyCounts_Tasks}{RGB}{240,223,178}
\definecolor{colorConiferPropertyCounts_Tasks}{RGB}{224,240,237}
\definecolor{colorCidarPropertyCounts_Tasks}{RGB}{224,240,237}
\definecolor{colorAyaPropertyCounts_Tasks}{RGB}{229,241,239}
\definecolor{colorAgentInstructPropertyCounts_Tasks}{RGB}{245,240,228}
\definecolor{colorInstArPropertyCounts_Tasks}{RGB}{206,235,231}
\definecolor{colorDynosaurPropertyCounts_Tasks}{RGB}{179,226,219}
\definecolor{colorMedicalMeadowPropertyCounts_Tasks}{RGB}{245,236,210}
\definecolor{colorOpen-PlatypusPropertyCounts_Tasks}{RGB}{215,237,234}
\definecolor{colorPMC-LLaMAInstr.PropertyCounts_Tasks}{RGB}{240,223,178}
\definecolor{colorCOIGPropertyCounts_Tasks}{RGB}{206,235,231}
\definecolor{colorDialogStudioPropertyCounts_Tasks}{RGB}{245,240,228}
\definecolor{colorReasoningPropertyCounts_Tasks}{RGB}{245,243,240}
\definecolor{colorRiddleSensePropertyCounts_Langs}{RGB}{240,223,178}
\definecolor{colorMathInstr.PropertyCounts_Langs}{RGB}{240,223,178}
\definecolor{colorNoRobotsPropertyCounts_Langs}{RGB}{240,223,178}
\definecolor{colorNectarPropertyCounts_Langs}{RGB}{240,223,178}
\definecolor{colorMetaMathQAPropertyCounts_Langs}{RGB}{240,223,178}
\definecolor{colorMegaWikaPropertyCounts_Langs}{RGB}{196,232,227}
\definecolor{colorMedInstr.PropertyCounts_Langs}{RGB}{240,223,178}
\definecolor{colorMathDialPropertyCounts_Langs}{RGB}{240,223,178}
\definecolor{colorPII-Masking-200kPropertyCounts_Langs}{RGB}{245,239,222}
\definecolor{colorPure-DovePropertyCounts_Langs}{RGB}{240,223,178}
\definecolor{colorLMSYS-Chat-1MPropertyCounts_Langs}{RGB}{245,241,230}
\definecolor{colorPygmalionAI-PIPPAPropertyCounts_Langs}{RGB}{240,223,178}
\definecolor{colorHelpSteerPropertyCounts_Langs}{RGB}{240,223,178}
\definecolor{colorSeaBenchPropertyCounts_Langs}{RGB}{244,244,244}
\definecolor{colorOpenAsst.v2PropertyCounts_Langs}{RGB}{222,239,237}
\definecolor{colorFeedbackColl.PropertyCounts_Langs}{RGB}{240,223,178}
\definecolor{colorGlaiveCodeAsst.PropertyCounts_Langs}{RGB}{245,234,202}
\definecolor{colorEverythingLMPropertyCounts_Langs}{RGB}{245,234,202}
\definecolor{colorBactrian-XPropertyCounts_Langs}{RGB}{245,242,234}
\definecolor{colorCOBRAFramesPropertyCounts_Langs}{RGB}{240,223,178}
\definecolor{colorUltraFeedbackArgillaPropertyCounts_Langs}{RGB}{240,223,178}
\definecolor{colorExpertQAPropertyCounts_Langs}{RGB}{240,223,178}
\definecolor{colorChatDoctorPropertyCounts_Langs}{RGB}{240,223,178}
\definecolor{colorCapybaraPropertyCounts_Langs}{RGB}{245,234,202}
\definecolor{colorUltraChat-200kPropertyCounts_Langs}{RGB}{240,223,178}
\definecolor{colorCollectiveCognitionPropertyCounts_Langs}{RGB}{240,223,178}
\definecolor{colorThaiGenAIPropertyCounts_Langs}{RGB}{240,223,178}
\definecolor{colorDeita10KPropertyCounts_Langs}{RGB}{240,223,178}
\definecolor{colorSelFeePropertyCounts_Langs}{RGB}{240,223,178}
\definecolor{colorChatbotArenaPropertyCounts_Langs}{RGB}{240,223,178}
\definecolor{colorOpenGPTHealthcarePropertyCounts_Langs}{RGB}{240,223,178}
\definecolor{colorOrca-MathPropertyCounts_Langs}{RGB}{240,223,178}
\definecolor{colorOpenMathInstr.-1PropertyCounts_Langs}{RGB}{240,223,178}
\definecolor{colorWildChatPropertyCounts_Langs}{RGB}{240,243,243}
\definecolor{colorMagpie-ProPropertyCounts_Langs}{RGB}{240,223,178}
\definecolor{color10kPromptRankedPropertyCounts_Langs}{RGB}{240,223,178}
\definecolor{colorSynth.-GSM8K-Refl.PropertyCounts_Langs}{RGB}{240,223,178}
\definecolor{colorLongAlign-10kPropertyCounts_Langs}{RGB}{240,223,178}
\definecolor{colorLlama2-MedTuned-Instr.PropertyCounts_Langs}{RGB}{240,223,178}
\definecolor{colorKIWIPropertyCounts_Langs}{RGB}{240,223,178}
\definecolor{colorIndic-Instr.PropertyCounts_Langs}{RGB}{245,234,202}
\definecolor{colorGretelText-to-SQLPropertyCounts_Langs}{RGB}{245,237,214}
\definecolor{colorConiferPropertyCounts_Langs}{RGB}{240,223,178}
\definecolor{colorCidarPropertyCounts_Langs}{RGB}{240,223,178}
\definecolor{colorAyaPropertyCounts_Langs}{RGB}{179,226,219}
\definecolor{colorAgentInstructPropertyCounts_Langs}{RGB}{240,223,178}
\definecolor{colorInstArPropertyCounts_Langs}{RGB}{240,223,178}
\definecolor{colorDynosaurPropertyCounts_Langs}{RGB}{240,223,178}
\definecolor{colorMedicalMeadowPropertyCounts_Langs}{RGB}{240,223,178}
\definecolor{colorOpen-PlatypusPropertyCounts_Langs}{RGB}{206,235,231}
\definecolor{colorPMC-LLaMAInstr.PropertyCounts_Langs}{RGB}{240,223,178}
\definecolor{colorCOIGPropertyCounts_Langs}{RGB}{245,234,202}
\definecolor{colorDialogStudioPropertyCounts_Langs}{RGB}{245,241,230}
\definecolor{colorReasoningPropertyCounts_Langs}{RGB}{240,223,178}
\definecolor{colorRiddleSensePropertyCounts_Domains}{RGB}{240,223,178}
\definecolor{colorMathInstr.PropertyCounts_Domains}{RGB}{240,223,178}
\definecolor{colorNoRobotsPropertyCounts_Domains}{RGB}{240,223,178}
\definecolor{colorNectarPropertyCounts_Domains}{RGB}{245,236,210}
\definecolor{colorMetaMathQAPropertyCounts_Domains}{RGB}{240,223,178}
\definecolor{colorMegaWikaPropertyCounts_Domains}{RGB}{240,223,178}
\definecolor{colorMedInstr.PropertyCounts_Domains}{RGB}{240,223,178}
\definecolor{colorMathDialPropertyCounts_Domains}{RGB}{245,243,238}
\definecolor{colorPII-Masking-200kPropertyCounts_Domains}{RGB}{240,223,178}
\definecolor{colorPure-DovePropertyCounts_Domains}{RGB}{240,223,178}
\definecolor{colorLMSYS-Chat-1MPropertyCounts_Domains}{RGB}{240,223,178}
\definecolor{colorPygmalionAI-PIPPAPropertyCounts_Domains}{RGB}{240,223,178}
\definecolor{colorHelpSteerPropertyCounts_Domains}{RGB}{240,223,178}
\definecolor{colorSeaBenchPropertyCounts_Domains}{RGB}{242,244,244}
\definecolor{colorOpenAsst.v2PropertyCounts_Domains}{RGB}{240,223,178}
\definecolor{colorFeedbackColl.PropertyCounts_Domains}{RGB}{240,223,178}
\definecolor{colorGlaiveCodeAsst.PropertyCounts_Domains}{RGB}{240,223,178}
\definecolor{colorEverythingLMPropertyCounts_Domains}{RGB}{240,223,178}
\definecolor{colorBactrian-XPropertyCounts_Domains}{RGB}{240,223,178}
\definecolor{colorCOBRAFramesPropertyCounts_Domains}{RGB}{245,236,210}
\definecolor{colorUltraFeedbackArgillaPropertyCounts_Domains}{RGB}{185,228,221}
\definecolor{colorExpertQAPropertyCounts_Domains}{RGB}{240,223,178}
\definecolor{colorChatDoctorPropertyCounts_Domains}{RGB}{245,236,210}
\definecolor{colorCapybaraPropertyCounts_Domains}{RGB}{240,223,178}
\definecolor{colorUltraChat-200kPropertyCounts_Domains}{RGB}{245,236,210}
\definecolor{colorCollectiveCognitionPropertyCounts_Domains}{RGB}{240,223,178}
\definecolor{colorThaiGenAIPropertyCounts_Domains}{RGB}{240,223,178}
\definecolor{colorDeita10KPropertyCounts_Domains}{RGB}{245,240,226}
\definecolor{colorSelFeePropertyCounts_Domains}{RGB}{240,223,178}
\definecolor{colorChatbotArenaPropertyCounts_Domains}{RGB}{240,223,178}
\definecolor{colorOpenGPTHealthcarePropertyCounts_Domains}{RGB}{240,223,178}
\definecolor{colorOrca-MathPropertyCounts_Domains}{RGB}{245,240,226}
\definecolor{colorOpenMathInstr.-1PropertyCounts_Domains}{RGB}{245,240,226}
\definecolor{colorWildChatPropertyCounts_Domains}{RGB}{240,223,178}
\definecolor{colorMagpie-ProPropertyCounts_Domains}{RGB}{240,223,178}
\definecolor{color10kPromptRankedPropertyCounts_Domains}{RGB}{245,243,238}
\definecolor{colorSynth.-GSM8K-Refl.PropertyCounts_Domains}{RGB}{240,223,178}
\definecolor{colorLongAlign-10kPropertyCounts_Domains}{RGB}{240,223,178}
\definecolor{colorLlama2-MedTuned-Instr.PropertyCounts_Domains}{RGB}{240,223,178}
\definecolor{colorKIWIPropertyCounts_Domains}{RGB}{245,236,210}
\definecolor{colorIndic-Instr.PropertyCounts_Domains}{RGB}{245,240,226}
\definecolor{colorGretelText-to-SQLPropertyCounts_Domains}{RGB}{240,223,178}
\definecolor{colorConiferPropertyCounts_Domains}{RGB}{245,236,210}
\definecolor{colorCidarPropertyCounts_Domains}{RGB}{240,223,178}
\definecolor{colorAyaPropertyCounts_Domains}{RGB}{240,223,178}
\definecolor{colorAgentInstructPropertyCounts_Domains}{RGB}{229,241,239}
\definecolor{colorInstArPropertyCounts_Domains}{RGB}{220,239,236}
\definecolor{colorDynosaurPropertyCounts_Domains}{RGB}{179,226,219}
\definecolor{colorMedicalMeadowPropertyCounts_Domains}{RGB}{245,240,226}
\definecolor{colorOpen-PlatypusPropertyCounts_Domains}{RGB}{224,240,237}
\definecolor{colorPMC-LLaMAInstr.PropertyCounts_Domains}{RGB}{245,236,210}
\definecolor{colorCOIGPropertyCounts_Domains}{RGB}{179,226,219}
\definecolor{colorDialogStudioPropertyCounts_Domains}{RGB}{245,240,226}
\definecolor{colorReasoningPropertyCounts_Domains}{RGB}{240,223,178}

\begin{longtable}{lc|cccc|c|cc}
\caption[\textbf{Alignment tuning (text) collections and properties}]{\textbf{Alignment tuning (text) collections and properties}. Collection properties include numbers of datasets, tasks, languages, and text domains. The \textsc{Source} column indicates whether a collection contains human-generated web text (\emojiglobe), language model outputs (\emojirobot) or both (\emojiglobe\emojirobot). The \textsc{Use} column indicates whether a collection includes data freely usable even for commercial purposes (\protect\CommercialDataCircle), data usable only for noncommercial purposes or academic research (\protect\NCDataCircle) and data whose license status is not specified precisely enough to allow us to determine commercial use permissions (\protect\UnspecifiedDataCircle). Note that each collection may have different datasets with one, two, or all three of these statuses. Finally, the \textsc{OAI} column indicates collections which include OpenAI model generations. Datasets are sorted chronologically to highlight trends over time.} \label{tab:collections-text} \\
\toprule
\textsc{Collection} & \textsc{} & \multicolumn{4}{c}{\textsc{Property Counts}} & \textsc{Types} & \multicolumn{2}{c}{\textsc{Permissions}} \\
 & \textsc{\thead{Year}} & \textsc{\thead{Datasets}} & \textsc{\thead{Tasks}} & \textsc{\thead{Langs}} & \textsc{\thead{Domains}} & \textsc{\thead{Source}} & \textsc{\thead{Use}} & \textsc{\thead{OAI}} \\
\midrule
\endfirsthead
\caption[]{\textbf{Alignment tuning (text) collections and properties}.} \\
\toprule
\textsc{Collection} & \textsc{} & \multicolumn{4}{c}{\textsc{Property Counts}} & \textsc{Types} & \multicolumn{2}{c}{\textsc{Permissions}} \\
 & \textsc{\thead{Year}} & \textsc{\thead{Datasets}} & \textsc{\thead{Tasks}} & \textsc{\thead{Langs}} & \textsc{\thead{Domains}} & \textsc{\thead{Source}} & \textsc{\thead{Use}} & \textsc{\thead{OAI}} \\
\midrule
\endhead
\midrule
\multicolumn{9}{r}{Continued on next page} \\
\midrule
\endfoot
\bottomrule
\endlastfoot
RiddleSense & 2021 & \cellcolor{colorReasoningPropertyCounts_Datasets}{1} & \cellcolor{colorDialogStudioPropertyCounts_Tasks}{3} & \cellcolor{colorReasoningPropertyCounts_Langs}{1} & \cellcolor{colorReasoningPropertyCounts_Domains}{1} & \emojiglobe\emojiblank & \CommercialDataCircle \TransparentCircle \TransparentCircle & \emojiblank \\
MathInstr. & 2023 & \cellcolor{colorReasoningPropertyCounts_Datasets}{1} & \cellcolor{colorDialogStudioPropertyCounts_Tasks}{3} & \cellcolor{colorReasoningPropertyCounts_Langs}{1} & \cellcolor{colorReasoningPropertyCounts_Domains}{1} & \emojiblank\emojirobot & \CommercialDataCircle \TransparentCircle \TransparentCircle & \greencheck \\
No Robots & 2023 & \cellcolor{colorReasoningPropertyCounts_Datasets}{1} & \cellcolor{colorCidarPropertyCounts_Tasks}{8} & \cellcolor{colorReasoningPropertyCounts_Langs}{1} & \cellcolor{colorReasoningPropertyCounts_Domains}{1} & \emojiglobe\emojiblank & \TransparentCircle \TransparentCircle \NCDataCircle & \emojiblank \\
Nectar & 2023 & \cellcolor{colorReasoningPropertyCounts_Datasets}{1} & \cellcolor{colorPMC-LLaMAInstr.PropertyCounts_Tasks}{1} & \cellcolor{colorReasoningPropertyCounts_Langs}{1} & \cellcolor{colorPMC-LLaMAInstr.PropertyCounts_Domains}{2} & \emojiblank\emojirobot & \CommercialDataCircle \TransparentCircle \NCDataCircle & \greencheck \\
MetaMathQA & 2023 & \cellcolor{colorMedicalMeadowPropertyCounts_Datasets}{8} & \cellcolor{colorMedicalMeadowPropertyCounts_Tasks}{2} & \cellcolor{colorReasoningPropertyCounts_Langs}{1} & \cellcolor{colorReasoningPropertyCounts_Domains}{1} & \emojiblank\emojirobot & \CommercialDataCircle \TransparentCircle \TransparentCircle & \greencheck \\
\end{longtable}
\setlength{\tabcolsep}{1.9pt}
\definecolor{colorTIMITPropertyCounts_Hr}{RGB}{244,229,189}
\definecolor{colorSwitchboardPropertyCounts_Hr}{RGB}{245,241,232}
\definecolor{colorAfricanAcc.FrenchPropertyCounts_Hr}{RGB}{245,235,206}
\definecolor{colorCSJPropertyCounts_Hr}{RGB}{245,244,242}
\definecolor{colorFisherPropertyCounts_Hr}{RGB}{236,243,242}
\definecolor{colorCSLU22Langs.PropertyCounts_Hr}{RGB}{245,238,220}
\definecolor{colorAMIPropertyCounts_Hr}{RGB}{245,239,222}
\definecolor{colorCSLU1.2PropertyCounts_Hr}{RGB}{245,235,208}
\definecolor{colorALLSSTARPropertyCounts_Hr}{RGB}{245,238,220}
\definecolor{colorTED-LIUM3PropertyCounts_Hr}{RGB}{245,243,238}
\definecolor{colorNSTNorwegianPropertyCounts_Hr}{RGB}{245,243,240}
\definecolor{colorNSTDanishPropertyCounts_Hr}{RGB}{245,243,240}
\definecolor{colorNSTSwedishPropertyCounts_Hr}{RGB}{245,242,234}
\definecolor{colorVystadialPropertyCounts_Hr}{RGB}{245,237,216}
\definecolor{colorTHCHS-30PropertyCounts_Hr}{RGB}{245,236,212}
\definecolor{colorLibriSpeechPropertyCounts_Hr}{RGB}{244,244,244}
\definecolor{colorTHUYG-20PropertyCounts_Hr}{RGB}{245,235,206}
\definecolor{colorVCTKPropertyCounts_Hr}{RGB}{245,237,214}
\definecolor{colorSpokenWikipediaPropertyCounts_Hr}{RGB}{244,244,244}
\definecolor{colorAISHELL-1PropertyCounts_Hr}{RGB}{245,243,240}
\definecolor{colorLJSpeechPropertyCounts_Hr}{RGB}{245,235,208}
\definecolor{colorClarinPLPropertyCounts_Hr}{RGB}{245,237,216}
\definecolor{colorAISHELL-2PropertyCounts_Hr}{RGB}{244,244,244}
\definecolor{colorRegionalAf.Am.Lang.PropertyCounts_Hr}{RGB}{245,240,228}
\definecolor{colorCrowdSourcedSpeechPropertyCounts_Hr}{RGB}{242,244,244}
\definecolor{colorZeroth-KoreanPropertyCounts_Hr}{RGB}{245,239,222}
\definecolor{colorRTVEPropertyCounts_Hr}{RGB}{245,244,242}
\definecolor{colorOpenSTTPropertyCounts_Hr}{RGB}{215,237,234}
\definecolor{colorMuST-CPropertyCounts_Hr}{RGB}{231,241,240}
\definecolor{colorM-AILABSPropertyCounts_Hr}{RGB}{244,244,244}
\definecolor{colorMAGICDATAPropertyCounts_Hr}{RGB}{245,244,244}
\definecolor{colorCommonVoice17PropertyCounts_Hr}{RGB}{211,237,233}
\definecolor{colorCoNASEPropertyCounts_Hr}{RGB}{193,231,226}
\definecolor{colorNigerianEnglishPropertyCounts_Hr}{RGB}{245,230,192}
\definecolor{colorNorwegianParl.SpeechPropertyCounts_Hr}{RGB}{245,240,226}
\definecolor{color120hSpanishSpeechPropertyCounts_Hr}{RGB}{245,239,224}
\definecolor{colorDiDiSpeechPropertyCounts_Hr}{RGB}{245,244,244}
\definecolor{colorCzechParliamentPropertyCounts_Hr}{RGB}{245,243,238}
\definecolor{colorCoVoST-2PropertyCounts_Hr}{RGB}{233,242,240}
\definecolor{colorKSCPropertyCounts_Hr}{RGB}{245,242,234}
\definecolor{colorBasq.,Cat.andGal.PropertyCounts_Hr}{RGB}{245,236,210}
\definecolor{colorKsponSpeechPropertyCounts_Hr}{RGB}{244,244,244}
\definecolor{colorSamromurPropertyCounts_Hr}{RGB}{245,240,226}
\definecolor{colorMultiling.LibriSpeechPropertyCounts_Hr}{RGB}{206,235,231}
\definecolor{colorMaSSPropertyCounts_Hr}{RGB}{245,240,228}
\definecolor{colorFTSPEECHPropertyCounts_Hr}{RGB}{236,243,242}
\definecolor{colorEng.Acc.inBrit.IslesPropertyCounts_Hr}{RGB}{245,236,210}
\definecolor{colorHighlandPueblaNahuatlPropertyCounts_Hr}{RGB}{245,240,226}
\definecolor{colorQASRPropertyCounts_Hr}{RGB}{236,243,242}
\definecolor{colorMultiling.TEDxPropertyCounts_Hr}{RGB}{245,244,244}
\definecolor{colorMinds14PropertyCounts_Hr}{RGB}{245,235,208}
\definecolor{colorGolosPropertyCounts_Hr}{RGB}{242,244,244}
\definecolor{colorMASCPropertyCounts_Hr}{RGB}{242,244,244}
\definecolor{colorLaboroTVSpeechPropertyCounts_Hr}{RGB}{236,243,242}
\definecolor{colorKeSpeechPropertyCounts_Hr}{RGB}{238,243,242}
\definecolor{colorJTUBESPEECHPropertyCounts_Hr}{RGB}{240,243,243}
\definecolor{colorGigaSpeechPropertyCounts_Hr}{RGB}{222,239,237}
\definecolor{colorVoxPopuliPropertyCounts_Hr}{RGB}{238,243,242}
\definecolor{colorSPGISpeechPropertyCounts_Hr}{RGB}{227,240,239}
\definecolor{colorWestAfr.RadioPropertyCounts_Hr}{RGB}{245,240,226}
\definecolor{colorAISHELL-4PropertyCounts_Hr}{RGB}{245,239,224}
\definecolor{colorWestAfr.Virt.Asst.PropertyCounts_Hr}{RGB}{239,221,175}
\definecolor{colorMediaSpeechPropertyCounts_Hr}{RGB}{245,236,212}
\definecolor{colorPeople'sSpeechPropertyCounts_Hr}{RGB}{211,237,233}
\definecolor{color1111HoursHindiPropertyCounts_Hr}{RGB}{245,239,224}
\definecolor{colorShrutilipiPropertyCounts_Hr}{RGB}{226,240,238}
\definecolor{colorWenetSpeechPropertyCounts_Hr}{RGB}{222,239,237}
\definecolor{colorSamromurChildrenPropertyCounts_Hr}{RGB}{245,240,226}
\definecolor{colorSDS-200PropertyCounts_Hr}{RGB}{245,241,230}
\definecolor{coloraidatatangPropertyCounts_Hr}{RGB}{245,241,230}
\definecolor{colorFleursPropertyCounts_Hr}{RGB}{240,243,243}
\definecolor{colorOLKAVSPropertyCounts_Hr}{RGB}{242,244,244}
\definecolor{colorNorwegianParl.PropertyCounts_Hr}{RGB}{245,240,226}
\definecolor{colorMagicData-RAMCPropertyCounts_Hr}{RGB}{245,240,228}
\definecolor{colorKathbathPropertyCounts_Hr}{RGB}{238,243,242}
\definecolor{colorHebrewKanPropertyCounts_Hr}{RGB}{245,232,196}
\definecolor{colorHebrewCourseraPropertyCounts_Hr}{RGB}{245,236,212}
\definecolor{colorBloomSpeechPropertyCounts_Hr}{RGB}{245,243,238}
\definecolor{colorEnglish-VietnamesePropertyCounts_Hr}{RGB}{245,243,240}
\definecolor{colorEarnings-22PropertyCounts_Hr}{RGB}{245,239,224}
\definecolor{colorYODASPropertyCounts_Hr}{RGB}{179,226,219}
\definecolor{colorAFRISPEECH-200PropertyCounts_Hr}{RGB}{245,241,230}
\definecolor{colorAaltoFinnishParl.PropertyCounts_Hr}{RGB}{233,242,240}
\definecolor{colorReazonSpeechPropertyCounts_Hr}{RGB}{209,236,232}
\definecolor{colorEdAccPropertyCounts_Hr}{RGB}{245,236,212}
\definecolor{colorRixVoxPropertyCounts_Hr}{RGB}{227,240,239}
\definecolor{colorJapaneseAnimeSpeechPropertyCounts_Hr}{RGB}{245,239,224}
\definecolor{colorSnowMountainPropertyCounts_Hr}{RGB}{245,241,232}
\definecolor{colorSamromurMilljonPropertyCounts_Hr}{RGB}{244,244,244}
\definecolor{colorBud500PropertyCounts_Hr}{RGB}{245,243,240}
\definecolor{colorVibraVoxPropertyCounts_Hr}{RGB}{245,234,204}
\definecolor{colorM2ASRPropertyCounts_Hr}{RGB}{245,243,238}
\definecolor{colorTIMITPropertyCounts_Spkr}{RGB}{215,237,234}
\definecolor{colorSwitchboardPropertyCounts_Spkr}{RGB}{215,237,234}
\definecolor{colorAfricanAcc.FrenchPropertyCounts_Spkr}{RGB}{218,238,235}
\definecolor{colorCSJPropertyCounts_Spkr}{RGB}{211,237,233}
\definecolor{colorFisherPropertyCounts_Spkr}{RGB}{202,234,230}
\definecolor{colorCSLU22Langs.PropertyCounts_Spkr}{RGB}{240,223,178}
\definecolor{colorAMIPropertyCounts_Spkr}{RGB}{240,223,178}
\definecolor{colorCSLU1.2PropertyCounts_Spkr}{RGB}{206,235,231}
\definecolor{colorALLSSTARPropertyCounts_Spkr}{RGB}{220,239,236}
\definecolor{colorTED-LIUM3PropertyCounts_Spkr}{RGB}{209,236,232}
\definecolor{colorNSTNorwegianPropertyCounts_Spkr}{RGB}{213,237,234}
\definecolor{colorNSTDanishPropertyCounts_Spkr}{RGB}{240,223,178}
\definecolor{colorNSTSwedishPropertyCounts_Spkr}{RGB}{240,223,178}
\definecolor{colorVystadialPropertyCounts_Spkr}{RGB}{240,223,178}
\definecolor{colorTHCHS-30PropertyCounts_Spkr}{RGB}{226,240,238}
\definecolor{colorLibriSpeechPropertyCounts_Spkr}{RGB}{208,236,232}
\definecolor{colorTHUYG-20PropertyCounts_Spkr}{RGB}{217,238,235}
\definecolor{colorVCTKPropertyCounts_Spkr}{RGB}{222,239,237}
\definecolor{colorSpokenWikipediaPropertyCounts_Spkr}{RGB}{213,237,234}
\definecolor{colorAISHELL-1PropertyCounts_Spkr}{RGB}{217,238,235}
\definecolor{colorLJSpeechPropertyCounts_Spkr}{RGB}{242,244,244}
\definecolor{colorClarinPLPropertyCounts_Spkr}{RGB}{217,238,235}
\definecolor{colorAISHELL-2PropertyCounts_Spkr}{RGB}{209,236,232}
\definecolor{colorRegionalAf.Am.Lang.PropertyCounts_Spkr}{RGB}{218,238,235}
\definecolor{colorCrowdSourcedSpeechPropertyCounts_Spkr}{RGB}{208,236,232}
\definecolor{colorZeroth-KoreanPropertyCounts_Spkr}{RGB}{220,239,236}
\definecolor{colorRTVEPropertyCounts_Spkr}{RGB}{240,223,178}
\definecolor{colorOpenSTTPropertyCounts_Spkr}{RGB}{240,223,178}
\definecolor{colorMuST-CPropertyCounts_Spkr}{RGB}{209,236,232}
\definecolor{colorM-AILABSPropertyCounts_Spkr}{RGB}{240,223,178}
\definecolor{colorMAGICDATAPropertyCounts_Spkr}{RGB}{211,237,233}
\definecolor{colorCommonVoice17PropertyCounts_Spkr}{RGB}{179,226,219}
\definecolor{colorCoNASEPropertyCounts_Spkr}{RGB}{240,223,178}
\definecolor{colorNigerianEnglishPropertyCounts_Spkr}{RGB}{240,223,178}
\definecolor{colorNorwegianParl.SpeechPropertyCounts_Spkr}{RGB}{217,238,235}
\definecolor{color120hSpanishSpeechPropertyCounts_Spkr}{RGB}{229,241,239}
\definecolor{colorDiDiSpeechPropertyCounts_Spkr}{RGB}{204,235,230}
\definecolor{colorCzechParliamentPropertyCounts_Spkr}{RGB}{218,238,235}
\definecolor{colorCoVoST-2PropertyCounts_Spkr}{RGB}{190,230,224}
\definecolor{colorKSCPropertyCounts_Spkr}{RGB}{240,223,178}
\definecolor{colorBasq.,Cat.andGal.PropertyCounts_Spkr}{RGB}{220,239,236}
\definecolor{colorKsponSpeechPropertyCounts_Spkr}{RGB}{209,236,232}
\definecolor{colorSamromurPropertyCounts_Spkr}{RGB}{202,234,230}
\definecolor{colorMultiling.LibriSpeechPropertyCounts_Spkr}{RGB}{204,235,230}
\definecolor{colorMaSSPropertyCounts_Spkr}{RGB}{240,223,178}
\definecolor{colorFTSPEECHPropertyCounts_Spkr}{RGB}{217,238,235}
\definecolor{colorEng.Acc.inBrit.IslesPropertyCounts_Spkr}{RGB}{222,239,237}
\definecolor{colorHighlandPueblaNahuatlPropertyCounts_Spkr}{RGB}{240,223,178}
\definecolor{colorQASRPropertyCounts_Spkr}{RGB}{202,234,230}
\definecolor{colorMultiling.TEDxPropertyCounts_Spkr}{RGB}{240,223,178}
\definecolor{colorMinds14PropertyCounts_Spkr}{RGB}{240,223,178}
\definecolor{colorGolosPropertyCounts_Spkr}{RGB}{240,223,178}
\definecolor{colorMASCPropertyCounts_Spkr}{RGB}{200,234,229}
\definecolor{colorLaboroTVSpeechPropertyCounts_Spkr}{RGB}{240,223,178}
\definecolor{colorKeSpeechPropertyCounts_Spkr}{RGB}{199,234,229}
\definecolor{colorJTUBESPEECHPropertyCounts_Spkr}{RGB}{240,223,178}
\definecolor{colorGigaSpeechPropertyCounts_Spkr}{RGB}{240,223,178}
\definecolor{colorVoxPopuliPropertyCounts_Spkr}{RGB}{206,235,231}
\definecolor{colorSPGISpeechPropertyCounts_Spkr}{RGB}{193,231,226}
\definecolor{colorWestAfr.RadioPropertyCounts_Spkr}{RGB}{240,223,178}
\definecolor{colorAISHELL-4PropertyCounts_Spkr}{RGB}{224,240,237}
\definecolor{colorWestAfr.Virt.Asst.PropertyCounts_Spkr}{RGB}{226,240,238}
\definecolor{colorMediaSpeechPropertyCounts_Spkr}{RGB}{240,223,178}
\definecolor{colorPeople'sSpeechPropertyCounts_Spkr}{RGB}{240,223,178}
\definecolor{color1111HoursHindiPropertyCounts_Spkr}{RGB}{240,223,178}
\definecolor{colorShrutilipiPropertyCounts_Spkr}{RGB}{240,223,178}
\definecolor{colorWenetSpeechPropertyCounts_Spkr}{RGB}{240,223,178}
\definecolor{colorSamromurChildrenPropertyCounts_Spkr}{RGB}{208,236,232}
\definecolor{colorSDS-200PropertyCounts_Spkr}{RGB}{206,235,231}
\definecolor{coloraidatatangPropertyCounts_Spkr}{RGB}{215,237,234}
\definecolor{colorFleursPropertyCounts_Spkr}{RGB}{240,223,178}
\definecolor{colorOLKAVSPropertyCounts_Spkr}{RGB}{211,237,233}
\definecolor{colorNorwegianParl.PropertyCounts_Spkr}{RGB}{218,238,235}
\definecolor{colorMagicData-RAMCPropertyCounts_Spkr}{RGB}{213,237,234}
\definecolor{colorKathbathPropertyCounts_Spkr}{RGB}{211,237,233}
\definecolor{colorHebrewKanPropertyCounts_Spkr}{RGB}{240,223,178}
\definecolor{colorHebrewCourseraPropertyCounts_Spkr}{RGB}{240,223,178}
\definecolor{colorBloomSpeechPropertyCounts_Spkr}{RGB}{240,223,178}
\definecolor{colorEnglish-VietnamesePropertyCounts_Spkr}{RGB}{240,223,178}
\definecolor{colorEarnings-22PropertyCounts_Spkr}{RGB}{222,239,237}
\definecolor{colorYODASPropertyCounts_Spkr}{RGB}{240,223,178}
\definecolor{colorAFRISPEECH-200PropertyCounts_Spkr}{RGB}{208,236,232}
\definecolor{colorAaltoFinnishParl.PropertyCounts_Spkr}{RGB}{215,237,234}
\definecolor{colorReazonSpeechPropertyCounts_Spkr}{RGB}{240,223,178}
\definecolor{colorEdAccPropertyCounts_Spkr}{RGB}{222,239,237}
\definecolor{colorRixVoxPropertyCounts_Spkr}{RGB}{240,223,178}
\definecolor{colorJapaneseAnimeSpeechPropertyCounts_Spkr}{RGB}{240,223,178}
\definecolor{colorSnowMountainPropertyCounts_Spkr}{RGB}{231,241,240}
\definecolor{colorSamromurMilljonPropertyCounts_Spkr}{RGB}{200,234,229}
\definecolor{colorBud500PropertyCounts_Spkr}{RGB}{240,223,178}
\definecolor{colorVibraVoxPropertyCounts_Spkr}{RGB}{218,238,235}
\definecolor{colorM2ASRPropertyCounts_Spkr}{RGB}{215,237,234}
\definecolor{colorTIMITPropertyCounts_Lang}{RGB}{240,223,178}
\definecolor{colorSwitchboardPropertyCounts_Lang}{RGB}{240,223,178}
\definecolor{colorAfricanAcc.FrenchPropertyCounts_Lang}{RGB}{240,223,178}
\definecolor{colorCSJPropertyCounts_Lang}{RGB}{240,223,178}
\definecolor{colorFisherPropertyCounts_Lang}{RGB}{240,223,178}
\definecolor{colorCSLU22Langs.PropertyCounts_Lang}{RGB}{233,242,240}
\definecolor{colorAMIPropertyCounts_Lang}{RGB}{240,223,178}
\definecolor{colorCSLU1.2PropertyCounts_Lang}{RGB}{240,223,178}
\definecolor{colorALLSSTARPropertyCounts_Lang}{RGB}{226,240,238}
\definecolor{colorTED-LIUM3PropertyCounts_Lang}{RGB}{240,223,178}
\definecolor{colorNSTNorwegianPropertyCounts_Lang}{RGB}{240,223,178}
\definecolor{colorNSTDanishPropertyCounts_Lang}{RGB}{240,223,178}
\definecolor{colorNSTSwedishPropertyCounts_Lang}{RGB}{240,223,178}
\definecolor{colorVystadialPropertyCounts_Lang}{RGB}{245,233,198}
\definecolor{colorTHCHS-30PropertyCounts_Lang}{RGB}{240,223,178}
\definecolor{colorLibriSpeechPropertyCounts_Lang}{RGB}{240,223,178}
\definecolor{colorTHUYG-20PropertyCounts_Lang}{RGB}{240,223,178}
\definecolor{colorVCTKPropertyCounts_Lang}{RGB}{240,223,178}
\definecolor{colorSpokenWikipediaPropertyCounts_Lang}{RGB}{245,236,210}
\definecolor{colorAISHELL-1PropertyCounts_Lang}{RGB}{240,223,178}
\definecolor{colorLJSpeechPropertyCounts_Lang}{RGB}{240,223,178}
\definecolor{colorClarinPLPropertyCounts_Lang}{RGB}{240,223,178}
\definecolor{colorAISHELL-2PropertyCounts_Lang}{RGB}{240,223,178}
\definecolor{colorRegionalAf.Am.Lang.PropertyCounts_Lang}{RGB}{240,223,178}
\definecolor{colorCrowdSourcedSpeechPropertyCounts_Lang}{RGB}{245,239,222}
\definecolor{colorZeroth-KoreanPropertyCounts_Lang}{RGB}{240,223,178}
\definecolor{colorRTVEPropertyCounts_Lang}{RGB}{240,223,178}
\definecolor{colorOpenSTTPropertyCounts_Lang}{RGB}{240,223,178}
\definecolor{colorMuST-CPropertyCounts_Lang}{RGB}{238,243,242}
\definecolor{colorM-AILABSPropertyCounts_Lang}{RGB}{245,242,234}
\definecolor{colorMAGICDATAPropertyCounts_Lang}{RGB}{240,223,178}
\definecolor{colorCommonVoice17PropertyCounts_Lang}{RGB}{187,229,223}
\definecolor{colorCoNASEPropertyCounts_Lang}{RGB}{240,223,178}
\definecolor{colorNigerianEnglishPropertyCounts_Lang}{RGB}{240,223,178}
\definecolor{colorNorwegianParl.SpeechPropertyCounts_Lang}{RGB}{240,223,178}
\definecolor{color120hSpanishSpeechPropertyCounts_Lang}{RGB}{240,223,178}
\definecolor{colorDiDiSpeechPropertyCounts_Lang}{RGB}{240,223,178}
\definecolor{colorCzechParliamentPropertyCounts_Lang}{RGB}{240,223,178}
\definecolor{colorCoVoST-2PropertyCounts_Lang}{RGB}{231,241,240}
\definecolor{colorKSCPropertyCounts_Lang}{RGB}{240,223,178}
\definecolor{colorBasq.,Cat.andGal.PropertyCounts_Lang}{RGB}{245,236,210}
\definecolor{colorKsponSpeechPropertyCounts_Lang}{RGB}{240,223,178}
\definecolor{colorSamromurPropertyCounts_Lang}{RGB}{240,223,178}
\definecolor{colorMultiling.LibriSpeechPropertyCounts_Lang}{RGB}{245,242,234}
\definecolor{colorMaSSPropertyCounts_Lang}{RGB}{245,242,234}
\definecolor{colorFTSPEECHPropertyCounts_Lang}{RGB}{240,223,178}
\definecolor{colorEng.Acc.inBrit.IslesPropertyCounts_Lang}{RGB}{240,223,178}
\definecolor{colorHighlandPueblaNahuatlPropertyCounts_Lang}{RGB}{240,223,178}
\definecolor{colorQASRPropertyCounts_Lang}{RGB}{240,223,178}
\definecolor{colorMultiling.TEDxPropertyCounts_Lang}{RGB}{245,243,238}
\definecolor{colorMinds14PropertyCounts_Lang}{RGB}{242,244,244}
\definecolor{colorGolosPropertyCounts_Lang}{RGB}{240,223,178}
\definecolor{colorMASCPropertyCounts_Lang}{RGB}{240,223,178}
\definecolor{colorLaboroTVSpeechPropertyCounts_Lang}{RGB}{245,233,198}
\definecolor{colorKeSpeechPropertyCounts_Lang}{RGB}{245,233,198}
\definecolor{colorJTUBESPEECHPropertyCounts_Lang}{RGB}{245,233,198}
\definecolor{colorGigaSpeechPropertyCounts_Lang}{RGB}{240,223,178}
\definecolor{colorVoxPopuliPropertyCounts_Lang}{RGB}{238,243,242}
\definecolor{colorSPGISpeechPropertyCounts_Lang}{RGB}{240,223,178}
\definecolor{colorWestAfr.RadioPropertyCounts_Lang}{RGB}{245,243,240}
\definecolor{colorAISHELL-4PropertyCounts_Lang}{RGB}{240,223,178}
\definecolor{colorWestAfr.Virt.Asst.PropertyCounts_Lang}{RGB}{245,236,210}
\definecolor{colorMediaSpeechPropertyCounts_Lang}{RGB}{245,237,216}
\definecolor{colorPeople'sSpeechPropertyCounts_Lang}{RGB}{240,223,178}
\definecolor{color1111HoursHindiPropertyCounts_Lang}{RGB}{240,223,178}
\definecolor{colorShrutilipiPropertyCounts_Lang}{RGB}{245,244,244}
\definecolor{colorWenetSpeechPropertyCounts_Lang}{RGB}{240,223,178}
\definecolor{colorSamromurChildrenPropertyCounts_Lang}{RGB}{240,223,178}
\definecolor{colorSDS-200PropertyCounts_Lang}{RGB}{240,223,178}
\definecolor{coloraidatatangPropertyCounts_Lang}{RGB}{240,223,178}
\definecolor{colorFleursPropertyCounts_Lang}{RGB}{193,231,226}
\definecolor{colorOLKAVSPropertyCounts_Lang}{RGB}{240,223,178}
\definecolor{colorNorwegianParl.PropertyCounts_Lang}{RGB}{240,223,178}
\definecolor{colorMagicData-RAMCPropertyCounts_Lang}{RGB}{240,223,178}
\definecolor{colorKathbathPropertyCounts_Lang}{RGB}{245,244,244}
\definecolor{colorHebrewKanPropertyCounts_Lang}{RGB}{240,223,178}
\definecolor{colorHebrewCourseraPropertyCounts_Lang}{RGB}{240,223,178}
\definecolor{colorBloomSpeechPropertyCounts_Lang}{RGB}{209,236,232}
\definecolor{colorEnglish-VietnamesePropertyCounts_Lang}{RGB}{245,233,198}
\definecolor{colorEarnings-22PropertyCounts_Lang}{RGB}{240,223,178}
\definecolor{colorYODASPropertyCounts_Lang}{RGB}{179,226,219}
\definecolor{colorAFRISPEECH-200PropertyCounts_Lang}{RGB}{233,242,240}
\definecolor{colorAaltoFinnishParl.PropertyCounts_Lang}{RGB}{240,223,178}
\definecolor{colorReazonSpeechPropertyCounts_Lang}{RGB}{240,223,178}
\definecolor{colorEdAccPropertyCounts_Lang}{RGB}{240,223,178}
\definecolor{colorRixVoxPropertyCounts_Lang}{RGB}{240,223,178}
\definecolor{colorJapaneseAnimeSpeechPropertyCounts_Lang}{RGB}{240,223,178}
\definecolor{colorSnowMountainPropertyCounts_Lang}{RGB}{242,244,244}
\definecolor{colorSamromurMilljonPropertyCounts_Lang}{RGB}{240,223,178}
\definecolor{colorBud500PropertyCounts_Lang}{RGB}{240,223,178}
\definecolor{colorVibraVoxPropertyCounts_Lang}{RGB}{240,223,178}
\definecolor{colorM2ASRPropertyCounts_Lang}{RGB}{245,237,216}
\definecolor{colorTIMITPropertyCounts_Creat}{RGB}{245,242,234}
\definecolor{colorSwitchboardPropertyCounts_Creat}{RGB}{240,223,178}
\definecolor{colorAfricanAcc.FrenchPropertyCounts_Creat}{RGB}{240,223,178}
\definecolor{colorCSJPropertyCounts_Creat}{RGB}{240,223,178}
\definecolor{colorFisherPropertyCounts_Creat}{RGB}{240,223,178}
\definecolor{colorCSLU22Langs.PropertyCounts_Creat}{RGB}{240,223,178}
\definecolor{colorAMIPropertyCounts_Creat}{RGB}{240,223,178}
\definecolor{colorCSLU1.2PropertyCounts_Creat}{RGB}{240,223,178}
\definecolor{colorALLSSTARPropertyCounts_Creat}{RGB}{240,223,178}
\definecolor{colorTED-LIUM3PropertyCounts_Creat}{RGB}{245,237,214}
\definecolor{colorNSTNorwegianPropertyCounts_Creat}{RGB}{240,223,178}
\definecolor{colorNSTDanishPropertyCounts_Creat}{RGB}{240,223,178}
\definecolor{colorNSTSwedishPropertyCounts_Creat}{RGB}{240,223,178}
\definecolor{colorVystadialPropertyCounts_Creat}{RGB}{240,223,178}
\definecolor{colorTHCHS-30PropertyCounts_Creat}{RGB}{240,223,178}
\definecolor{colorLibriSpeechPropertyCounts_Creat}{RGB}{240,223,178}
\definecolor{colorTHUYG-20PropertyCounts_Creat}{RGB}{245,237,214}
\definecolor{colorVCTKPropertyCounts_Creat}{RGB}{240,223,178}
\definecolor{colorSpokenWikipediaPropertyCounts_Creat}{RGB}{240,223,178}
\definecolor{colorAISHELL-1PropertyCounts_Creat}{RGB}{245,237,214}
\definecolor{colorLJSpeechPropertyCounts_Creat}{RGB}{240,223,178}
\definecolor{colorClarinPLPropertyCounts_Creat}{RGB}{240,223,178}
\definecolor{colorAISHELL-2PropertyCounts_Creat}{RGB}{245,237,214}
\definecolor{colorRegionalAf.Am.Lang.PropertyCounts_Creat}{RGB}{240,223,178}
\definecolor{colorCrowdSourcedSpeechPropertyCounts_Creat}{RGB}{240,223,178}
\definecolor{colorZeroth-KoreanPropertyCounts_Creat}{RGB}{240,223,178}
\definecolor{colorRTVEPropertyCounts_Creat}{RGB}{240,223,178}
\definecolor{colorOpenSTTPropertyCounts_Creat}{RGB}{245,237,214}
\definecolor{colorMuST-CPropertyCounts_Creat}{RGB}{245,237,214}
\definecolor{colorM-AILABSPropertyCounts_Creat}{RGB}{240,223,178}
\definecolor{colorMAGICDATAPropertyCounts_Creat}{RGB}{240,223,178}
\definecolor{colorCommonVoice17PropertyCounts_Creat}{RGB}{245,242,234}
\definecolor{colorCoNASEPropertyCounts_Creat}{RGB}{240,223,178}
\definecolor{colorNigerianEnglishPropertyCounts_Creat}{RGB}{240,223,178}
\definecolor{colorNorwegianParl.SpeechPropertyCounts_Creat}{RGB}{240,223,178}
\definecolor{color120hSpanishSpeechPropertyCounts_Creat}{RGB}{240,223,178}
\definecolor{colorDiDiSpeechPropertyCounts_Creat}{RGB}{240,223,178}
\definecolor{colorCzechParliamentPropertyCounts_Creat}{RGB}{240,223,178}
\definecolor{colorCoVoST-2PropertyCounts_Creat}{RGB}{240,223,178}
\definecolor{colorKSCPropertyCounts_Creat}{RGB}{240,223,178}
\definecolor{colorBasq.,Cat.andGal.PropertyCounts_Creat}{RGB}{240,223,178}
\definecolor{colorKsponSpeechPropertyCounts_Creat}{RGB}{240,223,178}
\definecolor{colorSamromurPropertyCounts_Creat}{RGB}{240,223,178}
\definecolor{colorMultiling.LibriSpeechPropertyCounts_Creat}{RGB}{240,223,178}
\definecolor{colorMaSSPropertyCounts_Creat}{RGB}{240,223,178}
\definecolor{colorFTSPEECHPropertyCounts_Creat}{RGB}{245,237,214}
\definecolor{colorEng.Acc.inBrit.IslesPropertyCounts_Creat}{RGB}{240,223,178}
\definecolor{colorHighlandPueblaNahuatlPropertyCounts_Creat}{RGB}{245,242,234}
\definecolor{colorQASRPropertyCounts_Creat}{RGB}{245,237,214}
\definecolor{colorMultiling.TEDxPropertyCounts_Creat}{RGB}{245,242,234}
\definecolor{colorMinds14PropertyCounts_Creat}{RGB}{240,223,178}
\definecolor{colorGolosPropertyCounts_Creat}{RGB}{245,242,234}
\definecolor{colorMASCPropertyCounts_Creat}{RGB}{245,242,234}
\definecolor{colorLaboroTVSpeechPropertyCounts_Creat}{RGB}{245,237,214}
\definecolor{colorKeSpeechPropertyCounts_Creat}{RGB}{240,223,178}
\definecolor{colorJTUBESPEECHPropertyCounts_Creat}{RGB}{242,244,244}
\definecolor{colorGigaSpeechPropertyCounts_Creat}{RGB}{206,235,231}
\definecolor{colorVoxPopuliPropertyCounts_Creat}{RGB}{240,223,178}
\definecolor{colorSPGISpeechPropertyCounts_Creat}{RGB}{242,244,244}
\definecolor{colorWestAfr.RadioPropertyCounts_Creat}{RGB}{245,237,214}
\definecolor{colorAISHELL-4PropertyCounts_Creat}{RGB}{242,244,244}
\definecolor{colorWestAfr.Virt.Asst.PropertyCounts_Creat}{RGB}{245,237,214}
\definecolor{colorMediaSpeechPropertyCounts_Creat}{RGB}{231,241,240}
\definecolor{colorPeople'sSpeechPropertyCounts_Creat}{RGB}{217,238,235}
\definecolor{color1111HoursHindiPropertyCounts_Creat}{RGB}{240,223,178}
\definecolor{colorShrutilipiPropertyCounts_Creat}{RGB}{245,237,214}
\definecolor{colorWenetSpeechPropertyCounts_Creat}{RGB}{242,244,244}
\definecolor{colorSamromurChildrenPropertyCounts_Creat}{RGB}{240,223,178}
\definecolor{colorSDS-200PropertyCounts_Creat}{RGB}{245,242,234}
\definecolor{coloraidatatangPropertyCounts_Creat}{RGB}{240,223,178}
\definecolor{colorFleursPropertyCounts_Creat}{RGB}{245,242,234}
\definecolor{colorOLKAVSPropertyCounts_Creat}{RGB}{245,237,214}
\definecolor{colorNorwegianParl.PropertyCounts_Creat}{RGB}{245,237,214}
\definecolor{colorMagicData-RAMCPropertyCounts_Creat}{RGB}{242,244,244}
\definecolor{colorKathbathPropertyCounts_Creat}{RGB}{245,237,214}
\definecolor{colorHebrewKanPropertyCounts_Creat}{RGB}{240,223,178}
\definecolor{colorHebrewCourseraPropertyCounts_Creat}{RGB}{240,223,178}
\definecolor{colorBloomSpeechPropertyCounts_Creat}{RGB}{231,241,240}
\definecolor{colorEnglish-VietnamesePropertyCounts_Creat}{RGB}{240,223,178}
\definecolor{colorEarnings-22PropertyCounts_Creat}{RGB}{240,223,178}
\definecolor{colorYODASPropertyCounts_Creat}{RGB}{245,242,234}
\definecolor{colorAFRISPEECH-200PropertyCounts_Creat}{RGB}{179,226,219}
\definecolor{colorAaltoFinnishParl.PropertyCounts_Creat}{RGB}{240,223,178}
\definecolor{colorReazonSpeechPropertyCounts_Creat}{RGB}{245,237,214}
\definecolor{colorEdAccPropertyCounts_Creat}{RGB}{240,223,178}
\definecolor{colorRixVoxPropertyCounts_Creat}{RGB}{240,223,178}
\definecolor{colorJapaneseAnimeSpeechPropertyCounts_Creat}{RGB}{240,223,178}
\definecolor{colorSnowMountainPropertyCounts_Creat}{RGB}{245,237,214}
\definecolor{colorSamromurMilljonPropertyCounts_Creat}{RGB}{240,223,178}
\definecolor{colorBud500PropertyCounts_Creat}{RGB}{240,223,178}
\definecolor{colorVibraVoxPropertyCounts_Creat}{RGB}{240,223,178}
\definecolor{colorM2ASRPropertyCounts_Creat}{RGB}{245,242,234}
\definecolor{colorTIMITPropertyCounts_Tasks}{RGB}{245,242,234}
\definecolor{colorSwitchboardPropertyCounts_Tasks}{RGB}{240,223,178}
\definecolor{colorAfricanAcc.FrenchPropertyCounts_Tasks}{RGB}{240,223,178}
\definecolor{colorCSJPropertyCounts_Tasks}{RGB}{240,223,178}
\definecolor{colorFisherPropertyCounts_Tasks}{RGB}{240,223,178}
\definecolor{colorCSLU22Langs.PropertyCounts_Tasks}{RGB}{240,223,178}
\definecolor{colorAMIPropertyCounts_Tasks}{RGB}{240,223,178}
\definecolor{colorCSLU1.2PropertyCounts_Tasks}{RGB}{240,223,178}
\definecolor{colorALLSSTARPropertyCounts_Tasks}{RGB}{240,223,178}
\definecolor{colorTED-LIUM3PropertyCounts_Tasks}{RGB}{245,237,214}
\definecolor{colorNSTNorwegianPropertyCounts_Tasks}{RGB}{240,223,178}
\definecolor{colorNSTDanishPropertyCounts_Tasks}{RGB}{240,223,178}
\definecolor{colorNSTSwedishPropertyCounts_Tasks}{RGB}{240,223,178}
\definecolor{colorVystadialPropertyCounts_Tasks}{RGB}{240,223,178}
\definecolor{colorTHCHS-30PropertyCounts_Tasks}{RGB}{240,223,178}
\definecolor{colorLibriSpeechPropertyCounts_Tasks}{RGB}{240,223,178}
\definecolor{colorTHUYG-20PropertyCounts_Tasks}{RGB}{245,237,214}
\definecolor{colorVCTKPropertyCounts_Tasks}{RGB}{240,223,178}
\definecolor{colorSpokenWikipediaPropertyCounts_Tasks}{RGB}{240,223,178}
\definecolor{colorAISHELL-1PropertyCounts_Tasks}{RGB}{245,237,214}
\definecolor{colorLJSpeechPropertyCounts_Tasks}{RGB}{240,223,178}
\definecolor{colorClarinPLPropertyCounts_Tasks}{RGB}{240,223,178}
\definecolor{colorAISHELL-2PropertyCounts_Tasks}{RGB}{245,237,214}
\definecolor{colorRegionalAf.Am.Lang.PropertyCounts_Tasks}{RGB}{240,223,178}
\definecolor{colorCrowdSourcedSpeechPropertyCounts_Tasks}{RGB}{240,223,178}
\definecolor{colorZeroth-KoreanPropertyCounts_Tasks}{RGB}{240,223,178}
\definecolor{colorRTVEPropertyCounts_Tasks}{RGB}{240,223,178}
\definecolor{colorOpenSTTPropertyCounts_Tasks}{RGB}{245,237,214}
\definecolor{colorMuST-CPropertyCounts_Tasks}{RGB}{245,237,214}
\definecolor{colorM-AILABSPropertyCounts_Tasks}{RGB}{240,223,178}
\definecolor{colorMAGICDATAPropertyCounts_Tasks}{RGB}{240,223,178}
\definecolor{colorCommonVoice17PropertyCounts_Tasks}{RGB}{245,242,234}
\definecolor{colorCoNASEPropertyCounts_Tasks}{RGB}{240,223,178}
\definecolor{colorNigerianEnglishPropertyCounts_Tasks}{RGB}{240,223,178}
\definecolor{colorNorwegianParl.SpeechPropertyCounts_Tasks}{RGB}{240,223,178}
\definecolor{color120hSpanishSpeechPropertyCounts_Tasks}{RGB}{240,223,178}
\definecolor{colorDiDiSpeechPropertyCounts_Tasks}{RGB}{240,223,178}
\definecolor{colorCzechParliamentPropertyCounts_Tasks}{RGB}{240,223,178}
\definecolor{colorCoVoST-2PropertyCounts_Tasks}{RGB}{240,223,178}
\definecolor{colorKSCPropertyCounts_Tasks}{RGB}{240,223,178}
\definecolor{colorBasq.,Cat.andGal.PropertyCounts_Tasks}{RGB}{240,223,178}
\definecolor{colorKsponSpeechPropertyCounts_Tasks}{RGB}{240,223,178}
\definecolor{colorSamromurPropertyCounts_Tasks}{RGB}{240,223,178}
\definecolor{colorMultiling.LibriSpeechPropertyCounts_Tasks}{RGB}{240,223,178}
\definecolor{colorMaSSPropertyCounts_Tasks}{RGB}{240,223,178}
\definecolor{colorFTSPEECHPropertyCounts_Tasks}{RGB}{245,237,214}
\definecolor{colorEng.Acc.inBrit.IslesPropertyCounts_Tasks}{RGB}{240,223,178}
\definecolor{colorHighlandPueblaNahuatlPropertyCounts_Tasks}{RGB}{245,242,234}
\definecolor{colorQASRPropertyCounts_Tasks}{RGB}{245,237,214}
\definecolor{colorMultiling.TEDxPropertyCounts_Tasks}{RGB}{245,242,234}
\definecolor{colorMinds14PropertyCounts_Tasks}{RGB}{240,223,178}
\definecolor{colorGolosPropertyCounts_Tasks}{RGB}{245,242,234}
\definecolor{colorMASCPropertyCounts_Tasks}{RGB}{245,242,234}
\definecolor{colorLaboroTVSpeechPropertyCounts_Tasks}{RGB}{245,237,214}
\definecolor{colorKeSpeechPropertyCounts_Tasks}{RGB}{240,223,178}
\definecolor{colorJTUBESPEECHPropertyCounts_Tasks}{RGB}{242,244,244}
\definecolor{colorGigaSpeechPropertyCounts_Tasks}{RGB}{206,235,231}
\definecolor{colorVoxPopuliPropertyCounts_Tasks}{RGB}{240,223,178}
\definecolor{colorSPGISpeechPropertyCounts_Tasks}{RGB}{242,244,244}
\definecolor{colorWestAfr.RadioPropertyCounts_Tasks}{RGB}{245,237,214}
\definecolor{colorAISHELL-4PropertyCounts_Tasks}{RGB}{242,244,244}
\definecolor{colorWestAfr.Virt.Asst.PropertyCounts_Tasks}{RGB}{245,237,214}
\definecolor{colorMediaSpeechPropertyCounts_Tasks}{RGB}{231,241,240}
\definecolor{colorPeople'sSpeechPropertyCounts_Tasks}{RGB}{217,238,235}
\definecolor{color1111HoursHindiPropertyCounts_Tasks}{RGB}{240,223,178}
\definecolor{colorShrutilipiPropertyCounts_Tasks}{RGB}{245,237,214}
\definecolor{colorWenetSpeechPropertyCounts_Tasks}{RGB}{242,244,244}
\definecolor{colorSamromurChildrenPropertyCounts_Tasks}{RGB}{240,223,178}
\definecolor{colorSDS-200PropertyCounts_Tasks}{RGB}{245,242,234}
\definecolor{coloraidatatangPropertyCounts_Tasks}{RGB}{240,223,178}
\definecolor{colorFleursPropertyCounts_Tasks}{RGB}{245,242,234}
\definecolor{colorOLKAVSPropertyCounts_Tasks}{RGB}{245,237,214}
\definecolor{colorNorwegianParl.PropertyCounts_Tasks}{RGB}{245,237,214}
\definecolor{colorMagicData-RAMCPropertyCounts_Tasks}{RGB}{242,244,244}
\definecolor{colorKathbathPropertyCounts_Tasks}{RGB}{245,237,214}
\definecolor{colorHebrewKanPropertyCounts_Tasks}{RGB}{240,223,178}
\definecolor{colorHebrewCourseraPropertyCounts_Tasks}{RGB}{240,223,178}
\definecolor{colorBloomSpeechPropertyCounts_Tasks}{RGB}{231,241,240}
\definecolor{colorEnglish-VietnamesePropertyCounts_Tasks}{RGB}{240,223,178}
\definecolor{colorEarnings-22PropertyCounts_Tasks}{RGB}{240,223,178}
\definecolor{colorYODASPropertyCounts_Tasks}{RGB}{245,242,234}
\definecolor{colorAFRISPEECH-200PropertyCounts_Tasks}{RGB}{179,226,219}
\definecolor{colorAaltoFinnishParl.PropertyCounts_Tasks}{RGB}{240,223,178}
\definecolor{colorReazonSpeechPropertyCounts_Tasks}{RGB}{245,237,214}
\definecolor{colorEdAccPropertyCounts_Tasks}{RGB}{240,223,178}
\definecolor{colorRixVoxPropertyCounts_Tasks}{RGB}{240,223,178}
\definecolor{colorJapaneseAnimeSpeechPropertyCounts_Tasks}{RGB}{240,223,178}
\definecolor{colorSnowMountainPropertyCounts_Tasks}{RGB}{245,237,214}
\definecolor{colorSamromurMilljonPropertyCounts_Tasks}{RGB}{240,223,178}
\definecolor{colorBud500PropertyCounts_Tasks}{RGB}{240,223,178}
\definecolor{colorVibraVoxPropertyCounts_Tasks}{RGB}{240,223,178}
\definecolor{colorM2ASRPropertyCounts_Tasks}{RGB}{245,242,234}
\definecolor{colorTIMITPropertyCounts_Src}{RGB}{240,223,178}
\definecolor{colorSwitchboardPropertyCounts_Src}{RGB}{240,223,178}
\definecolor{colorAfricanAcc.FrenchPropertyCounts_Src}{RGB}{240,223,178}
\definecolor{colorCSJPropertyCounts_Src}{RGB}{240,223,178}
\definecolor{colorFisherPropertyCounts_Src}{RGB}{240,223,178}
\definecolor{colorCSLU22Langs.PropertyCounts_Src}{RGB}{240,223,178}
\definecolor{colorAMIPropertyCounts_Src}{RGB}{245,237,216}
\definecolor{colorCSLU1.2PropertyCounts_Src}{RGB}{240,223,178}
\definecolor{colorALLSSTARPropertyCounts_Src}{RGB}{240,223,178}
\definecolor{colorTED-LIUM3PropertyCounts_Src}{RGB}{240,223,178}
\definecolor{colorNSTNorwegianPropertyCounts_Src}{RGB}{240,223,178}
\definecolor{colorNSTDanishPropertyCounts_Src}{RGB}{240,223,178}
\definecolor{colorNSTSwedishPropertyCounts_Src}{RGB}{240,223,178}
\definecolor{colorVystadialPropertyCounts_Src}{RGB}{245,237,216}
\definecolor{colorTHCHS-30PropertyCounts_Src}{RGB}{240,223,178}
\definecolor{colorLibriSpeechPropertyCounts_Src}{RGB}{240,223,178}
\definecolor{colorTHUYG-20PropertyCounts_Src}{RGB}{240,223,178}
\definecolor{colorVCTKPropertyCounts_Src}{RGB}{240,223,178}
\definecolor{colorSpokenWikipediaPropertyCounts_Src}{RGB}{240,223,178}
\definecolor{colorAISHELL-1PropertyCounts_Src}{RGB}{245,237,216}
\definecolor{colorLJSpeechPropertyCounts_Src}{RGB}{240,223,178}
\definecolor{colorClarinPLPropertyCounts_Src}{RGB}{245,237,216}
\definecolor{colorAISHELL-2PropertyCounts_Src}{RGB}{240,223,178}
\definecolor{colorRegionalAf.Am.Lang.PropertyCounts_Src}{RGB}{240,223,178}
\definecolor{colorCrowdSourcedSpeechPropertyCounts_Src}{RGB}{240,223,178}
\definecolor{colorZeroth-KoreanPropertyCounts_Src}{RGB}{240,223,178}
\definecolor{colorRTVEPropertyCounts_Src}{RGB}{240,223,178}
\definecolor{colorOpenSTTPropertyCounts_Src}{RGB}{245,237,216}
\definecolor{colorMuST-CPropertyCounts_Src}{RGB}{240,223,178}
\definecolor{colorM-AILABSPropertyCounts_Src}{RGB}{240,223,178}
\definecolor{colorMAGICDATAPropertyCounts_Src}{RGB}{240,223,178}
\definecolor{colorCommonVoice17PropertyCounts_Src}{RGB}{240,223,178}
\definecolor{colorCoNASEPropertyCounts_Src}{RGB}{240,223,178}
\definecolor{colorNigerianEnglishPropertyCounts_Src}{RGB}{240,223,178}
\definecolor{colorNorwegianParl.SpeechPropertyCounts_Src}{RGB}{240,223,178}
\definecolor{color120hSpanishSpeechPropertyCounts_Src}{RGB}{240,223,178}
\definecolor{colorDiDiSpeechPropertyCounts_Src}{RGB}{240,223,178}
\definecolor{colorCzechParliamentPropertyCounts_Src}{RGB}{240,223,178}
\definecolor{colorCoVoST-2PropertyCounts_Src}{RGB}{245,237,216}
\definecolor{colorKSCPropertyCounts_Src}{RGB}{240,223,178}
\definecolor{colorBasq.,Cat.andGal.PropertyCounts_Src}{RGB}{240,223,178}
\definecolor{colorKsponSpeechPropertyCounts_Src}{RGB}{240,223,178}
\definecolor{colorSamromurPropertyCounts_Src}{RGB}{240,223,178}
\definecolor{colorMultiling.LibriSpeechPropertyCounts_Src}{RGB}{240,223,178}
\definecolor{colorMaSSPropertyCounts_Src}{RGB}{240,223,178}
\definecolor{colorFTSPEECHPropertyCounts_Src}{RGB}{240,223,178}
\definecolor{colorEng.Acc.inBrit.IslesPropertyCounts_Src}{RGB}{240,223,178}
\definecolor{colorHighlandPueblaNahuatlPropertyCounts_Src}{RGB}{240,223,178}
\definecolor{colorQASRPropertyCounts_Src}{RGB}{240,223,178}
\definecolor{colorMultiling.TEDxPropertyCounts_Src}{RGB}{240,223,178}
\definecolor{colorMinds14PropertyCounts_Src}{RGB}{245,237,216}
\definecolor{colorGolosPropertyCounts_Src}{RGB}{240,223,178}
\definecolor{colorMASCPropertyCounts_Src}{RGB}{240,223,178}
\definecolor{colorLaboroTVSpeechPropertyCounts_Src}{RGB}{240,223,178}
\definecolor{colorKeSpeechPropertyCounts_Src}{RGB}{240,223,178}
\definecolor{colorJTUBESPEECHPropertyCounts_Src}{RGB}{240,223,178}
\definecolor{colorGigaSpeechPropertyCounts_Src}{RGB}{245,243,238}
\definecolor{colorVoxPopuliPropertyCounts_Src}{RGB}{240,223,178}
\definecolor{colorSPGISpeechPropertyCounts_Src}{RGB}{240,223,178}
\definecolor{colorWestAfr.RadioPropertyCounts_Src}{RGB}{240,223,178}
\definecolor{colorAISHELL-4PropertyCounts_Src}{RGB}{245,237,216}
\definecolor{colorWestAfr.Virt.Asst.PropertyCounts_Src}{RGB}{240,223,178}
\definecolor{colorMediaSpeechPropertyCounts_Src}{RGB}{179,226,219}
\definecolor{colorPeople'sSpeechPropertyCounts_Src}{RGB}{245,237,216}
\definecolor{color1111HoursHindiPropertyCounts_Src}{RGB}{240,223,178}
\definecolor{colorShrutilipiPropertyCounts_Src}{RGB}{240,223,178}
\definecolor{colorWenetSpeechPropertyCounts_Src}{RGB}{245,237,216}
\definecolor{colorSamromurChildrenPropertyCounts_Src}{RGB}{240,223,178}
\definecolor{colorSDS-200PropertyCounts_Src}{RGB}{240,223,178}
\definecolor{coloraidatatangPropertyCounts_Src}{RGB}{240,223,178}
\definecolor{colorFleursPropertyCounts_Src}{RGB}{240,223,178}
\definecolor{colorOLKAVSPropertyCounts_Src}{RGB}{240,223,178}
\definecolor{colorNorwegianParl.PropertyCounts_Src}{RGB}{240,223,178}
\definecolor{colorMagicData-RAMCPropertyCounts_Src}{RGB}{240,223,178}
\definecolor{colorKathbathPropertyCounts_Src}{RGB}{240,223,178}
\definecolor{colorHebrewKanPropertyCounts_Src}{RGB}{240,223,178}
\definecolor{colorHebrewCourseraPropertyCounts_Src}{RGB}{240,223,178}
\definecolor{colorBloomSpeechPropertyCounts_Src}{RGB}{240,223,178}
\definecolor{colorEnglish-VietnamesePropertyCounts_Src}{RGB}{240,223,178}
\definecolor{colorEarnings-22PropertyCounts_Src}{RGB}{245,243,238}
\definecolor{colorYODASPropertyCounts_Src}{RGB}{240,223,178}
\definecolor{colorAFRISPEECH-200PropertyCounts_Src}{RGB}{240,223,178}
\definecolor{colorAaltoFinnishParl.PropertyCounts_Src}{RGB}{240,223,178}
\definecolor{colorReazonSpeechPropertyCounts_Src}{RGB}{240,223,178}
\definecolor{colorEdAccPropertyCounts_Src}{RGB}{240,223,178}
\definecolor{colorRixVoxPropertyCounts_Src}{RGB}{240,223,178}
\definecolor{colorJapaneseAnimeSpeechPropertyCounts_Src}{RGB}{240,223,178}
\definecolor{colorSnowMountainPropertyCounts_Src}{RGB}{240,223,178}
\definecolor{colorSamromurMilljonPropertyCounts_Src}{RGB}{240,223,178}
\definecolor{colorBud500PropertyCounts_Src}{RGB}{245,237,216}
\definecolor{colorVibraVoxPropertyCounts_Src}{RGB}{240,223,178}
\definecolor{colorM2ASRPropertyCounts_Src}{RGB}{240,223,178}
\definecolor{colorTIMITPropertyCounts_Top}{RGB}{245,242,234}
\definecolor{colorSwitchboardPropertyCounts_Top}{RGB}{196,232,227}
\definecolor{colorAfricanAcc.FrenchPropertyCounts_Top}{RGB}{245,242,234}
\definecolor{colorCSJPropertyCounts_Top}{RGB}{245,233,200}
\definecolor{colorFisherPropertyCounts_Top}{RGB}{213,237,234}
\definecolor{colorCSLU22Langs.PropertyCounts_Top}{RGB}{245,242,234}
\definecolor{colorAMIPropertyCounts_Top}{RGB}{245,233,200}
\definecolor{colorCSLU1.2PropertyCounts_Top}{RGB}{240,223,178}
\definecolor{colorALLSSTARPropertyCounts_Top}{RGB}{245,236,212}
\definecolor{colorTED-LIUM3PropertyCounts_Top}{RGB}{240,223,178}
\definecolor{colorNSTNorwegianPropertyCounts_Top}{RGB}{245,242,234}
\definecolor{colorNSTDanishPropertyCounts_Top}{RGB}{245,242,234}
\definecolor{colorNSTSwedishPropertyCounts_Top}{RGB}{245,242,234}
\definecolor{colorVystadialPropertyCounts_Top}{RGB}{245,236,212}
\definecolor{colorTHCHS-30PropertyCounts_Top}{RGB}{240,223,178}
\definecolor{colorLibriSpeechPropertyCounts_Top}{RGB}{179,226,219}
\definecolor{colorTHUYG-20PropertyCounts_Top}{RGB}{245,236,212}
\definecolor{colorVCTKPropertyCounts_Top}{RGB}{240,223,178}
\definecolor{colorSpokenWikipediaPropertyCounts_Top}{RGB}{240,223,178}
\definecolor{colorAISHELL-1PropertyCounts_Top}{RGB}{244,244,244}
\definecolor{colorLJSpeechPropertyCounts_Top}{RGB}{240,223,178}
\definecolor{colorClarinPLPropertyCounts_Top}{RGB}{245,242,234}
\definecolor{colorAISHELL-2PropertyCounts_Top}{RGB}{245,243,238}
\definecolor{colorRegionalAf.Am.Lang.PropertyCounts_Top}{RGB}{245,243,238}
\definecolor{colorCrowdSourcedSpeechPropertyCounts_Top}{RGB}{240,223,178}
\definecolor{colorZeroth-KoreanPropertyCounts_Top}{RGB}{245,242,234}
\definecolor{colorRTVEPropertyCounts_Top}{RGB}{245,242,234}
\definecolor{colorOpenSTTPropertyCounts_Top}{RGB}{245,241,230}
\definecolor{colorMuST-CPropertyCounts_Top}{RGB}{245,238,220}
\definecolor{colorM-AILABSPropertyCounts_Top}{RGB}{217,238,235}
\definecolor{colorMAGICDATAPropertyCounts_Top}{RGB}{240,223,178}
\definecolor{colorCommonVoice17PropertyCounts_Top}{RGB}{240,223,178}
\definecolor{colorCoNASEPropertyCounts_Top}{RGB}{245,241,230}
\definecolor{colorNigerianEnglishPropertyCounts_Top}{RGB}{245,242,234}
\definecolor{colorNorwegianParl.SpeechPropertyCounts_Top}{RGB}{245,242,234}
\definecolor{color120hSpanishSpeechPropertyCounts_Top}{RGB}{245,242,234}
\definecolor{colorDiDiSpeechPropertyCounts_Top}{RGB}{245,233,200}
\definecolor{colorCzechParliamentPropertyCounts_Top}{RGB}{245,242,234}
\definecolor{colorCoVoST-2PropertyCounts_Top}{RGB}{240,223,178}
\definecolor{colorKSCPropertyCounts_Top}{RGB}{245,240,226}
\definecolor{colorBasq.,Cat.andGal.PropertyCounts_Top}{RGB}{245,233,200}
\definecolor{colorKsponSpeechPropertyCounts_Top}{RGB}{245,241,230}
\definecolor{colorSamromurPropertyCounts_Top}{RGB}{245,240,226}
\definecolor{colorMultiling.LibriSpeechPropertyCounts_Top}{RGB}{217,238,235}
\definecolor{colorMaSSPropertyCounts_Top}{RGB}{240,223,178}
\definecolor{colorFTSPEECHPropertyCounts_Top}{RGB}{245,233,200}
\definecolor{colorEng.Acc.inBrit.IslesPropertyCounts_Top}{RGB}{245,238,220}
\definecolor{colorHighlandPueblaNahuatlPropertyCounts_Top}{RGB}{245,242,234}
\definecolor{colorQASRPropertyCounts_Top}{RGB}{245,242,234}
\definecolor{colorMultiling.TEDxPropertyCounts_Top}{RGB}{245,242,234}
\definecolor{colorMinds14PropertyCounts_Top}{RGB}{245,242,234}
\definecolor{colorGolosPropertyCounts_Top}{RGB}{245,241,230}
\definecolor{colorMASCPropertyCounts_Top}{RGB}{235,242,241}
\definecolor{colorLaboroTVSpeechPropertyCounts_Top}{RGB}{245,242,234}
\definecolor{colorKeSpeechPropertyCounts_Top}{RGB}{240,223,178}
\definecolor{colorJTUBESPEECHPropertyCounts_Top}{RGB}{245,242,234}
\definecolor{colorGigaSpeechPropertyCounts_Top}{RGB}{224,240,237}
\definecolor{colorVoxPopuliPropertyCounts_Top}{RGB}{240,223,178}
\definecolor{colorSPGISpeechPropertyCounts_Top}{RGB}{245,233,200}
\definecolor{colorWestAfr.RadioPropertyCounts_Top}{RGB}{245,236,212}
\definecolor{colorAISHELL-4PropertyCounts_Top}{RGB}{245,241,230}
\definecolor{colorWestAfr.Virt.Asst.PropertyCounts_Top}{RGB}{245,233,200}
\definecolor{colorMediaSpeechPropertyCounts_Top}{RGB}{240,223,178}
\definecolor{colorPeople'sSpeechPropertyCounts_Top}{RGB}{236,243,242}
\definecolor{color1111HoursHindiPropertyCounts_Top}{RGB}{245,240,226}
\definecolor{colorShrutilipiPropertyCounts_Top}{RGB}{240,223,178}
\definecolor{colorWenetSpeechPropertyCounts_Top}{RGB}{245,244,244}
\definecolor{colorSamromurChildrenPropertyCounts_Top}{RGB}{245,240,226}
\definecolor{colorSDS-200PropertyCounts_Top}{RGB}{245,233,200}
\definecolor{coloraidatatangPropertyCounts_Top}{RGB}{245,242,234}
\definecolor{colorFleursPropertyCounts_Top}{RGB}{244,244,244}
\definecolor{colorOLKAVSPropertyCounts_Top}{RGB}{236,243,242}
\definecolor{colorNorwegianParl.PropertyCounts_Top}{RGB}{245,233,200}
\definecolor{colorMagicData-RAMCPropertyCounts_Top}{RGB}{235,242,241}
\definecolor{colorKathbathPropertyCounts_Top}{RGB}{245,236,212}
\definecolor{colorHebrewKanPropertyCounts_Top}{RGB}{245,236,212}
\definecolor{colorHebrewCourseraPropertyCounts_Top}{RGB}{245,242,234}
\definecolor{colorBloomSpeechPropertyCounts_Top}{RGB}{245,243,238}
\definecolor{colorEnglish-VietnamesePropertyCounts_Top}{RGB}{245,242,234}
\definecolor{colorEarnings-22PropertyCounts_Top}{RGB}{245,233,200}
\definecolor{colorYODASPropertyCounts_Top}{RGB}{240,223,178}
\definecolor{colorAFRISPEECH-200PropertyCounts_Top}{RGB}{245,241,230}
\definecolor{colorAaltoFinnishParl.PropertyCounts_Top}{RGB}{245,233,200}
\definecolor{colorReazonSpeechPropertyCounts_Top}{RGB}{240,223,178}
\definecolor{colorEdAccPropertyCounts_Top}{RGB}{245,243,238}
\definecolor{colorRixVoxPropertyCounts_Top}{RGB}{245,233,200}
\definecolor{colorJapaneseAnimeSpeechPropertyCounts_Top}{RGB}{245,242,234}
\definecolor{colorSnowMountainPropertyCounts_Top}{RGB}{240,223,178}
\definecolor{colorSamromurMilljonPropertyCounts_Top}{RGB}{245,240,226}
\definecolor{colorBud500PropertyCounts_Top}{RGB}{245,238,220}
\definecolor{colorVibraVoxPropertyCounts_Top}{RGB}{240,223,178}
\definecolor{colorM2ASRPropertyCounts_Top}{RGB}{245,244,242}

\begin{longtable}{lc|ccccccc|ccc|c}
\caption[\textbf{Audio collections and properties}]{\textbf{Audio collections and properties}. Collection properties include numbers of audio hours (\textsc{HR}), speakers (\textsc{SPKR}), languages (\textsc{Lang}), creator institutions (\textsc{Creat}), tasks (\textsc{Tasks}), data sources (\textsc{Src}), and topics (\textsc{Topics}). The number of datasets is not listed because all collections include only one dataset, except for M2ASR which has four. The \textsc{US} column indicates datasets from or partly from the United States, the \textsc{Ac} column datasets created by academic institutions, and the \textsc{Ind} column datasets created by industry. Note that a dataset can have all of these, none of them, or any combination of them. The \textsc{Use} column indicates whether a collection includes data freely usable even for commercial purposes (\protect\CommercialDataCircle), data usable only for noncommercial purposes or academic research (\protect\NCDataCircle) and data whose license status is not specified precisely enough to allow us to determine commercial use permissions (\protect\UnspecifiedDataCircle). Note that each collection may have different datasets with one, two, or all three of these statuses. Datasets are sorted chronologically to highlight trends over time.} \label{tab:collections-speech} \\
\toprule
\textsc{Collection} & \textsc{} & \multicolumn{7}{c}{\textsc{Property Counts}} & \multicolumn{3}{c}{\textsc{Category}} & \textsc{Perm} \\
 & \textsc{\thead{Year}} & \textsc{\thead{Hr}} & \textsc{\thead{Spkr}} & \textsc{\thead{Lang}} & \textsc{\thead{Creat}} & \textsc{\thead{Tasks}} & \textsc{\thead{Src}} & \textsc{\thead{Top}} & \textsc{\thead{US}} & \textsc{\thead{Ac}} & \textsc{\thead{Ind}} & \textsc{\thead{Use}} \\
\midrule
\endfirsthead
\caption[]{\textbf{Audio collections and properties}.} \\
\toprule
\textsc{Collection} & \textsc{} & \multicolumn{7}{c}{\textsc{Property Counts}} & \multicolumn{3}{c}{\textsc{Category}} & \textsc{Perm} \\
 & \textsc{\thead{Year}} & \textsc{\thead{Hr}} & \textsc{\thead{Spkr}} & \textsc{\thead{Lang}} & \textsc{\thead{Creat}} & \textsc{\thead{Tasks}} & \textsc{\thead{Src}} & \textsc{\thead{Top}} & \textsc{\thead{US}} & \textsc{\thead{Ac}} & \textsc{\thead{Ind}} & \textsc{\thead{Use}} \\
\midrule
\endhead
\midrule
\multicolumn{13}{r}{Continued on next page} \\
\midrule
\endfoot
\bottomrule
\endlastfoot
TIMIT & 1990 & \cellcolor{colorTIMITPropertyCounts_Hr}{5} & \cellcolor{colorTIMITPropertyCounts_Spkr}{630} & \cellcolor{colorVibraVoxPropertyCounts_Lang}{1} & \cellcolor{colorM2ASRPropertyCounts_Creat}{3} & \cellcolor{colorM2ASRPropertyCounts_Tasks}{3} & \cellcolor{colorM2ASRPropertyCounts_Src}{1} & \cellcolor{colorJapaneseAnimeSpeechPropertyCounts_Top}{7} & \greencheck & \greencheck & \greencheck & \TransparentCircle \TransparentCircle \NCDataCircle \\
Switchboard & 1992 & \cellcolor{colorSwitchboardPropertyCounts_Hr}{250} & \cellcolor{colorSwitchboardPropertyCounts_Spkr}{543} & \cellcolor{colorVibraVoxPropertyCounts_Lang}{1} & \cellcolor{colorVibraVoxPropertyCounts_Creat}{1} & \cellcolor{colorVibraVoxPropertyCounts_Tasks}{1} & \cellcolor{colorM2ASRPropertyCounts_Src}{1} & \cellcolor{colorSwitchboardPropertyCounts_Top}{70} & \greencheck & \emojiblank & \greencheck & \TransparentCircle \TransparentCircle \NCDataCircle \\
African Acc. French & 2003 & \cellcolor{colorAfricanAcc.FrenchPropertyCounts_Hr}{22} & \cellcolor{colorAfricanAcc.FrenchPropertyCounts_Spkr}{232} & \cellcolor{colorVibraVoxPropertyCounts_Lang}{1} & \cellcolor{colorVibraVoxPropertyCounts_Creat}{1} & \cellcolor{colorVibraVoxPropertyCounts_Tasks}{1} & \cellcolor{colorM2ASRPropertyCounts_Src}{1} & \cellcolor{colorJapaneseAnimeSpeechPropertyCounts_Top}{7} & \greencheck & \emojiblank & \emojiblank & \CommercialDataCircle \TransparentCircle \TransparentCircle \\
CSJ & 2003 & \cellcolor{colorCSJPropertyCounts_Hr}{661} & \cellcolor{colorCSJPropertyCounts_Spkr}{1k} & \cellcolor{colorVibraVoxPropertyCounts_Lang}{1} & \cellcolor{colorVibraVoxPropertyCounts_Creat}{1} & \cellcolor{colorVibraVoxPropertyCounts_Tasks}{1} & \cellcolor{colorM2ASRPropertyCounts_Src}{1} & \cellcolor{colorRixVoxPropertyCounts_Top}{2} & \emojiblank & \emojiblank & \emojiblank & \TransparentCircle \UnspecifiedDataCircle \TransparentCircle \\
\end{longtable}

\setlength{\tabcolsep}{1.9pt}
\definecolor{colorHOLLYWOOD2PropertyCounts_Hours}{RGB}{229,241,239}
\definecolor{colorCollectivePropertyCounts_Hours}{RGB}{240,223,178}
\definecolor{colorHMDBPropertyCounts_Hours}{RGB}{204,235,230}
\definecolor{colorUCF101PropertyCounts_Hours}{RGB}{227,240,239}
\definecolor{colorYouCookPropertyCounts_Hours}{RGB}{213,237,234}
\definecolor{color50SaladsPropertyCounts_Hours}{RGB}{226,240,238}
\definecolor{colorStoryGraphsPropertyCounts_Hours}{RGB}{235,242,241}
\definecolor{colorHollywoodExt.PropertyCounts_Hours}{RGB}{233,242,240}
\definecolor{colorBreakfastPropertyCounts_Hours}{RGB}{224,240,237}
\definecolor{colorSports-1MPropertyCounts_Hours}{RGB}{187,229,223}
\definecolor{colorTHUMOSPropertyCounts_Hours}{RGB}{218,238,235}
\definecolor{colorVideoStoryPropertyCounts_Hours}{RGB}{213,237,234}
\definecolor{colorSumMePropertyCounts_Hours}{RGB}{242,244,244}
\definecolor{colorTVSumPropertyCounts_Hours}{RGB}{236,243,242}
\definecolor{colorVolleyballPropertyCounts_Hours}{RGB}{240,223,178}
\definecolor{colorActivityNetPropertyCounts_Hours}{RGB}{213,237,234}
\definecolor{colorMovieQAPropertyCounts_Hours}{RGB}{217,238,235}
\definecolor{colorMarsPropertyCounts_Hours}{RGB}{240,223,178}
\definecolor{colorNTURGB+DPropertyCounts_Hours}{RGB}{224,240,237}
\definecolor{colorMSR-VTTPropertyCounts_Hours}{RGB}{226,240,238}
\definecolor{colorCharadesPropertyCounts_Hours}{RGB}{224,240,237}
\definecolor{colorVTWPropertyCounts_Hours}{RGB}{218,238,235}
\definecolor{colorYoutube-8MPropertyCounts_Hours}{RGB}{179,226,219}
\definecolor{colorNarratedInstr.Vid.PropertyCounts_Hours}{RGB}{235,242,241}
\definecolor{colorTGIFPropertyCounts_Hours}{RGB}{224,240,237}
\definecolor{colorMultiTHUMOSPropertyCounts_Hours}{RGB}{227,240,239}
\definecolor{colorImageNet-VidPropertyCounts_Hours}{RGB}{233,242,240}
\definecolor{colorPKU-MMDPropertyCounts_Hours}{RGB}{226,240,238}
\definecolor{color20BN-SOMETHINGPropertyCounts_Hours}{RGB}{222,239,237}
\definecolor{colorYouCook2PropertyCounts_Hours}{RGB}{220,239,236}
\definecolor{colorVoxCelebPropertyCounts_Hours}{RGB}{209,236,232}
\definecolor{colorDavisPropertyCounts_Hours}{RGB}{240,223,178}
\definecolor{colorQFVSPropertyCounts_Hours}{RGB}{229,241,239}
\definecolor{colorDiDeMoPropertyCounts_Hours}{RGB}{218,238,235}
\definecolor{colorSOAPropertyCounts_Hours}{RGB}{211,237,233}
\definecolor{colorCharades-EgoPropertyCounts_Hours}{RGB}{224,240,237}
\definecolor{colorEPIC-KITCHENSPropertyCounts_Hours}{RGB}{222,239,237}
\definecolor{colorMovieGraphsPropertyCounts_Hours}{RGB}{222,239,237}
\definecolor{colorHow2PropertyCounts_Hours}{RGB}{209,236,232}
\definecolor{colorVLOGPropertyCounts_Hours}{RGB}{217,238,235}
\definecolor{colorVaTeXPropertyCounts_Hours}{RGB}{222,239,237}
\definecolor{color20BN-jesterPropertyCounts_Hours}{RGB}{231,241,240}
\definecolor{colorHowTo100MPropertyCounts_Hours}{RGB}{187,229,223}
\definecolor{colorCOINPropertyCounts_Hours}{RGB}{215,237,234}
\definecolor{colorMMActPropertyCounts_Hours}{RGB}{222,239,237}
\definecolor{colorHACSPropertyCounts_Hours}{RGB}{213,237,234}
\definecolor{colorCrossTaskPropertyCounts_Hours}{RGB}{217,238,235}
\definecolor{colorMomentsinTimePropertyCounts_Hours}{RGB}{213,237,234}
\definecolor{colorTRECVidPropertyCounts_Hours}{RGB}{213,237,234}
\definecolor{colorMSAPropertyCounts_Hours}{RGB}{215,237,234}
\definecolor{colorToyotaSmarthomePropertyCounts_Hours}{RGB}{218,238,235}
\definecolor{colorTITANPropertyCounts_Hours}{RGB}{238,243,242}
\definecolor{colorVIOLINPropertyCounts_Hours}{RGB}{215,237,234}
\definecolor{colorRareActPropertyCounts_Hours}{RGB}{229,241,239}
\definecolor{colorTinyVIRATPropertyCounts_Hours}{RGB}{231,241,240}
\definecolor{color100DOHPropertyCounts_Hours}{RGB}{206,235,231}
\definecolor{colorOops!PropertyCounts_Hours}{RGB}{226,240,238}
\definecolor{colorOmniSource-WebPropertyCounts_Hours}{RGB}{200,234,229}
\definecolor{colorCondensedMoviesPropertyCounts_Hours}{RGB}{211,237,233}
\definecolor{colorMovieScenesPropertyCounts_Hours}{RGB}{218,238,235}
\definecolor{colorEEVPropertyCounts_Hours}{RGB}{217,238,235}
\definecolor{colorMovie-NetPropertyCounts_Hours}{RGB}{208,236,232}
\definecolor{colorFineGymPropertyCounts_Hours}{RGB}{213,237,234}
\definecolor{colorHAA500PropertyCounts_Hours}{RGB}{235,242,241}
\definecolor{colorLEMMAPropertyCounts_Hours}{RGB}{231,241,240}
\definecolor{colorHVUPropertyCounts_Hours}{RGB}{190,230,224}
\definecolor{colorApesPropertyCounts_Hours}{RGB}{227,240,239}
\definecolor{colorWebVidPropertyCounts_Hours}{RGB}{200,234,229}
\definecolor{colorVideoLTPropertyCounts_Hours}{RGB}{200,234,229}
\definecolor{colorHOMAGEPropertyCounts_Hours}{RGB}{227,240,239}
\definecolor{colorUAV-HumanPropertyCounts_Hours}{RGB}{229,241,239}
\definecolor{colorHD-VILA-100MPropertyCounts_Hours}{RGB}{217,238,235}
\definecolor{colorM-MiTPropertyCounts_Hours}{RGB}{213,237,234}
\definecolor{colorMimeticsPropertyCounts_Hours}{RGB}{242,244,244}
\definecolor{colorSpokenMomentsPropertyCounts_Hours}{RGB}{217,238,235}
\definecolor{colorQuerYDPropertyCounts_Hours}{RGB}{218,238,235}
\definecolor{colorMADPropertyCounts_Hours}{RGB}{211,237,233}
\definecolor{colorFERV39kPropertyCounts_Hours}{RGB}{231,241,240}
\definecolor{colorCDADPropertyCounts_Hours}{RGB}{218,238,235}
\definecolor{colorMVBenchPropertyCounts_Hours}{RGB}{240,223,178}
\definecolor{colorVidPromPropertyCounts_Hours}{RGB}{182,227,220}
\definecolor{colorShareGPT4VideoPropertyCounts_Hours}{RGB}{208,236,232}
\definecolor{colorOpenVid-1MPropertyCounts_Hours}{RGB}{193,231,226}
\definecolor{colorFineVideoPropertyCounts_Hours}{RGB}{208,236,232}
\definecolor{colorDisneyVid.Gen.PropertyCounts_Hours}{RGB}{235,242,241}
\definecolor{colorKineticsPropertyCounts_Hours}{RGB}{206,235,231}
\definecolor{colorEgo4DPropertyCounts_Hours}{RGB}{206,235,231}
\definecolor{colorMPIIPropertyCounts_Hours}{RGB}{222,239,237}
\definecolor{colorProject-AriaPropertyCounts_Hours}{RGB}{211,237,233}
\definecolor{colorAvaPropertyCounts_Hours}{RGB}{220,239,236}
\definecolor{colorLSMDCPropertyCounts_Hours}{RGB}{217,238,235}
\definecolor{colorHOLLYWOOD2PropertyCounts_Datasets}{RGB}{240,223,178}
\definecolor{colorCollectivePropertyCounts_Datasets}{RGB}{240,223,178}
\definecolor{colorHMDBPropertyCounts_Datasets}{RGB}{240,223,178}
\definecolor{colorUCF101PropertyCounts_Datasets}{RGB}{240,223,178}
\definecolor{colorYouCookPropertyCounts_Datasets}{RGB}{240,223,178}
\definecolor{color50SaladsPropertyCounts_Datasets}{RGB}{240,223,178}
\definecolor{colorStoryGraphsPropertyCounts_Datasets}{RGB}{240,223,178}
\definecolor{colorHollywoodExt.PropertyCounts_Datasets}{RGB}{240,223,178}
\definecolor{colorBreakfastPropertyCounts_Datasets}{RGB}{240,223,178}
\definecolor{colorSports-1MPropertyCounts_Datasets}{RGB}{240,223,178}
\definecolor{colorTHUMOSPropertyCounts_Datasets}{RGB}{240,223,178}
\definecolor{colorVideoStoryPropertyCounts_Datasets}{RGB}{240,223,178}
\definecolor{colorSumMePropertyCounts_Datasets}{RGB}{240,223,178}
\definecolor{colorTVSumPropertyCounts_Datasets}{RGB}{240,223,178}
\definecolor{colorVolleyballPropertyCounts_Datasets}{RGB}{240,223,178}
\definecolor{colorActivityNetPropertyCounts_Datasets}{RGB}{240,223,178}
\definecolor{colorMovieQAPropertyCounts_Datasets}{RGB}{240,223,178}
\definecolor{colorMarsPropertyCounts_Datasets}{RGB}{240,223,178}
\definecolor{colorNTURGB+DPropertyCounts_Datasets}{RGB}{240,223,178}
\definecolor{colorMSR-VTTPropertyCounts_Datasets}{RGB}{240,223,178}
\definecolor{colorCharadesPropertyCounts_Datasets}{RGB}{240,223,178}
\definecolor{colorVTWPropertyCounts_Datasets}{RGB}{240,223,178}
\definecolor{colorYoutube-8MPropertyCounts_Datasets}{RGB}{240,223,178}
\definecolor{colorNarratedInstr.Vid.PropertyCounts_Datasets}{RGB}{240,223,178}
\definecolor{colorTGIFPropertyCounts_Datasets}{RGB}{240,223,178}
\definecolor{colorMultiTHUMOSPropertyCounts_Datasets}{RGB}{240,223,178}
\definecolor{colorImageNet-VidPropertyCounts_Datasets}{RGB}{240,223,178}
\definecolor{colorPKU-MMDPropertyCounts_Datasets}{RGB}{240,223,178}
\definecolor{color20BN-SOMETHINGPropertyCounts_Datasets}{RGB}{240,223,178}
\definecolor{colorYouCook2PropertyCounts_Datasets}{RGB}{240,223,178}
\definecolor{colorVoxCelebPropertyCounts_Datasets}{RGB}{240,223,178}
\definecolor{colorDavisPropertyCounts_Datasets}{RGB}{240,223,178}
\definecolor{colorQFVSPropertyCounts_Datasets}{RGB}{240,223,178}
\definecolor{colorDiDeMoPropertyCounts_Datasets}{RGB}{240,223,178}
\definecolor{colorSOAPropertyCounts_Datasets}{RGB}{240,223,178}
\definecolor{colorCharades-EgoPropertyCounts_Datasets}{RGB}{240,223,178}
\definecolor{colorEPIC-KITCHENSPropertyCounts_Datasets}{RGB}{240,223,178}
\definecolor{colorMovieGraphsPropertyCounts_Datasets}{RGB}{240,223,178}
\definecolor{colorHow2PropertyCounts_Datasets}{RGB}{240,223,178}
\definecolor{colorVLOGPropertyCounts_Datasets}{RGB}{240,223,178}
\definecolor{colorVaTeXPropertyCounts_Datasets}{RGB}{240,223,178}
\definecolor{color20BN-jesterPropertyCounts_Datasets}{RGB}{240,223,178}
\definecolor{colorHowTo100MPropertyCounts_Datasets}{RGB}{240,223,178}
\definecolor{colorCOINPropertyCounts_Datasets}{RGB}{240,223,178}
\definecolor{colorMMActPropertyCounts_Datasets}{RGB}{240,223,178}
\definecolor{colorHACSPropertyCounts_Datasets}{RGB}{240,223,178}
\definecolor{colorCrossTaskPropertyCounts_Datasets}{RGB}{240,223,178}
\definecolor{colorMomentsinTimePropertyCounts_Datasets}{RGB}{240,223,178}
\definecolor{colorTRECVidPropertyCounts_Datasets}{RGB}{240,223,178}
\definecolor{colorMSAPropertyCounts_Datasets}{RGB}{240,223,178}
\definecolor{colorToyotaSmarthomePropertyCounts_Datasets}{RGB}{240,223,178}
\definecolor{colorTITANPropertyCounts_Datasets}{RGB}{240,223,178}
\definecolor{colorVIOLINPropertyCounts_Datasets}{RGB}{240,223,178}
\definecolor{colorRareActPropertyCounts_Datasets}{RGB}{240,223,178}
\definecolor{colorTinyVIRATPropertyCounts_Datasets}{RGB}{240,223,178}
\definecolor{color100DOHPropertyCounts_Datasets}{RGB}{240,223,178}
\definecolor{colorOops!PropertyCounts_Datasets}{RGB}{240,223,178}
\definecolor{colorOmniSource-WebPropertyCounts_Datasets}{RGB}{240,223,178}
\definecolor{colorCondensedMoviesPropertyCounts_Datasets}{RGB}{240,223,178}
\definecolor{colorMovieScenesPropertyCounts_Datasets}{RGB}{240,223,178}
\definecolor{colorEEVPropertyCounts_Datasets}{RGB}{240,223,178}
\definecolor{colorMovie-NetPropertyCounts_Datasets}{RGB}{240,223,178}
\definecolor{colorFineGymPropertyCounts_Datasets}{RGB}{240,223,178}
\definecolor{colorHAA500PropertyCounts_Datasets}{RGB}{240,223,178}
\definecolor{colorLEMMAPropertyCounts_Datasets}{RGB}{240,223,178}
\definecolor{colorHVUPropertyCounts_Datasets}{RGB}{240,223,178}
\definecolor{colorApesPropertyCounts_Datasets}{RGB}{240,223,178}
\definecolor{colorWebVidPropertyCounts_Datasets}{RGB}{240,223,178}
\definecolor{colorVideoLTPropertyCounts_Datasets}{RGB}{240,223,178}
\definecolor{colorHOMAGEPropertyCounts_Datasets}{RGB}{240,223,178}
\definecolor{colorUAV-HumanPropertyCounts_Datasets}{RGB}{240,223,178}
\definecolor{colorHD-VILA-100MPropertyCounts_Datasets}{RGB}{240,223,178}
\definecolor{colorM-MiTPropertyCounts_Datasets}{RGB}{240,223,178}
\definecolor{colorMimeticsPropertyCounts_Datasets}{RGB}{240,223,178}
\definecolor{colorSpokenMomentsPropertyCounts_Datasets}{RGB}{240,223,178}
\definecolor{colorQuerYDPropertyCounts_Datasets}{RGB}{240,223,178}
\definecolor{colorMADPropertyCounts_Datasets}{RGB}{240,223,178}
\definecolor{colorFERV39kPropertyCounts_Datasets}{RGB}{240,223,178}
\definecolor{colorCDADPropertyCounts_Datasets}{RGB}{240,223,178}
\definecolor{colorMVBenchPropertyCounts_Datasets}{RGB}{240,223,178}
\definecolor{colorVidPromPropertyCounts_Datasets}{RGB}{240,223,178}
\definecolor{colorShareGPT4VideoPropertyCounts_Datasets}{RGB}{240,223,178}
\definecolor{colorOpenVid-1MPropertyCounts_Datasets}{RGB}{240,223,178}
\definecolor{colorFineVideoPropertyCounts_Datasets}{RGB}{240,223,178}
\definecolor{colorDisneyVid.Gen.PropertyCounts_Datasets}{RGB}{240,223,178}
\definecolor{colorKineticsPropertyCounts_Datasets}{RGB}{179,226,219}
\definecolor{colorEgo4DPropertyCounts_Datasets}{RGB}{229,241,239}
\definecolor{colorMPIIPropertyCounts_Datasets}{RGB}{179,226,219}
\definecolor{colorProject-AriaPropertyCounts_Datasets}{RGB}{229,241,239}
\definecolor{colorAvaPropertyCounts_Datasets}{RGB}{229,241,239}
\definecolor{colorLSMDCPropertyCounts_Datasets}{RGB}{229,241,239}
\definecolor{colorHOLLYWOOD2PropertyCounts_Countries}{RGB}{239,221,175}
\definecolor{colorCollectivePropertyCounts_Countries}{RGB}{239,221,175}
\definecolor{colorHMDBPropertyCounts_Countries}{RGB}{245,244,244}
\definecolor{colorUCF101PropertyCounts_Countries}{RGB}{239,221,175}
\definecolor{colorYouCookPropertyCounts_Countries}{RGB}{239,221,175}
\definecolor{color50SaladsPropertyCounts_Countries}{RGB}{239,221,175}
\definecolor{colorStoryGraphsPropertyCounts_Countries}{RGB}{239,221,175}
\definecolor{colorHollywoodExt.PropertyCounts_Countries}{RGB}{239,221,175}
\definecolor{colorBreakfastPropertyCounts_Countries}{RGB}{245,244,244}
\definecolor{colorSports-1MPropertyCounts_Countries}{RGB}{239,221,175}
\definecolor{colorTHUMOSPropertyCounts_Countries}{RGB}{245,244,244}
\definecolor{colorVideoStoryPropertyCounts_Countries}{RGB}{239,221,175}
\definecolor{colorSumMePropertyCounts_Countries}{RGB}{245,244,244}
\definecolor{colorTVSumPropertyCounts_Countries}{RGB}{239,221,175}
\definecolor{colorVolleyballPropertyCounts_Countries}{RGB}{239,221,175}
\definecolor{colorActivityNetPropertyCounts_Countries}{RGB}{245,244,244}
\definecolor{colorMovieQAPropertyCounts_Countries}{RGB}{211,237,233}
\definecolor{colorMarsPropertyCounts_Countries}{RGB}{239,221,175}
\definecolor{colorNTURGB+DPropertyCounts_Countries}{RGB}{239,221,175}
\definecolor{colorMSR-VTTPropertyCounts_Countries}{RGB}{239,221,175}
\definecolor{colorCharadesPropertyCounts_Countries}{RGB}{245,244,244}
\definecolor{colorVTWPropertyCounts_Countries}{RGB}{245,244,244}
\definecolor{colorYoutube-8MPropertyCounts_Countries}{RGB}{239,221,175}
\definecolor{colorNarratedInstr.Vid.PropertyCounts_Countries}{RGB}{245,244,244}
\definecolor{colorTGIFPropertyCounts_Countries}{RGB}{239,221,175}
\definecolor{colorMultiTHUMOSPropertyCounts_Countries}{RGB}{245,244,244}
\definecolor{colorImageNet-VidPropertyCounts_Countries}{RGB}{239,221,175}
\definecolor{colorPKU-MMDPropertyCounts_Countries}{RGB}{239,221,175}
\definecolor{color20BN-SOMETHINGPropertyCounts_Countries}{RGB}{239,221,175}
\definecolor{colorYouCook2PropertyCounts_Countries}{RGB}{239,221,175}
\definecolor{colorVoxCelebPropertyCounts_Countries}{RGB}{245,244,244}
\definecolor{colorDavisPropertyCounts_Countries}{RGB}{239,221,175}
\definecolor{colorQFVSPropertyCounts_Countries}{RGB}{239,221,175}
\definecolor{colorDiDeMoPropertyCounts_Countries}{RGB}{239,221,175}
\definecolor{colorSOAPropertyCounts_Countries}{RGB}{239,221,175}
\definecolor{colorCharades-EgoPropertyCounts_Countries}{RGB}{239,221,175}
\definecolor{colorEPIC-KITCHENSPropertyCounts_Countries}{RGB}{211,237,233}
\definecolor{colorMovieGraphsPropertyCounts_Countries}{RGB}{239,221,175}
\definecolor{colorHow2PropertyCounts_Countries}{RGB}{239,221,175}
\definecolor{colorVLOGPropertyCounts_Countries}{RGB}{239,221,175}
\definecolor{colorVaTeXPropertyCounts_Countries}{RGB}{245,244,244}
\definecolor{color20BN-jesterPropertyCounts_Countries}{RGB}{239,221,175}
\definecolor{colorHowTo100MPropertyCounts_Countries}{RGB}{245,244,244}
\definecolor{colorCOINPropertyCounts_Countries}{RGB}{239,221,175}
\definecolor{colorMMActPropertyCounts_Countries}{RGB}{245,244,244}
\definecolor{colorHACSPropertyCounts_Countries}{RGB}{239,221,175}
\definecolor{colorCrossTaskPropertyCounts_Countries}{RGB}{179,226,219}
\definecolor{colorMomentsinTimePropertyCounts_Countries}{RGB}{239,221,175}
\definecolor{colorTRECVidPropertyCounts_Countries}{RGB}{239,221,175}
\definecolor{colorMSAPropertyCounts_Countries}{RGB}{245,244,244}
\definecolor{colorToyotaSmarthomePropertyCounts_Countries}{RGB}{239,221,175}
\definecolor{colorTITANPropertyCounts_Countries}{RGB}{239,221,175}
\definecolor{colorVIOLINPropertyCounts_Countries}{RGB}{239,221,175}
\definecolor{colorRareActPropertyCounts_Countries}{RGB}{211,237,233}
\definecolor{colorTinyVIRATPropertyCounts_Countries}{RGB}{239,221,175}
\definecolor{color100DOHPropertyCounts_Countries}{RGB}{239,221,175}
\definecolor{colorOops!PropertyCounts_Countries}{RGB}{239,221,175}
\definecolor{colorOmniSource-WebPropertyCounts_Countries}{RGB}{239,221,175}
\definecolor{colorCondensedMoviesPropertyCounts_Countries}{RGB}{239,221,175}
\definecolor{colorMovieScenesPropertyCounts_Countries}{RGB}{245,244,244}
\definecolor{colorEEVPropertyCounts_Countries}{RGB}{239,221,175}
\definecolor{colorMovie-NetPropertyCounts_Countries}{RGB}{239,221,175}
\definecolor{colorFineGymPropertyCounts_Countries}{RGB}{239,221,175}
\definecolor{colorHAA500PropertyCounts_Countries}{RGB}{245,244,244}
\definecolor{colorLEMMAPropertyCounts_Countries}{RGB}{239,221,175}
\definecolor{colorHVUPropertyCounts_Countries}{RGB}{211,237,233}
\definecolor{colorApesPropertyCounts_Countries}{RGB}{211,237,233}
\definecolor{colorWebVidPropertyCounts_Countries}{RGB}{245,244,244}
\definecolor{colorVideoLTPropertyCounts_Countries}{RGB}{245,244,244}
\definecolor{colorHOMAGEPropertyCounts_Countries}{RGB}{239,221,175}
\definecolor{colorUAV-HumanPropertyCounts_Countries}{RGB}{245,244,244}
\definecolor{colorHD-VILA-100MPropertyCounts_Countries}{RGB}{239,221,175}
\definecolor{colorM-MiTPropertyCounts_Countries}{RGB}{239,221,175}
\definecolor{colorMimeticsPropertyCounts_Countries}{RGB}{239,221,175}
\definecolor{colorSpokenMomentsPropertyCounts_Countries}{RGB}{239,221,175}
\definecolor{colorQuerYDPropertyCounts_Countries}{RGB}{239,221,175}
\definecolor{colorMADPropertyCounts_Countries}{RGB}{239,221,175}
\definecolor{colorFERV39kPropertyCounts_Countries}{RGB}{239,221,175}
\definecolor{colorCDADPropertyCounts_Countries}{RGB}{239,221,175}
\definecolor{colorMVBenchPropertyCounts_Countries}{RGB}{239,221,175}
\definecolor{colorVidPromPropertyCounts_Countries}{RGB}{245,244,244}
\definecolor{colorShareGPT4VideoPropertyCounts_Countries}{RGB}{239,221,175}
\definecolor{colorOpenVid-1MPropertyCounts_Countries}{RGB}{239,221,175}
\definecolor{colorFineVideoPropertyCounts_Countries}{RGB}{239,221,175}
\definecolor{colorDisneyVid.Gen.PropertyCounts_Countries}{RGB}{239,221,175}
\definecolor{colorKineticsPropertyCounts_Countries}{RGB}{239,221,175}
\definecolor{colorEgo4DPropertyCounts_Countries}{RGB}{239,221,175}
\definecolor{colorMPIIPropertyCounts_Countries}{RGB}{239,221,175}
\definecolor{colorProject-AriaPropertyCounts_Countries}{RGB}{239,221,175}
\definecolor{colorAvaPropertyCounts_Countries}{RGB}{239,221,175}
\definecolor{colorLSMDCPropertyCounts_Countries}{RGB}{179,226,219}
\definecolor{colorHOLLYWOOD2PropertyCounts_Creators}{RGB}{204,235,230}
\definecolor{colorCollectivePropertyCounts_Creators}{RGB}{204,235,230}
\definecolor{colorHMDBPropertyCounts_Creators}{RGB}{193,231,226}
\definecolor{colorUCF101PropertyCounts_Creators}{RGB}{204,235,230}
\definecolor{colorYouCookPropertyCounts_Creators}{RGB}{204,235,230}
\definecolor{color50SaladsPropertyCounts_Creators}{RGB}{204,235,230}
\definecolor{colorStoryGraphsPropertyCounts_Creators}{RGB}{204,235,230}
\definecolor{colorHollywoodExt.PropertyCounts_Creators}{RGB}{204,235,230}
\definecolor{colorBreakfastPropertyCounts_Creators}{RGB}{199,234,229}
\definecolor{colorSports-1MPropertyCounts_Creators}{RGB}{204,235,230}
\definecolor{colorTHUMOSPropertyCounts_Creators}{RGB}{190,230,224}
\definecolor{colorVideoStoryPropertyCounts_Creators}{RGB}{204,235,230}
\definecolor{colorSumMePropertyCounts_Creators}{RGB}{193,231,226}
\definecolor{colorTVSumPropertyCounts_Creators}{RGB}{204,235,230}
\definecolor{colorVolleyballPropertyCounts_Creators}{RGB}{204,235,230}
\definecolor{colorActivityNetPropertyCounts_Creators}{RGB}{199,234,229}
\definecolor{colorMovieQAPropertyCounts_Creators}{RGB}{193,231,226}
\definecolor{colorMarsPropertyCounts_Creators}{RGB}{190,230,224}
\definecolor{colorNTURGB+DPropertyCounts_Creators}{RGB}{204,235,230}
\definecolor{colorMSR-VTTPropertyCounts_Creators}{RGB}{204,235,230}
\definecolor{colorCharadesPropertyCounts_Creators}{RGB}{190,230,224}
\definecolor{colorVTWPropertyCounts_Creators}{RGB}{199,234,229}
\definecolor{colorYoutube-8MPropertyCounts_Creators}{RGB}{204,235,230}
\definecolor{colorNarratedInstr.Vid.PropertyCounts_Creators}{RGB}{190,230,224}
\definecolor{colorTGIFPropertyCounts_Creators}{RGB}{193,231,226}
\definecolor{colorMultiTHUMOSPropertyCounts_Creators}{RGB}{193,231,226}
\definecolor{colorImageNet-VidPropertyCounts_Creators}{RGB}{204,235,230}
\definecolor{colorPKU-MMDPropertyCounts_Creators}{RGB}{199,234,229}
\definecolor{color20BN-SOMETHINGPropertyCounts_Creators}{RGB}{204,235,230}
\definecolor{colorYouCook2PropertyCounts_Creators}{RGB}{199,234,229}
\definecolor{colorVoxCelebPropertyCounts_Creators}{RGB}{204,235,230}
\definecolor{colorDavisPropertyCounts_Creators}{RGB}{199,234,229}
\definecolor{colorQFVSPropertyCounts_Creators}{RGB}{199,234,229}
\definecolor{colorDiDeMoPropertyCounts_Creators}{RGB}{204,235,230}
\definecolor{colorSOAPropertyCounts_Creators}{RGB}{204,235,230}
\definecolor{colorCharades-EgoPropertyCounts_Creators}{RGB}{204,235,230}
\definecolor{colorEPIC-KITCHENSPropertyCounts_Creators}{RGB}{193,231,226}
\definecolor{colorMovieGraphsPropertyCounts_Creators}{RGB}{193,231,226}
\definecolor{colorHow2PropertyCounts_Creators}{RGB}{204,235,230}
\definecolor{colorVLOGPropertyCounts_Creators}{RGB}{204,235,230}
\definecolor{colorVaTeXPropertyCounts_Creators}{RGB}{199,234,229}
\definecolor{color20BN-jesterPropertyCounts_Creators}{RGB}{204,235,230}
\definecolor{colorHowTo100MPropertyCounts_Creators}{RGB}{190,230,224}
\definecolor{colorCOINPropertyCounts_Creators}{RGB}{199,234,229}
\definecolor{colorMMActPropertyCounts_Creators}{RGB}{199,234,229}
\definecolor{colorHACSPropertyCounts_Creators}{RGB}{193,231,226}
\definecolor{colorCrossTaskPropertyCounts_Creators}{RGB}{187,229,223}
\definecolor{colorMomentsinTimePropertyCounts_Creators}{RGB}{204,235,230}
\definecolor{colorTRECVidPropertyCounts_Creators}{RGB}{204,235,230}
\definecolor{colorMSAPropertyCounts_Creators}{RGB}{199,234,229}
\definecolor{colorToyotaSmarthomePropertyCounts_Creators}{RGB}{204,235,230}
\definecolor{colorTITANPropertyCounts_Creators}{RGB}{204,235,230}
\definecolor{colorVIOLINPropertyCounts_Creators}{RGB}{204,235,230}
\definecolor{colorRareActPropertyCounts_Creators}{RGB}{187,229,223}
\definecolor{colorTinyVIRATPropertyCounts_Creators}{RGB}{204,235,230}
\definecolor{color100DOHPropertyCounts_Creators}{RGB}{199,234,229}
\definecolor{colorOops!PropertyCounts_Creators}{RGB}{204,235,230}
\definecolor{colorOmniSource-WebPropertyCounts_Creators}{RGB}{204,235,230}
\definecolor{colorCondensedMoviesPropertyCounts_Creators}{RGB}{204,235,230}
\definecolor{colorMovieScenesPropertyCounts_Creators}{RGB}{199,234,229}
\definecolor{colorEEVPropertyCounts_Creators}{RGB}{199,234,229}
\definecolor{colorMovie-NetPropertyCounts_Creators}{RGB}{204,235,230}
\definecolor{colorFineGymPropertyCounts_Creators}{RGB}{204,235,230}
\definecolor{colorHAA500PropertyCounts_Creators}{RGB}{190,230,224}
\definecolor{colorLEMMAPropertyCounts_Creators}{RGB}{204,235,230}
\definecolor{colorHVUPropertyCounts_Creators}{RGB}{187,229,223}
\definecolor{colorApesPropertyCounts_Creators}{RGB}{193,231,226}
\definecolor{colorWebVidPropertyCounts_Creators}{RGB}{199,234,229}
\definecolor{colorVideoLTPropertyCounts_Creators}{RGB}{190,230,224}
\definecolor{colorHOMAGEPropertyCounts_Creators}{RGB}{199,234,229}
\definecolor{colorUAV-HumanPropertyCounts_Creators}{RGB}{199,234,229}
\definecolor{colorHD-VILA-100MPropertyCounts_Creators}{RGB}{204,235,230}
\definecolor{colorM-MiTPropertyCounts_Creators}{RGB}{204,235,230}
\definecolor{colorMimeticsPropertyCounts_Creators}{RGB}{204,235,230}
\definecolor{colorSpokenMomentsPropertyCounts_Creators}{RGB}{193,231,226}
\definecolor{colorQuerYDPropertyCounts_Creators}{RGB}{204,235,230}
\definecolor{colorMADPropertyCounts_Creators}{RGB}{204,235,230}
\definecolor{colorFERV39kPropertyCounts_Creators}{RGB}{204,235,230}
\definecolor{colorCDADPropertyCounts_Creators}{RGB}{199,234,229}
\definecolor{colorMVBenchPropertyCounts_Creators}{RGB}{187,229,223}
\definecolor{colorVidPromPropertyCounts_Creators}{RGB}{199,234,229}
\definecolor{colorShareGPT4VideoPropertyCounts_Creators}{RGB}{190,230,224}
\definecolor{colorOpenVid-1MPropertyCounts_Creators}{RGB}{193,231,226}
\definecolor{colorFineVideoPropertyCounts_Creators}{RGB}{204,235,230}
\definecolor{colorDisneyVid.Gen.PropertyCounts_Creators}{RGB}{240,223,178}
\definecolor{colorKineticsPropertyCounts_Creators}{RGB}{204,235,230}
\definecolor{colorEgo4DPropertyCounts_Creators}{RGB}{199,234,229}
\definecolor{colorMPIIPropertyCounts_Creators}{RGB}{199,234,229}
\definecolor{colorProject-AriaPropertyCounts_Creators}{RGB}{204,235,230}
\definecolor{colorAvaPropertyCounts_Creators}{RGB}{204,235,230}
\definecolor{colorLSMDCPropertyCounts_Creators}{RGB}{179,226,219}
\definecolor{colorHOLLYWOOD2PropertyCounts_Sources}{RGB}{240,223,178}
\definecolor{colorCollectivePropertyCounts_Sources}{RGB}{240,223,178}
\definecolor{colorHMDBPropertyCounts_Sources}{RGB}{227,240,239}
\definecolor{colorUCF101PropertyCounts_Sources}{RGB}{240,223,178}
\definecolor{colorYouCookPropertyCounts_Sources}{RGB}{240,223,178}
\definecolor{color50SaladsPropertyCounts_Sources}{RGB}{240,223,178}
\definecolor{colorStoryGraphsPropertyCounts_Sources}{RGB}{240,223,178}
\definecolor{colorHollywoodExt.PropertyCounts_Sources}{RGB}{240,223,178}
\definecolor{colorBreakfastPropertyCounts_Sources}{RGB}{240,223,178}
\definecolor{colorSports-1MPropertyCounts_Sources}{RGB}{240,223,178}
\definecolor{colorTHUMOSPropertyCounts_Sources}{RGB}{240,223,178}
\definecolor{colorVideoStoryPropertyCounts_Sources}{RGB}{240,223,178}
\definecolor{colorSumMePropertyCounts_Sources}{RGB}{240,223,178}
\definecolor{colorTVSumPropertyCounts_Sources}{RGB}{240,223,178}
\definecolor{colorVolleyballPropertyCounts_Sources}{RGB}{240,223,178}
\definecolor{colorActivityNetPropertyCounts_Sources}{RGB}{240,223,178}
\definecolor{colorMovieQAPropertyCounts_Sources}{RGB}{240,223,178}
\definecolor{colorMarsPropertyCounts_Sources}{RGB}{240,223,178}
\definecolor{colorNTURGB+DPropertyCounts_Sources}{RGB}{240,223,178}
\definecolor{colorMSR-VTTPropertyCounts_Sources}{RGB}{240,223,178}
\definecolor{colorCharadesPropertyCounts_Sources}{RGB}{240,223,178}
\definecolor{colorVTWPropertyCounts_Sources}{RGB}{240,223,178}
\definecolor{colorYoutube-8MPropertyCounts_Sources}{RGB}{240,223,178}
\definecolor{colorNarratedInstr.Vid.PropertyCounts_Sources}{RGB}{240,223,178}
\definecolor{colorTGIFPropertyCounts_Sources}{RGB}{240,223,178}
\definecolor{colorMultiTHUMOSPropertyCounts_Sources}{RGB}{240,223,178}
\definecolor{colorImageNet-VidPropertyCounts_Sources}{RGB}{240,223,178}
\definecolor{colorPKU-MMDPropertyCounts_Sources}{RGB}{240,223,178}
\definecolor{color20BN-SOMETHINGPropertyCounts_Sources}{RGB}{240,223,178}
\definecolor{colorYouCook2PropertyCounts_Sources}{RGB}{240,223,178}
\definecolor{colorVoxCelebPropertyCounts_Sources}{RGB}{240,223,178}
\definecolor{colorDavisPropertyCounts_Sources}{RGB}{240,223,178}
\definecolor{colorQFVSPropertyCounts_Sources}{RGB}{240,223,178}
\definecolor{colorDiDeMoPropertyCounts_Sources}{RGB}{240,223,178}
\definecolor{colorSOAPropertyCounts_Sources}{RGB}{240,223,178}
\definecolor{colorCharades-EgoPropertyCounts_Sources}{RGB}{240,223,178}
\definecolor{colorEPIC-KITCHENSPropertyCounts_Sources}{RGB}{240,223,178}
\definecolor{colorMovieGraphsPropertyCounts_Sources}{RGB}{240,223,178}
\definecolor{colorHow2PropertyCounts_Sources}{RGB}{240,223,178}
\definecolor{colorVLOGPropertyCounts_Sources}{RGB}{240,223,178}
\definecolor{colorVaTeXPropertyCounts_Sources}{RGB}{240,223,178}
\definecolor{color20BN-jesterPropertyCounts_Sources}{RGB}{240,223,178}
\definecolor{colorHowTo100MPropertyCounts_Sources}{RGB}{240,223,178}
\definecolor{colorCOINPropertyCounts_Sources}{RGB}{240,223,178}
\definecolor{colorMMActPropertyCounts_Sources}{RGB}{240,223,178}
\definecolor{colorHACSPropertyCounts_Sources}{RGB}{240,223,178}
\definecolor{colorCrossTaskPropertyCounts_Sources}{RGB}{240,223,178}
\definecolor{colorMomentsinTimePropertyCounts_Sources}{RGB}{187,229,223}
\definecolor{colorTRECVidPropertyCounts_Sources}{RGB}{245,237,216}
\definecolor{colorMSAPropertyCounts_Sources}{RGB}{240,223,178}
\definecolor{colorToyotaSmarthomePropertyCounts_Sources}{RGB}{240,223,178}
\definecolor{colorTITANPropertyCounts_Sources}{RGB}{240,223,178}
\definecolor{colorVIOLINPropertyCounts_Sources}{RGB}{240,223,178}
\definecolor{colorRareActPropertyCounts_Sources}{RGB}{240,223,178}
\definecolor{colorTinyVIRATPropertyCounts_Sources}{RGB}{240,223,178}
\definecolor{color100DOHPropertyCounts_Sources}{RGB}{240,223,178}
\definecolor{colorOops!PropertyCounts_Sources}{RGB}{240,223,178}
\definecolor{colorOmniSource-WebPropertyCounts_Sources}{RGB}{245,243,238}
\definecolor{colorCondensedMoviesPropertyCounts_Sources}{RGB}{240,223,178}
\definecolor{colorMovieScenesPropertyCounts_Sources}{RGB}{240,223,178}
\definecolor{colorEEVPropertyCounts_Sources}{RGB}{240,223,178}
\definecolor{colorMovie-NetPropertyCounts_Sources}{RGB}{240,223,178}
\definecolor{colorFineGymPropertyCounts_Sources}{RGB}{240,223,178}
\definecolor{colorHAA500PropertyCounts_Sources}{RGB}{240,223,178}
\definecolor{colorLEMMAPropertyCounts_Sources}{RGB}{245,237,216}
\definecolor{colorHVUPropertyCounts_Sources}{RGB}{240,223,178}
\definecolor{colorApesPropertyCounts_Sources}{RGB}{240,223,178}
\definecolor{colorWebVidPropertyCounts_Sources}{RGB}{240,223,178}
\definecolor{colorVideoLTPropertyCounts_Sources}{RGB}{240,223,178}
\definecolor{colorHOMAGEPropertyCounts_Sources}{RGB}{240,223,178}
\definecolor{colorUAV-HumanPropertyCounts_Sources}{RGB}{240,223,178}
\definecolor{colorHD-VILA-100MPropertyCounts_Sources}{RGB}{240,223,178}
\definecolor{colorM-MiTPropertyCounts_Sources}{RGB}{245,237,216}
\definecolor{colorMimeticsPropertyCounts_Sources}{RGB}{240,223,178}
\definecolor{colorSpokenMomentsPropertyCounts_Sources}{RGB}{187,229,223}
\definecolor{colorQuerYDPropertyCounts_Sources}{RGB}{245,237,216}
\definecolor{colorMADPropertyCounts_Sources}{RGB}{240,223,178}
\definecolor{colorFERV39kPropertyCounts_Sources}{RGB}{240,223,178}
\definecolor{colorCDADPropertyCounts_Sources}{RGB}{240,223,178}
\definecolor{colorMVBenchPropertyCounts_Sources}{RGB}{179,226,219}
\definecolor{colorVidPromPropertyCounts_Sources}{RGB}{227,240,239}
\definecolor{colorShareGPT4VideoPropertyCounts_Sources}{RGB}{227,240,239}
\definecolor{colorOpenVid-1MPropertyCounts_Sources}{RGB}{227,240,239}
\definecolor{colorFineVideoPropertyCounts_Sources}{RGB}{240,223,178}
\definecolor{colorDisneyVid.Gen.PropertyCounts_Sources}{RGB}{245,237,216}
\definecolor{colorKineticsPropertyCounts_Sources}{RGB}{245,237,216}
\definecolor{colorEgo4DPropertyCounts_Sources}{RGB}{240,223,178}
\definecolor{colorMPIIPropertyCounts_Sources}{RGB}{245,237,216}
\definecolor{colorProject-AriaPropertyCounts_Sources}{RGB}{240,223,178}
\definecolor{colorAvaPropertyCounts_Sources}{RGB}{245,237,216}
\definecolor{colorLSMDCPropertyCounts_Sources}{RGB}{240,223,178}

\begin{longtable}{lc|ccccc|cc}
\caption[\textbf{Video collections and properties}]{\textbf{Video collections and properties}. Collection properties include numbers of hours of video, datasets, creator institutions, countries of creator institutions, and data sources. The \textsc{Use} column indicates whether a collection includes data freely usable even for commercial purposes (\protect\CommercialDataCircle), data usable only for noncommercial purposes or academic research (\protect\NCDataCircle) and data whose license status is not specified precisely enough to allow us to determine commercial use permissions (\protect\UnspecifiedDataCircle). Note that each collection may have different datasets with one, two, or all three of these statuses. Finally, the \textsc{Avail} column indicates whether a dataset is available online (\greencheck) or has been taken down, usually for legal reasons (\redcross). Datasets are sorted chronologically to highlight trends over time.} \label{tab:collections-video} \\
\toprule
\textsc{Collection} & \textsc{} & \multicolumn{5}{c}{\textsc{Property Counts}} & \multicolumn{2}{c}{\textsc{Permissions}} \\
 & \textsc{\thead{Year}} & \textsc{\thead{Hours}} & \textsc{\thead{Datasets}} & \textsc{\thead{Countries}} & \textsc{\thead{Creators}} & \textsc{\thead{Sources}} & \textsc{\thead{Use}} & \textsc{\thead{Avail}} \\
\midrule
\endfirsthead
\caption[]{\textbf{Video collections and properties}.} \\
\toprule
\textsc{Collection} & \textsc{} & \multicolumn{5}{c}{\textsc{Property Counts}} & \multicolumn{2}{c}{\textsc{Permissions}} \\
 & \textsc{\thead{Year}} & \textsc{\thead{Hours}} & \textsc{\thead{Datasets}} & \textsc{\thead{Countries}} & \textsc{\thead{Creators}} & \textsc{\thead{Sources}} & \textsc{\thead{Use}} & \textsc{\thead{Avail}} \\
\midrule
\endhead
\midrule
\multicolumn{9}{r}{Continued on next page} \\
\midrule
\endfoot
\bottomrule
\endlastfoot
HOLLYWOOD2 & 2009 & \cellcolor{colorQFVSPropertyCounts_Hours}{20} & \cellcolor{colorDisneyVid.Gen.PropertyCounts_Datasets}{1} & \cellcolor{colorAvaPropertyCounts_Countries}{1} & \cellcolor{colorAvaPropertyCounts_Creators}{1} & \cellcolor{colorLSMDCPropertyCounts_Sources}{1} & \TransparentCircle \UnspecifiedDataCircle \TransparentCircle & \greencheck \\
Collective & 2009 & - & \cellcolor{colorDisneyVid.Gen.PropertyCounts_Datasets}{1} & \cellcolor{colorAvaPropertyCounts_Countries}{1} & \cellcolor{colorAvaPropertyCounts_Creators}{1} & \cellcolor{colorLSMDCPropertyCounts_Sources}{1} & \TransparentCircle \UnspecifiedDataCircle \TransparentCircle & \greencheck \\
HMDB & 2011 & \cellcolor{colorHMDBPropertyCounts_Hours}{7k} & \cellcolor{colorDisneyVid.Gen.PropertyCounts_Datasets}{1} & \cellcolor{colorVidPromPropertyCounts_Countries}{2} & \cellcolor{colorOpenVid-1MPropertyCounts_Creators}{3} & \cellcolor{colorOpenVid-1MPropertyCounts_Sources}{5} & \CommercialDataCircle \TransparentCircle \TransparentCircle & \greencheck \\
UCF101 & 2012 & \cellcolor{colorUCF101PropertyCounts_Hours}{26} & \cellcolor{colorDisneyVid.Gen.PropertyCounts_Datasets}{1} & \cellcolor{colorAvaPropertyCounts_Countries}{1} & \cellcolor{colorAvaPropertyCounts_Creators}{1} & \cellcolor{colorLSMDCPropertyCounts_Sources}{1} & \TransparentCircle \UnspecifiedDataCircle \TransparentCircle & \greencheck \\
YouCook & 2013 & \cellcolor{colorTRECVidPropertyCounts_Hours}{1k} & \cellcolor{colorDisneyVid.Gen.PropertyCounts_Datasets}{1} & \cellcolor{colorAvaPropertyCounts_Countries}{1} & \cellcolor{colorAvaPropertyCounts_Creators}{1} & \cellcolor{colorLSMDCPropertyCounts_Sources}{1} & \TransparentCircle \UnspecifiedDataCircle \TransparentCircle & \greencheck \\
\end{longtable}
\section{Contributions}
\label{app:contributions}
Here we break down contributions to this work. Contributors are listed alphabetically, except for team leads who are placed first.

\begin{itemize}

    \item \textbf{Text Datasets} \; Shayne Longpre (lead), Jad Kabbara (lead),  Ahmad Anis, Deividas Mataciunas, Diganta Misra, Emad Alghamdi, Enrico Shippole, Jianguo Zhang, Kun Qian, Lester Miranda, Manan Dey, Minnie Liang, Mohammed Hamdy, Nayan Saxena, Niklas Muennighoff, Naana Obeng-Marnu, Robert Mahari, Seonghyeon Ye, Seungone Kim, Shayne Longpre, Shrestha Mohanty, Vipul Gupta, Vivek Sharma, Vu Minh Chien, William Brannon, Xuhui Zhou, Yizhi Li, An Dinh, Caroline Chitongo, Christopher Klamm, Da Yin, Damien Sileo, Ariel Lee

    \item \textbf{Reviewing Text Dataset Metadata} \; Jad Kabbara (lead), Shayne Longpre (lead), Robert Mahari, Damien Sileo, Niklas Muennighoff, William Brannon,
    
    \item \textbf{Data Explorer Features} \; Shayne Longpre (lead), Christopher Klamm, Vu Minh Chien, 

    \item \textbf{Speech Datasets} Nikhil Singh (lead), Manuel Cherep (lead), An Dinh, Minnie Liang, Shrestha Mohanty

    \item \textbf{Video Datasets} \; Kush Tiwary (lead), Joanna Materzynska (lead), Vivek Sharma, Shayne Longpre, Robert Mahari, Jad Kabbara, William Brannon, Tobin South, Shrestha Mohanty, Nikhil Singh, Manuel Cherep
    
    \item \textbf{Data Analysis} \; Shayne Longpre (lead), Nikhil Singh (lead), Manuel Cherep (lead), Kush Tiwary (lead), Joanna Materzynska (lead), Naana Obeng-Marnu (lead), William Brannon (lead), 

    \item \textbf{Writing} \; Shayne Longpre (lead), Jad Kabbara (lead), Nikhil Singh, Manuel Cherep, Kush Tiwary, Joanna Materzynska, Robert Mahari
    
    \item \textbf{Legal Analysis} \; Robert Mahari (lead), Luis Villa
    
    \item \textbf{Visualizations \& Visual Data Analysis} \; Nikhil Singh (lead), Manuel Cherep (lead), Kush Tiwary (lead), Joanna Materzynska (lead), Naana Obeng-Marnu (lead), William Brannon (lead), Shayne Longpre (lead), Ariel Lee, Hamidah Oderinwale, Campbell Lund

    \item \textbf{Senior Advisors} \; Stella Biderman, Sara Hooker, Jad Kabbara, Sandy Pentland, Luis Villa, Caiming Xiong
\end{itemize}
\captionsetup{width=0.8\textwidth}

\section{Attribution Card}
\label{sec:attribution-card}

\noindent Here we provide detailed information about the licenses of each data collection and its constituent datasets, and cite all of the papers (455 in all) which introduced datasets we consider. Text datasets are laid out in \autoref{tab:refs-licenses-text}, audio datasets in \autoref{tab:refs-licenses-audio}, and video datasets in \autoref{tab:refs-licenses-video}. Because of the large number of references, we include a second bibliography after the tables (named `Attribution Card References'), with numbered citations in this section referring to that second bibliography.

\noindent\rule{\textwidth}{0.4pt}

\begin{refsection}
    \begingroup
        \renewcommand*{\arraystretch}{1.2}
        \begin{longtable}{p{5cm}|p{5cm}|p{5cm}}
\caption[\textbf{References and licenses: alignment tuning (text}]{\textbf{References and licenses for alignment-tuning (text)} dataset collections presented in this paper. Collections containing material under more than three distinct licenses are marked as having ``Various'' licenses, and we refer readers to our raw data for the full details. Datasets are sorted alphabetically for ease of dataset lookup.} \label{tab:refs-licenses-text} \\
\toprule
Collection & Licenses & Cite \\
\midrule
\endfirsthead
\caption[]{\textbf{References and licenses for alignment-tuning (text)} dataset collections presented in this paper. Collections containing material under more than three distinct licenses are marked as having ``Various'' licenses, and we refer readers to our raw data for the full details. Datasets are sorted alphabetically for ease of dataset lookup.} \\
\toprule
Collection & Licenses & Cite \\
\midrule
\endhead
\midrule
\multicolumn{3}{r}{Continued on next page} \\
\midrule
\endfoot
\bottomrule
\endlastfoot
10k Prompt Ranked & Unspecified & -- \\
AgentInstruct & Unspecified, CC BY 4.0, MIT License & \autocite{shridharALFWorldAligningText2021,yaoWebShopScalableRealWorld2023,liuAgentBenchEvaluatingLLMs2023,zengAgentTuningEnablingGeneralized2023,dengMind2WebGeneralistAgent2023} \\
Aya & Apache License 2.0 & \autocite{singhAyaDatasetOpenAccess2024} \\
Bactrian-X & CC BY-SA 3.0, CC BY-NC 4.0 & \autocite{liBactrianXMultilingualReplicable2023} \\
COBRA Frames & BigScience OpenRAIL-M & \autocite{zhouCOBRAFramesContextual2023} \\
COIG & Various & \autocite{zhangChineseOpenInstruction2023,baiCOIGCQIAQualityAll2024} \\
Capybara & Various & -- \\
ChatDoctor & Unspecified & \autocite{liChatDoctorMedicalChat2023} \\
ChatbotArena & CC BY 4.0, CC BY-NC 4.0 & \autocite{zhengJudgingLLMasaJudgeMTBench2023} \\
Cidar & CC BY-NC 4.0 & \autocite{alyafeaiCIDARCulturallyRelevant2024} \\
CollectiveCognition & MIT License & -- \\
Conifer & Apache License 2.0 & \autocite{sunConiferImprovingComplex2024} \\
Deita 10K & Apache License 2.0, CC BY-NC 4.0 & \autocite{liuWhatMakesGood2024} \\
DialogStudio & Various & \autocite{chenActionBasedConversationsDataset2021,weiAirDialogueEnvironmentGoalOriented2018,linBiToDBilingualMultiDomain2021,chawlaCaSiNoCorpusCampsite2021,heDecouplingStrategyGeneration2018,mrksicNeuralBeliefTracker2017,qianDatabaseSearchResults2022,liuDuRecDialBilingualParallel2021,elasriFramesCorpusAdding2017,quanGECOREndtoEndGenerative2019,chenSemanticallyConditionedDialog2019,chenKETODKnowledgeEnrichedTaskOriented2022,ericKeyValueRetrievalNetworks2017,zangMultiWOZDialogueDataset2020,shalyminovFewShotDialogueGeneration2019,martinMuDoCoCorpusMultidomain2020,peskovMultiDomainGoalOrientedDialogues2019,ericMultiWOZConsolidatedMultiDomain2019,moonOpenDialKGExplainableConversational2019,rastogiScalableMultidomainConversational2020,mosigSTARSchemaGuidedDialog2020,chiuSalesBotTransitioningChitChat2022,shahBuildingConversationalAgent2018,byrneTaskmaster1RealisticDiverse2019,mrksicFullyStatisticalNeural2018,shangUnsupervisedAbstractiveMeeting2018,rameshkumarStorytellingDialogueCritical2020,fabbriConvoSummConversationSummarization2021,chenDialogSumRealLifeScenario2021,mukherjeeECTSumNewBenchmark2022,shangUnsupervisedAbstractiveMeeting2018,zhuMediaSumLargescaleMedia2021,zhongQMSumNewBenchmark2021,gliwaSAMSumCorpusHumanannotated2019,chenSummScreenDatasetAbstractive2022,feigenblatTWEETSUMMDialogSummarization2021,liEndtoEndTrainableNonCollaborative2019,dinanSecondConversationalIntelligence2019,rashkinEmpatheticOpendomainConversation2019,baiTrainingHelpfulHarmless2022,chenPLACESPromptingLanguage2023,kimProsocialDialogProsocialBackbone2022,myersConversationalScaffoldingAnalogybased2020,reddyCoQAConversationalQuestion2019,yuCoSQLConversationalTexttoSQL2019,talmorWebKnowledgeBaseAnswering2018,nanDARTOpenDomainStructured2021,nanFeTaQAFreeformTable2022,guThreeLevelsGeneralization2021,chenHybridQADatasetMultiHop2020,guptaMMQAMultidomainMultilingual2018,liMTOPComprehensiveMultilingual2021,talmorMultiModalQAComplexQuestion2021,yuSParCCrossDomainSemantic2019,iyyerSearchbasedNeuralStructured2017,yuSpiderLargeScaleHumanLabeled2019,parikhToTToControlledTableToText2020,yihValueSemanticParse2016,zhongSeq2SQLGeneratingStructured2017,pasupatCompositionalSemanticParsing2015,komeiliInternetAugmentedDialogueGeneration2022,dinanWizardWikipediaKnowledgePowered2019,hemphillATISSpokenLanguage1990,casanuevaEfficientIntentDetection2020,zhangArePretrainedTransformers2022,larsonEvaluationDatasetIntent2019,rastogiScalableMultidomainConversational2020,liuBenchmarkingNaturalLanguage2019,liuAsgardPortableArchitecture2013,coopeSpanConveRTFewshotSpan2020,couckeSnipsVoicePlatform2018,guptaSemanticParsingTask2018} \\
Dynosaur & Various & \autocite{adlakhaTopiOCQAOpendomainConversational2022,agarwalKnowledgeGraphBased2021,akyurekTracingFactualKnowledge2022,aminiMathQAInterpretableMath2019,ardanuyLivingMachinesStudy2020,austinProgramSynthesisLarge2021,azerbayevProofNetAutoformalizingFormally2023,baiTrainingHelpfulHarmless2022,bajajMSMARCOHuman2018,balakrishnanConstrainedDecodingNeural2019,bartoloBeatAIInvestigating2020,biskPIQAReasoningPhysical2019,boratkoProtoQAQuestionAnswering2020,bothaLearningSplitRephrase2018,boudinRedefiningAbsentKeyphrases2021,bravoExtractionRelationsGenes2015,brownLanguageModelsAre2020,byrneTaskmaster1RealisticDiverse2019,caoControllableOpenendedQuestion2021,caoKQAProDataset2022,casanuevaEfficientIntentDetection2020,cetoliNamedEntityDisambiguation2019,chalkidisLargeScaleMultiLabelText2019,chalkidisNeuralLegalJudgment2019,chalkidisRegulatoryComplianceDoc2Doc2021,chanFewshotAdaptationWorks2022,chapuisHierarchicalPretrainingSequence2021,chenFindingFriendsFlipping2020,chengHiTabHierarchicalTable2022,chouldechovaFairPredictionDisparate2017,christmannLookYouHop2019,clarkBoolQExploringSurprising2019,clarkThinkYouHave2018,cobbeTrainingVerifiersSolve2021,couckeSnipsVoicePlatform2018,dankersCanTransformerBe2022,dasigiQuorefReadingComprehension2019,devarajParagraphlevelSimplificationMedical2021,deyoungMS2MultiDocumentSummarization2021,diggelmannCLIMATEFEVERDatasetVerification2021,emelinMoralStoriesSituated2020,fabbriMultiNewsLargeScaleMultiDocument2019,faruquiIdentifyingWellformedNatural2018,fengMultiDoc2DialModelingDialogues2021,gallinaKPTimesLargeScaleDataset2019,ganesanOpinosisGraphBased2010,gazzolaPredicaoComplexidadeTextual2019,georgeConversationalImplicaturesEnglish2019,gevaDiscoFuseLargeScaleDataset2019,gliwaSAMSumCorpusHumanannotated2019,gorrellRumourEval2019Determining2018,guDomainSpecificLanguageModel2022,guptaDisflQABenchmarkDataset2021,haNeuralRepresentationSketch2017,haagsmaMAGPIELargeCorpus2020,hazoomTextSQLWildNaturallyOccurring2021,hendersonPileLawLearning2022,hendrycksCUADExpertAnnotatedNLP2021,huangCosmosQAMachine2019,huangEfficientAttentionsLong2021,huangTisThyName2022,irwinZINC20AFreeUltralargeScale2020,ivgiEfficientLongTextUnderstanding2022,iyerLearningNeuralSemantic2017,jiangHoVerDatasetManyHop2020,jiangNeuralCRFModel2021,jinPubMedQADatasetBiomedical2019,joshiTriviaQALargeScale2017,juraskaViGGOVideoGame2019,jurczykSelQANewBenchmark2016,kanadeLearningEvaluatingContextual2020,kaushikLearningDifferenceThat2020,khotQASCDatasetQuestion2020,khotSciTaiLTextualEntailment2018,kimSemanticSentenceMatching2018,kornilovaBillSumCorpusAutomatic2019,kuryChiaLargeAnnotated2020,laiRACELargescaleReAding2017,lakeGeneralizationSystematicityCompositional2018,lebretNeuralTextGeneration2016,lewisDealNoDeal2017,liCompetitionLevelCodeGeneration2022,liNeuralCodeSearch2019,linBirdsHaveFour2020,linCommonGenConstrainedText2020,linReasoningParagraphEffects2019,linTruthfulQAMeasuringHow2022,lingProgramInductionRationale2017,liuBenchmarkingNaturalLanguage2019,louisIdRatherJust2020,loweUbuntuDialogueCorpus2016,maloGoodDebtBad2013,martinEventRepresentationsAutomated2018,merityPointerSentinelMixture2016,mihaylovCanSuitArmor2018,mishraLilaUnifiedBenchmark2023,monizMultiSourceSocialFeedback2018,mostafazadehGLUCOSEGeneraLizedCOntextualized2020,nanDARTOpenDomainStructured2021,narayanDontGiveMe2018,nguyenAdvancedSemanticsCommonsense2021,nieAdversarialNLINew2020,novikovaE2EDatasetNew2017,paikWorldOctopusHow2021,pakhomovSemanticSimilarityRelatedness2010,pangSeeingStarsExploiting2005,pavlichenkoCrowdSpeechVoxDIYBenchmark2021,pedersenMeasuresSemanticSimilarity2007,perez-beltrachiniGeneratingSummariesTopic2019,petroniLanguageModelsKnowledge2019,phamPiCPhraseContextDataset2023,rajaniExplainYourselfLeveraging2019,rajpurkarSQuAD100000Questions2016,rameshkumarStorytellingDialogueCritical2020,rashkinEmpatheticOpendomainConversation2019,rastogiScalableMultidomainConversational2020,royerXGANUnsupervisedImageImage2018,rushNeuralAttentionModel2015,rustLanguageModellingPixels2023,saeidiInterpretationNaturalLanguage2018,sahaDuoRCComplexLanguage2018,sakaguchiWinoGrandeAdversarialWinograd2019,sanhMultitaskPromptedTraining2022,sapSocialIQACommonsenseReasoning2019,schulzBiomedicalConceptRelatedness2020,seeGetPointSummarization2017,sharmaBIGPATENTLargeScaleDataset2019,shribergCanProsodyAid1998,sileoProbingNeuralLanguage2023,soleimaniNLQuADNonFactoidLong2021,stolckeDialogueActModeling2000,tafjordQuaRTzOpenDomainDataset2019,tafjordQuaRelDatasetModels2018,talmorCommonsenseQAQuestionAnswering2019,tandonWIQADatasetWhat2019,tangRapidlyBootstrappingQuestion2020,thawaniNumeracyEnhancesLiteracy2021,thorneFEVERLargescaleDataset2018,tylecekSpatialPatternTemplates2013,ullrichCsFEVERCTKFactsAcquiring2023,ushioGenerativeLanguageModels2023,wangAdversarialGLUEMultiTask2022,wangDetectingCodeClones2020,wangGLUEMultiTaskBenchmark2019,wangSelfInstructAligningLanguage2023,warstadtBLiMPBenchmarkLinguistic2023,welblConstructingDatasetsMultihop2018,welblCrowdsourcingMultipleChoice2017,wellerLearningTaskDescriptions2020,westonAICompleteQuestionAnswering2015,williamsANLIzingAdversarialNatural2020,wuMoleculeNetBenchmarkMolecular2018,wuMoleculeNetBenchmarkMolecular2018,xiongTWEETQASocialMedia2019,yangHotpotQADatasetDiverse2018,yuSpiderLargeScaleHumanLabeled2019,zellersHellaSwagCanMachine2019,zhangCharacterlevelConvolutionalNetworks2016,zhangFengshenbang10Being2023,zhangPAWSParaphraseAdversaries2019,zhouDevignEffectiveVulnerability2019,zhouDocPromptingGeneratingCode2023,zhuXLCoSTBenchmarkDataset2022} \\
EverythingLM & MIT License & -- \\
ExpertQA & MIT License & \autocite{malaviyaExpertQAExpertCuratedQuestions2024} \\
Feedback Coll. & MIT License & \autocite{kimPrometheusInducingFinegrained2024} \\
Glaive Code Asst. & Apache License 2.0 & -- \\
Gretel Text-to-SQL & Apache License 2.0 & -- \\
HelpSteer & CC BY 4.0 & \autocite{wangHelpSteerMultiattributeHelpfulness2023} \\
Indic-Instr. & Various & \autocite{galaAiravataIntroducingHindi2024} \\
InstAr & Various & \autocite{huXTREMEMassivelyMultilingual2020,eineaSANADSinglelabelArabic2019,mozannarNeuralArabicQuestion2019,pratapaMultilingualEventLinking2022,chouikhiGemmArEnhancingLLMs2024,abbasEvaluationTopicIdentification2011,abdelghanyDoc2VecApproachIdentify2020,orabiClassicalArabicPoetry2020,elsaharBuildingLargeArabic2015,elnagarBRAD10Book2016,pieriBiMediXBilingualMedical2024,alghamdiArMATHDatasetSolving2022,abdallahArabicaQAComprehensiveDataset2024,biltawiArabicReadingComprehension2020,aloui101BillionArabic2024,el-khair15BillionWords2016} \\
KIWI & CC BY-SA 4.0 & \autocite{xuKIWIDatasetKnowledgeIntensive2024} \\
LMSYS-Chat-1M & LMSYS-Chat-1M Dataset License, Anthropic, Llama 2 & \autocite{zhengLMSYSChat1MLargeScaleRealWorld2024} \\
Llama2-MedTuned-Instr. & CC BY-NC 4.0 & \autocite{rohanianExploringEffectivenessInstruction2023} \\
LongAlign-10k & Anthropic, Apache License 2.0 & \autocite{baiLongAlignRecipeLong2024} \\
Magpie-Pro & Meta Llama3 Community License & \autocite{xuMagpieAlignmentData2024} \\
MathDial & CC BY-SA 4.0, MIT License & \autocite{macinaMathDialDialogueTutoring2023} \\
MathInstr. & MIT License & \autocite{yueMAmmoTHBuildingMath2023} \\
MedInstr. & Unspecified & \autocite{zhangAlpaCareInstructiontunedLarge2024} \\
Medical Meadow & Various & \autocite{hanMedAlpacaOpenSourceCollection2023,wangCORD19COVID19Open2020,jinWhatDiseaseDoes2020,saveryQuestionDrivenSummarizationAnswers2020} \\
MegaWika & CC BY-SA 4.0 & \autocite{barhamMegaWikaMillionsReports2023} \\
MetaMathQA & MIT License & \autocite{yuMetaMathBootstrapYour2023} \\
Nectar & Various & -- \\
No Robots & CC BY-NC 4.0 & -- \\
Open Asst. v2 & Apache License 2.0 & \autocite{kopfOpenAssistantConversationsDemocratizing2023} \\
Open-Platypus & Various & \autocite{sawadaARBAdvancedReasoning2023,dettmersQLoRAEfficientFinetuning2023,lightmanLetsVerifyStep2023,yuReClorReadingComprehension2020,wangSciBenchEvaluatingCollegeLevel2024,luLearnExplainMultimodal2022,chenTheoremQATheoremdrivenQuestion2023} \\
OpenGPT Healthcare & Unspecified, OGL 3.0 & -- \\
OpenMathInstr.-1 & Custom, MIT License, Apache License 2.0 & \autocite{toshniwalOpenMathInstruct1MillionMath2024} \\
Orca-Math & Various & \autocite{mitraOrcaMathUnlockingPotential2024} \\
PII-Masking-200k & Non Commercial & -- \\
PMC-LLaMA Instr. & Unspecified, Apache License 2.0 & \autocite{wuPMCLLaMABuildingOpensource2023,jinPubMedQADatasetBiomedical2019} \\
Pure-Dove & Apache License 2.0 & -- \\
PygmalionAI-PIPPA & Apache License 2.0 & \autocite{goslingPIPPAPartiallySynthetic2023} \\
Reasoning & Apache License 2.0 & -- \\
RiddleSense & MIT License & \autocite{linRiddleSenseReasoningRiddle2021} \\
SeaBench & Apache License 2.0 & \autocite{nguyenSeaLLMsLargeLanguage2023} \\
SelFee & MIT License & \autocite{yeSelFeeIterativeSelfRevising2023} \\
Synth.-GSM8K-Refl. & Meta Llama3 Community License & -- \\
Thai Gen AI & Various & -- \\
UltraChat-200k & CC BY-NC 4.0 & \autocite{dingEnhancingChatLanguage2023} \\
UltraFeedback Argilla & Various & -- \\
WildChat & AI2 ImpACT License - Low Risk & \autocite{zhaoWildChat1MChatGPT2023} \\
\end{longtable}
        
        \clearpage
        \begin{longtable}{p{5cm}|p{5cm}|p{5cm}}
\caption[\textbf{References and licenses: audio}]{\textbf{References and licenses for audio} dataset collections presented in this paper. Collections containing material under more than three distinct licenses are marked as having ``Various'' licenses, and we refer readers to our raw data for the full details. Datasets are sorted alphabetically for ease of dataset lookup.} \label{tab:refs-licenses-audio} \\
\toprule
Collection & Licenses & Cite \\
\midrule
\endfirsthead
\caption[]{\textbf{References and licenses for audio} dataset collections presented in this paper. Collections containing material under more than three distinct licenses are marked as having ``Various'' licenses, and we refer readers to our raw data for the full details. Datasets are sorted alphabetically for ease of dataset lookup.} \\
\toprule
Collection & Licenses & Cite \\
\midrule
\endhead
\midrule
\multicolumn{3}{r}{Continued on next page} \\
\midrule
\endfoot
\bottomrule
\endlastfoot
1111 Hours Hindi & Custom & \autocite{bhanushaliGramVaaniASR2022} \\
120h Spanish Speech & CC0 1.0 & -- \\
AFRISPEECH-200 & CC BY-NC-SA 4.0 & \autocite{olatunjiAfriSpeech200PanAfricanAccented2023} \\
AISHELL-1 & Apache 2.0 & \autocite{buAISHELL1OpenSourceMandarin2017} \\
AISHELL-2 & Unspecified & \autocite{duAISHELL2TransformingMandarin2018} \\
AISHELL-4 & CC BY-SA 4.0 & \autocite{fuAISHELL4OpenSource2021} \\
ALLSSTAR & CC BY 4.0 & \autocite{bradlowALLSSTARArchiveL12010} \\
AMI & CC BY 4.0 & \autocite{carlettaAMIMeetingCorpus2006} \\
Aalto Finnish Parl. & Custom & \autocite{virkkunenFinnishParliamentASR2022} \\
African Acc. French & Apache 2.0 & -- \\
Basq., Cat. and Gal. & CC BY-SA 4.0 & \autocite{kjartanssonOpenSourceHighQuality2020} \\
Bloom Speech & Various & \autocite{leongBloomLibraryMultimodal2022} \\
Bud500 & Apache 2.0, CC BY-NC-SA 4.0 & -- \\
CSJ & Custom & \autocite{maekawaCorpusSpontaneousJapanese2003} \\
CSLU 1.2 & CSLU Agreement & \autocite{landertCSLUForeignAccented2007} \\
CSLU 22 Langs. & CSLU Agreement & \autocite{landertCSLU22Languages2005} \\
ClarinPL & CC BY 4.0 & \autocite{korzinekPolishReadSpeech2017} \\
CoNASE & Custom & \autocite{coatsCorpusRegionalAmerican2019} \\
CoVoST-2 & CC0 1.0 & \autocite{wangCoVoSTMassivelyMultilingual2020} \\
Common Voice 17 & CC0 1.0 & \autocite{ardilaCommonVoiceMassivelyMultilingual2020} \\
Crowd Sourced Speech & CC BY-SA 4.0 & \autocite{kjartanssonCrowdSourcedSpeechCorpora2018} \\
Czech Parliament & CC BY 4.0 & \autocite{kratochvilLargeCorpusCzech2020} \\
DiDiSpeech & Unspecified & \autocite{guoDiDiSpeechLargeScale2021} \\
Earnings-22 & Unspecified & \autocite{delrioEarnings22PracticalBenchmark2022} \\
EdAcc & CC BY-SA 4.0 & \autocite{sanabriaEdinburghInternationalAccents2023} \\
Eng. Acc. in Brit. Isles & CC BY-SA 4.0 & \autocite{demirsahinOpensourceMultispeakerCorpora2020} \\
English-Vietnamese & CC BY-NC-ND 4.0 & \autocite{nguyenHighQualityLargeScaleDataset2022} \\
FT SPEECH & Custom & \autocite{kirkedalFTSpeechDanish2020} \\
Fisher & LDC User Agreement & \autocite{cieriFisherCorpusResource2004} \\
Fleurs & CC BY 4.0 & \autocite{conneauFLEURSFewshotLearning2022} \\
GigaSpeech & Apache 2.0 & \autocite{chenGigaSpeechEvolvingMultidomain2021} \\
Golos & Custom & \autocite{karpovGolosRussianDataset2021} \\
Hebrew Coursera & Unspecified & -- \\
Hebrew Kan & Unspecified & -- \\
Highland Puebla Nahuatl & CC BY-NC-SA 3.0 & \autocite{shiHighlandPueblaNahuatl2021} \\
JTUBESPEECH & Unspecified & \autocite{takamichiJTubeSpeechCorpusJapanese2021} \\
Japanese Anime Speech & CC0 1.0 & -- \\
KSC & CC BY 4.0 & \autocite{khassanovCrowdsourcedOpenSourceKazakh2021} \\
Kathbath & CC0 1.0 & \autocite{javedIndicSUPERBSpeechProcessing2022} \\
KeSpeech & Custom & \autocite{tangKeSpeechOpenSource2021} \\
KsponSpeech & Unspecified & \autocite{bangKsponSpeechKoreanSpontaneous2020} \\
LJSpeech & Public Domain & \autocite{itoLJSpeechDataset2017} \\
LaboroTVSpeech & Custom & \autocite{andoConstructionLargescaleJapanese2021} \\
LibriSpeech & CC BY 4.0 & \autocite{panayotovLibrispeechASRCorpus2015} \\
M-AILABS & Custom & \autocite{solakMAILABSSpeechDataset2024} \\
M2ASR & Unspecified & \autocite{shiFreeKazakhSpeech2017,mamtiminM2ASRKIRGHIZFreeKirghiz2023,zhiM2ASRMONGOFreeMongolian2021,liFreeLinguisticSpeech2017} \\
MAGICDATA & CC BY-NC-ND 4.0 & -- \\
MASC & CC BY 4.0 & \autocite{al-fetyaniMASCMassiveArabic2023} \\
MaSS & Unspecified & \autocite{boitoMaSSLargeClean2020} \\
MagicData-RAMC & CC BY-NC-ND 4.0 & \autocite{yangOpenSourceMagicDataRAMC2022} \\
MediaSpeech & CC BY 4.0 & \autocite{kolobovMediaSpeechMultilanguageASR2021} \\
Minds14 & CC BY 4.0 & \autocite{gerzMultilingualCrossLingualIntent2021} \\
MuST-C & CC BY-NC-ND 4.0 & \autocite{digangiMuSTCMultilingualSpeech2019} \\
Multiling. LibriSpeech & CC BY 4.0 & \autocite{pratapMLSLargeScaleMultilingual2020} \\
Multiling. TEDx & CC BY-NC-ND 4.0 & \autocite{saleskyMultilingualTEDxCorpus2021} \\
NST Danish & CC0 1.0 & -- \\
NST Norwegian & CC0 1.0 & -- \\
NST Swedish & CC0 1.0 & -- \\
Nigerian English & CC BY-SA 4.0 & -- \\
Norwegian Parl. & CC0 1.0 & \autocite{solbergNorwegianParliamentarySpeech2022} \\
Norwegian Parl. Speech & CC0 1.0 & \autocite{solbergNorwegianParliamentarySpeech2022} \\
OLKAVS & Custom & \autocite{parkOLKAVSOpenLargeScale2023} \\
OpenSTT & CC BY-NC 4.0 & \autocite{andrusenkoExplorationEndtoEndASR2020} \\
People's Speech & Various & \autocite{galvezPeopleSpeechLargeScale2021} \\
QASR & Unspecified & \autocite{mubarakQASRQCRIAljazeera2021} \\
RTVE & Custom & -- \\
ReazonSpeech & CDLA-Sharing-1.0 & \autocite{yinReazonSpeechFreeMassive2023} \\
Regional Af. Am. Lang. & CC BY-NC-SA 4.0 & -- \\
RixVox & CC BY 4.0 & -- \\
SDS-200 & Custom & \autocite{plussSDS200SwissGerman2022} \\
SPGISpeech & Custom & \autocite{oneillSPGISpeech000Hours2021} \\
Samromur & CC BY 4.0 & \autocite{mollbergSamromurCrowdsourcingData2020} \\
Samromur Children & CC BY 4.0 & \autocite{hernandezmenaSamromurChildrenIcelandic2022} \\
Samromur Milljon & CC BY 4.0 & -- \\
Shrutilipi & CC0 1.0 & \autocite{bhogaleEffectivenessMiningAudio2022} \\
Snow Mountain & CC BY-SA 4.0 & \autocite{rajuSnowMountainDataset2023} \\
Spoken Wikipedia & CC BY-SA 4.0 & \autocite{baumannSpokenWikipediaCorpus2019} \\
Switchboard & LDC User Agreement & \autocite{godfreySWITCHBOARDTelephoneSpeech1992} \\
TED-LIUM3 & CC BY-NC-ND 3.0 & \autocite{hernandezTEDLIUMTwiceMuch2018} \\
THCHS-30 & Apache 2.0 & \autocite{wangTHCHS30FreeChinese2015} \\
THUYG-20 & Apache 2.0 & \autocite{roziOpenFreeDatabase2015} \\
TIMIT & LDC User Agreement & \autocite{garofolojohns.TIMITAcousticPhoneticContinuous1993} \\
VCTK & CC BY 4.0 & -- \\
VibraVox & CC BY 4.0 & -- \\
VoxPopuli & CC0 1.0 & \autocite{wangVoxPopuliLargeScaleMultilingual2021} \\
Vystadial & CC BY-SA 3.0 & \autocite{korvasFreeEnglishCzech2014} \\
WenetSpeech & CC BY 4.0 & \autocite{zhangWenetSpeech10000Hours2022} \\
West Afr. Radio & CC BY-SA 4.0 & \autocite{doumbouyaUsingRadioArchives2021} \\
West Afr. Virt. Asst. & CC BY-SA 4.0 & \autocite{doumbouyaUsingRadioArchives2021} \\
YODAS & CC BY 3.0 & \autocite{liYodasYoutubeOrientedDataset2023} \\
Zeroth-Korean & CC BY 4.0 & -- \\
aidatatang & CC BY-NC-ND 4.0 & -- \\
\end{longtable}
        
        \clearpage
        \begin{longtable}{p{5cm}|p{5cm}|p{5cm}}
\caption[\textbf{References and licenses: video}]{\textbf{References and licenses for video} dataset collections presented in this paper. Collections containing material under more than three distinct licenses are marked as having ``Various'' licenses, and we refer readers to our raw data for the full details. Datasets are sorted alphabetically for ease of dataset lookup.} \label{tab:refs-licenses-video} \\
\toprule
Collection & Licenses & Cite \\
\midrule
\endfirsthead
\caption[]{\textbf{References and licenses for video} dataset collections presented in this paper. Collections containing material under more than three distinct licenses are marked as having ``Various'' licenses, and we refer readers to our raw data for the full details. Datasets are sorted alphabetically for ease of dataset lookup.} \\
\toprule
Collection & Licenses & Cite \\
\midrule
\endhead
\midrule
\multicolumn{3}{r}{Continued on next page} \\
\midrule
\endfoot
\bottomrule
\endlastfoot
100DOH & Custom & \autocite{shanUnderstandingHumanHands2020} \\
20BN-SOMETHING & Custom & \autocite{goyalSomethingSomethingVideo2017} \\
20BN-jester & Custom & \autocite{materzynskaJesterDatasetLargeScale2019} \\
50 Salads & CC BY-NC-SA 4.0 & \autocite{steinCombiningEmbeddedAccelerometers2013} \\
ActivityNet & MIT License & \autocite{heilbronActivityNetLargescaleVideo2015} \\
Apes & Unspecified & \autocite{alcazarAPESAudiovisualPerson2021} \\
Ava & CC BY 4.0 & \autocite{rothAVAActiveSpeakerAudioVisualDataset2019,guAVAVideoDataset2018} \\
Breakfast & CC BY 4.0 & \autocite{kuehneLanguageActionsRecovering2014} \\
CDAD & Unspecified & \autocite{xiangCDADCommonDaily2022} \\
COIN & Custom & \autocite{tangCOINLargescaleDataset2019} \\
Charades & Custom & \autocite{sigurdssonHollywoodHomesCrowdsourcing2016} \\
Charades-Ego & Custom & \autocite{sigurdssonActorObserverJoint2018} \\
Collective & Unspecified & \autocite{wongunchoiWhatAreThey2009} \\
Condensed Movies & CC BY 4.0 & \autocite{bainCondensedMoviesStory2020} \\
CrossTask & Unspecified & \autocite{zhukovCrosstaskWeaklySupervised2019} \\
Davis & Custom & \autocite{perazziBenchmarkDatasetEvaluation2016} \\
DiDeMo & BSD 2-Clause License & \autocite{hendricksLocalizingMomentsVideo2018} \\
Disney Vid. Gen. & Apache 2.0 & -- \\
EEV & CC BY 4.0 & \autocite{sunEEVLargeScaleDataset2021} \\
EPIC-KITCHENS & CC BY-NC 4.0 & \autocite{damenScalingEgocentricVision2018} \\
Ego4D & Custom, MIT License & \autocite{graumanEgo4DWorld0002022} \\
FERV39k & CC BY-NC 4.0 & \autocite{wangFERV39kLargeScaleMultiScene2022} \\
FineGym & CC BY-NC 4.0 & \autocite{shaoFineGymHierarchicalVideo2020} \\
FineVideo & CC BY 4.0 & -- \\
HAA500 & Unspecified & \autocite{chungHAA500HumanCentricAtomic2021} \\
HACS & Custom & \autocite{zhaoHACSHumanAction2019} \\
HD-VILA-100M & Custom & \autocite{xueAdvancingHighResolutionVideoLanguage2022} \\
HMDB & CC BY 4.0 & \autocite{kuehneHMDBLargeVideo2011} \\
HOLLYWOOD2 & Unspecified & \autocite{marszalekActionsContext2009} \\
HOMAGE & Unspecified & \autocite{raiHomeActionGenome2021} \\
HVU & Custom & \autocite{dibaLargeScaleHolistic2020} \\
Hollywood Ext. & MIT License & \autocite{bojanowskiWeaklySupervisedAction2014} \\
How2 & Various & \autocite{sanabriaHow2LargescaleDataset2018} \\
HowTo100M & Unspecified & \autocite{miechHowTo100MLearningTextVideo2019} \\
ImageNet-Vid & CC BY-NC 4.0 & \autocite{russakovskyImageNetLargeScale2015} \\
Kinetics & Unspecified & \autocite{kayKineticsHumanAction2017,carreiraShortNoteKinetics6002018,smairaShortNoteKinetics70020202020} \\
LEMMA & Unspecified & \autocite{jiaLEMMAMultiviewDataset2020} \\
LSMDC & Custom, MIT License & \autocite{rohrbachMovieDescription2016,sharmaDeepMultimodalFeature2020} \\
M-MiT & Unspecified & \autocite{monfortMultiMomentsTimeLearning2021} \\
MAD & Custom & \autocite{soldanMADScalableDataset2022} \\
MMAct & Custom & \autocite{kongMMActLargeScaleDataset2019} \\
MPII & Unspecified, Custom & \autocite{rohrbachRecognizingFineGrainedComposite2016,rohrbachDatasetMovieDescription2015} \\
MSA & Unspecified & \autocite{xiongGraphBasedFrameworkBridge2019} \\
MSR-VTT & Unspecified & \autocite{xuMSRVTTLargeVideo2016} \\
MVBench & MIT License & \autocite{liMVBenchComprehensiveMultimodal2024} \\
Mars & Unspecified & \autocite{zhengMARSVideoBenchmark2016} \\
Mimetics & Unspecified & \autocite{weinzaepfelMimeticsUnderstandingHuman2021} \\
Moments in Time & Custom & \autocite{monfortMomentsTimeDataset2019} \\
Movie-Net & Unspecified & \autocite{huangMovieNetHolisticDataset2020} \\
MovieGraphs & Custom & \autocite{vicolMovieGraphsUnderstandingHumanCentric2018} \\
MovieQA & Unspecified & \autocite{tapaswiMovieQAUnderstandingStories2016} \\
MovieScenes & Unspecified & \autocite{raoLocaltoGlobalApproachMultimodal2020} \\
MultiTHUMOS & CC BY 4.0 & \autocite{yeungEveryMomentCounts2017} \\
NTU RGB+D & Custom & \autocite{shahroudyNTURGBLarge2016} \\
Narrated Instr. Vid. & MIT License & \autocite{alayracUnsupervisedLearningNarrated2016} \\
OmniSource-Web & Apache License 2.0 & \autocite{duanOmnisourcedWeblysupervisedLearning2020} \\
Oops! & CC BY-NC-SA 4.0 & \autocite{epsteinOopsPredictingUnintentional2020} \\
OpenVid-1M & CC-BY-4.0 & \autocite{nanOpenVid1MLargeScaleHighQuality2024} \\
PKU-MMD & Unspecified & \autocite{liuPKUMMDLargeScale2017} \\
Project-Aria & Apache License 2.0 & \autocite{panAriaDigitalTwin2023,lvAriaEverydayActivities2024} \\
QFVS & Unspecified & \autocite{sharghiQueryFocusedVideoSummarization2017} \\
QuerYD & Unspecified & \autocite{oncescuQuerYDVideoDataset2021} \\
RareAct & Unspecified & \autocite{miechRareActVideoDataset2020} \\
SOA & Unspecified & \autocite{dibaLargeScaleHolistic2020} \\
ShareGPT4Video & Attribution-NonCommercial 4.0 International & \autocite{chenShareGPT4VideoImprovingVideo2024} \\
Spoken Moments & Custom & \autocite{monfortSpokenMomentsLearning2021} \\
Sports-1M & CC BY 3.0 & \autocite{karpathyLargeScaleVideoClassification2014} \\
StoryGraphs & Unspecified & \autocite{tapaswiStoryGraphsVisualizingCharacter2014} \\
SumMe & Unspecified & \autocite{gygliCreatingSummariesUser2014} \\
TGIF & Custom & \autocite{liTGIFNewDataset2016} \\
THUMOS & Custom & \autocite{idreesTHUMOSChallengeAction2017} \\
TITAN & Non Commercial & \autocite{mallaTITANFutureForecast2020} \\
TRECVid & CC BY-NC-SA 4.0 & \autocite{awadTRECVID2019Evaluation2020} \\
TVSum & CC BY 3.0 & \autocite{yalesongTVSumSummarizingWeb2015} \\
TinyVIRAT & Unspecified & \autocite{demirTinyVIRATLowresolutionVideo2020} \\
Toyota Smarthome & Custom & \autocite{dasToyotaSmarthomeRealWorld2019} \\
UAV-Human & Custom & \autocite{liUAVHumanLargeBenchmark2021} \\
UCF101 & Unspecified & \autocite{soomroUCF101Dataset1012012} \\
VIOLIN & Unspecified & \autocite{liuVIOLINLargeScaleDataset2020} \\
VLOG & Custom & \autocite{fouheyLifestyleVlogsEveryday2017} \\
VTW & Unspecified & \autocite{zengTitleGenerationUser2016} \\
VaTeX & CC BY 4.0 & \autocite{wangVATEXLargeScaleHighQuality2020} \\
VidProm & CC-BY-NC 4.0 & \autocite{wangVidProMMillionscaleReal2024} \\
VideoLT & Non Commercial & \autocite{zhangVideoLTLargescaleLongtailed2021} \\
VideoStory & Unspecified & \autocite{habibianVideoStoryNewMultimedia2014} \\
Volleyball & Unspecified & \autocite{ibrahimHierarchicalDeepTemporal2016} \\
VoxCeleb & Custom & \autocite{nagraniVoxCelebLargescaleSpeaker2018} \\
WebVid & Custom & \autocite{bainFrozenTimeJoint2022} \\
YouCook & Unspecified & \autocite{dasThousandFramesJust2013} \\
YouCook2 & MIT License & \autocite{zhouAutomaticLearningProcedures2017} \\
Youtube-8M & Unspecified & \autocite{abu-el-haijaYouTube8MLargeScaleVideo2016} \\
\end{longtable}
    \endgroup
    
    \printbibliography[
        heading=subbibliography,
        title={Attribution Card References},
    ]
\end{refsection}

\end{document}